\documentclass[letterpaper]{article} 
\usepackage[table,xcdraw]{xcolor}
\usepackage[]{aaai25}  
\usepackage{times}  
\usepackage{helvet}  
\usepackage{courier}  
\usepackage[hyphens]{url}  
\usepackage{graphicx} 
\usepackage{natbib}  
\usepackage{caption} 
\usepackage{algorithm}
\usepackage{algorithmic}

\usepackage{newfloat}
\usepackage{listings}
\DeclareCaptionStyle{ruled}{labelfont=normalfont,labelsep=colon,strut=off} 
\floatstyle{ruled}
\newfloat{listing}{tb}{lst}{}
\floatname{listing}{Listing}
\usepackage{tcolorbox}
\usepackage{amsmath}
\usepackage{multirow}

\usepackage{hhline}
\definecolor{bg}{rgb}{0.95, 0.95, 0.95}
\usepackage{bibentry}

\usepackage{fancyvrb,fvextra}

\title{Comprehensive and Practical Evaluation of Retrieval-Augmented Generation Systems for Medical Question Answering}

\author{
    Nghia Trung Ngo\textsuperscript{\rm 1}, 
    Chien Van Nguyen\textsuperscript{\rm 1}, 
    Franck Dernoncourt\textsuperscript{\rm 2}, 
    Thien Huu Nguyen \textsuperscript{\rm 1}
}
\affiliations{
    \textsuperscript{\rm 1}Department of Computer Science,University of Oregon, OR, USA \\
    \textsuperscript{\rm 2}Adobe Research, USA \\
    \{nghian,chienn,thien@cs\}@uoregon.edu,
    franck.dernoncourt@adobe.com

}

\begin{document}

\maketitle

\begin{abstract}
Retrieval-augmented generation (RAG) has emerged as a promising approach to enhance the performance of large language models (LLMs) in knowledge-intensive tasks such as those from medical domain. 
However, the sensitive nature of the medical domain necessitates a completely accurate and trustworthy system. 
While existing RAG benchmarks primarily focus on the standard retrieve-answer setting, they overlook many practical scenarios that measure crucial aspects of a reliable medical system.
This paper addresses this gap by providing a comprehensive evaluation framework for medical question-answering (QA) systems in a RAG setting for these situations, including sufficiency, integration, and robustness.
We introduce Medical Retrieval-Augmented Generation Benchmark (MedRGB) that provides various supplementary elements to four medical QA datasets for testing LLMs' ability to handle these specific scenarios.
Utilizing MedRGB, we conduct extensive evaluations of both state-of-the-art commercial LLMs and open-source models across multiple retrieval conditions.
Our experimental results reveals current models' limited ability to handle noise and misinformation in the retrieved documents.
We further analyze the  LLMs' reasoning processes to provides valuable insights and future directions for developing RAG systems in this critical medical domain.








\end{abstract}

%


\section{Introduction}

Large language models (LLMs) have demonstrated remarkable capabilities in solving complex medical problems, achieving state-of-the-art performance across various benchmarks. 
However, ensuring the reliability and truthworthiness of an artificial intelligent (AI) medical system remains a critical challenge, especially in healthcare applications.
Retrieval-augmented generation (RAG) has emerged as a promising approach to reduce LLMs' hallucination problem by integrating external knowledge sources.

\begin{figure}[ht!]
\begin{center} 
\includegraphics[width=0.48\textwidth]{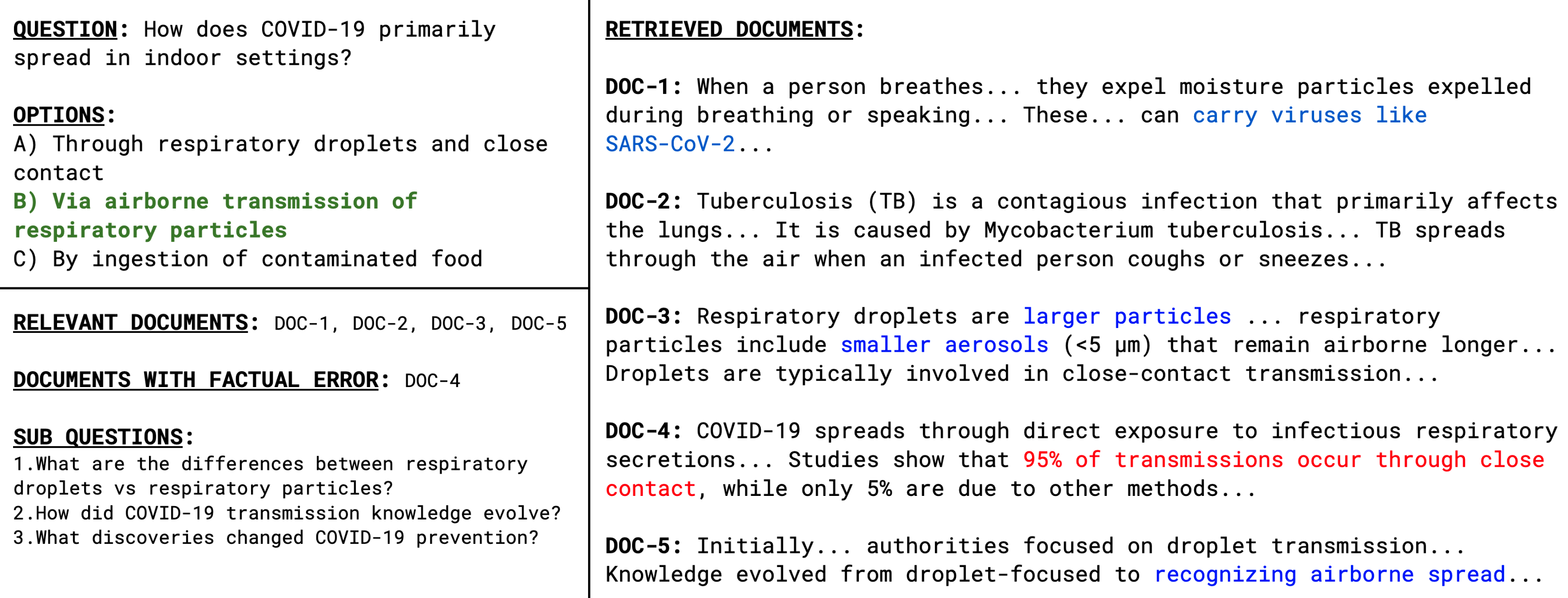}
\caption{\small Blue texts are useful information that should be extract to help determine the answer. Red texts are factual errors that potentially mislead the LLMs.} 
\label{fig:intro-example}
\end{center}
\end{figure}

While RAG has potential to improve the factual accuracy of LLMs' response, incorporating an information retriever also presents new complexities that warrant careful evaluation.
Consider the example in Fig. \ref{fig:intro-example}. The retrieved documents can contain not only useful knowledge that helps determine the true answer, but also noise information, or more serious, factual errors that can misleads the LLMs.
To consciously apply RAG for medical QA, we must consider these practical scenarios and evaluate LLMs ability to interact with retrieved documents reliably.

Recent efforts have been made to evaluate AI systems with LLMs in the medical domain \cite{nori-23, medeval, Xiong-24}.
For example, MedEval \cite{medeval} presents a large-scale, expert-annotated benchmark that cover various medical tasks and domains.
\cite{Xiong-24} evaluates RAG extensively based on their MIRAGE benchmark which cover 5 medical QA datasets.
However, they only focus on the effect of RAG modules on target accuracy, missing other important aspects of a AI medical system.

Several recent works have explore RAG evaluation more comprehensively in general domain \cite{ragas, rgb}, 
RAGAS \cite{ragas} assess 3 qualities of RAG's outputs for QA tasks including:
Faithfulness - degree to which responses align with the provided context, 
Answer Relevance - the extent to which generated responses address the actual question posed, and
Context Precision-Recall - the quality of retrieved context. 
We follow the work from \cite{rgb} which establishes Retrieval-Augmented Generation Benchmark (RGB) to measure 4 abilities required for RAG, including noise robustness, negative rejection, information integration, and counterfactual robustness.
In particular, using questions from 4 medical QA datasets from MIRAGE as basis, we create Medical Retrieval-Augmented Generation Benchmark (MedRGB) to evaluate RAG system in the following 4 test scenarios:
\begin{itemize}
    \item \textbf{Standard-RAG: } evaluates LLMs performance when presented with multiple retrieved signal documents to create a context to answer to question.
    \item \textbf{Sufficiency: } evaluates LLMs reliability when there are noise documents within the retrieved context. By adding "Insufficient Information" as an additional response option, LLMs should only answer when they are confident to have enough information to determine the correct answer. This
    requires LLMs to not only be aware of its own internal knowledge, but also be able to filter out noisy information from external documents.
    \item \textbf{Integration: } evaluates LLMs ability to answer multiple supporting questions and integrate the extracted information to help address the main question.
    \item \textbf{Robustness: } evaluates LLMs resiliency to factual errors in the retrieved context. A trustworthy AI medical system should be able detect factually incorrect documents and provide the corrected information.
\end{itemize}

In total, MedRGB consists of 3480 instances for 4 test scenarios, which is over 5 times that of RGB.
Using MedRGB, we evaluation 7 LLMs, including both state-of-the art commercial LLMs and open-source models. 
In summary, our contributions are three-fold:
\begin{itemize}
    \item We establish MedRGB with four test scenarios to evaluate LLMs for medical QA tasks in RAG settings. To best of our knowledge, it is the ﬁrst benchmark comprehensively assess medical RAG systems in these practical setting.
    \item Using MedRGB, we extensively evaluate 7 LLMs, including both state-of-the art commercial LLMs and open-source models, across multiple RAG conditions. Experiment results demonstrate their limitation in addressing the more complex scenarios.
    \item We analyzed the errors of the LLMs and their reasoning process to provide insights and suggest future directions for developing more reliable and trustworthy medical RAG systems.
\end{itemize}

\begin{figure*}[ht!]
\captionsetup{skip=2pt}
\addtolength{\belowcaptionskip}{-3mm}
\begin{center} 
\includegraphics[width=0.9\textwidth]{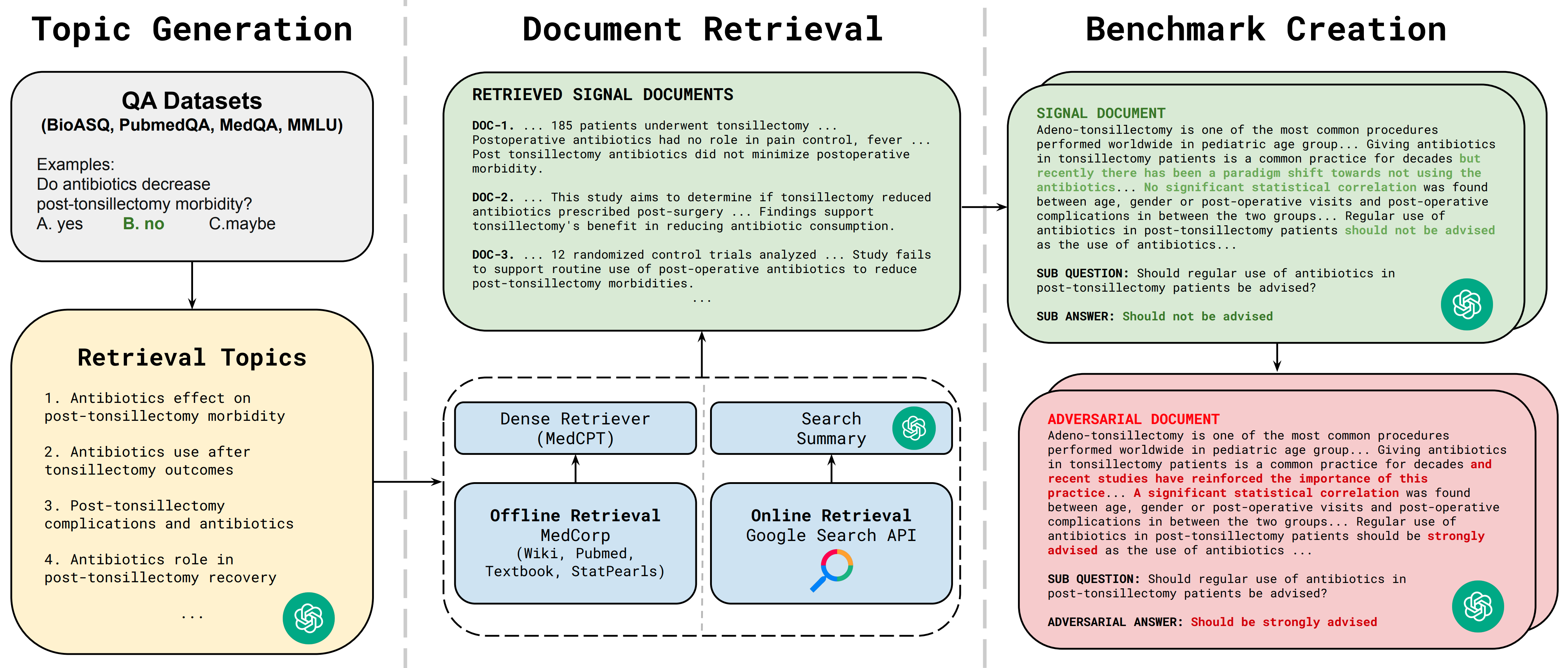}
\caption{\small The overall construction process of MedRGB. The green OpenAI symbol implies that the block involves data generation using the GPT-4o model.}
\label{fig:DataGenFig}
\end{center}
\end{figure*}

\section{Related Work}
\subsection{Medical Retrieval-augmented Generation}
The application of LLMs in medical domain demands a high level of accuracy and reliablity, which most of the current LLMs still struggle with \cite{zhou-23}.
Retrieval-augmented Generation (RAG) \cite{lewis-20} addresses this problem by helping LLMs integrating external knowledge sources in their generation process.
Recent works has achieved success in leverage RAG for knowledge-intensive tasks \cite{cui-23, peng-23, ram-23}.
Specifically for medical domain, \cite{Hiesinger-23, wang-24, Xiong-24} explore RAG for healthcare and clinical tasks.

\subsection{Medical Benchmarks}
Previous medical benchmarks usually focus solely on target performance of medical problems, which consist of only QA pairs \cite{medqa, pubmedqa, bioasq}.
Some recent benchmark also included evidence for LLMs to reasoning on \cite{chen-24}.
Most current systematic evaluations of LLMs in medical domain do not involve RAG \cite{medeval, nori-23}.
\cite{Xiong-24} attempts to provide a systematic evaluations of RAG systems in medicine.
We build on their work to further evaluate important criteria of a medical RAG system for variety of practical settings.

\begin{figure}
\captionsetup{skip=2pt}
\centering \scriptsize
\fcolorbox{bg}{bg}{
\begin{minipage}{32em}
You are a medical expert. Generate ranked search topics to help answer a medical question. Follow these guidelines: \\
\hspace*{0.5em}1. Rank topics by importance to the question. \\
\hspace*{0.5em}2. Ensure relevance to the question and answer options. \\
\hspace*{0.5em}3. The topics should be differentiable and efficient for information retrieval.
\end{minipage}
}
\caption{\small Retrieval topic generation prompt (shorten version).}
\label{fig:Topic-Gen-Prompt}
\end{figure}

\begin{figure}
\captionsetup{skip=2pt}
\centering \scriptsize
\fcolorbox{bg}{bg}{
\begin{minipage}{32em}
You are a medical expert answering a multiple-choice question using provided documents. Follow these instructions: \\
\hspace*{0.5em}1. Analyze the provide documents and question. \\
\hspace*{0.5em}2. Think step-by-step and determine the correct answer. \\
\end{minipage}
}
\caption{\small Standard-RAG test inference prompt (shorten version).}
\label{fig:sRAG-Inf-Prompt}
\end{figure}

\begin{figure}
\captionsetup{skip=2pt}
\centering \scriptsize
\fcolorbox{bg}{bg}{
\begin{minipage}{32em}
You are a medical expert answering a multiple-choice question using provided documents. Some documents may be irrelevant.
Follow these instructions: \\
\hspace*{0.5em}1. Identify relevant documents. \\
\hspace*{0.5em}2. Think step-by-step to determine the correct answer. \\
If all documents are irrelevant: \\
\hspace*{0.5em}1. Answer based on your knowledge if certain. \\
\hspace*{0.5em}2. Return \texttt{insufficient information} if unsure. \\
\end{minipage}
}
\caption{\small Sufficiency test inference prompt (shorten version).}
\label{fig:Suff-Inf-Prompt}
\end{figure}

\begin{figure}
\captionsetup{skip=2pt}
\centering \scriptsize
\fcolorbox{bg}{bg}{
\begin{minipage}{32em}
You are a medical research expert. For each provided document related to a main medical question, create a sub Q-A pair following these guidelines: \\
1. Explore different aspects related to the main question. \\
2. Sub-question must be specific to a document. \\
3. Sub-answer must be a short string extracted from the corresponding document.
\end{minipage}
}
\caption{\small Integration test data generation prompt (shorten version).}
\label{fig:Inin-Gen-Prompt}
\end{figure}

\begin{figure}
\captionsetup{skip=2pt}
\centering \scriptsize
\fcolorbox{bg}{bg}{
\begin{minipage}{32em}
You are a medical expert answering a multiple-choice main question and related sub-questions using provided documents. Some documents may be irrelevant.
Follow these instructions: \\
\hspace*{0.5em}1. Analyze all documents. \\
\hspace*{0.5em}2. Answer each sub-question using the most relevant document. Each sub-answer should be a concise string from the corresponding document.\\
\hspace*{0.5em}3. Think step-by-step, use sub-question information to determine the correct answer to the main question.
\end{minipage}
}
\caption{\small Integration test inference prompt (shorten version).}
\label{fig:Inin-Inf-Prompt}
\end{figure}

\begin{figure}
\captionsetup{skip=2pt}
\centering \scriptsize
\fcolorbox{bg}{bg}{
\begin{minipage}{32em}
You are a medical expert tasked with creating a deliberately incorrect answer-document pair for a given medical question.
Follow these instructions: \\
\hspace*{0.5em}1. Analyze the provided question, original answer, and document. \\
\hspace*{0.5em}2. Generate a deliberately incorrect new answer. \\
\hspace*{0.5em}3. Minimally edit the original document to create a persuasive but factually incorrect new document supporting the incorrect answer. \\
\end{minipage}
}
\caption{\small Robustness test data generation prompt (shorten version).}
\label{fig:Fact-Gen-Prompt}
\end{figure}

\begin{figure}
\captionsetup{skip=2pt}
\centering \scriptsize
\fcolorbox{bg}{bg}{
\begin{minipage}{32em}
You are a medical expert answering a multiple-choice main question and sub-questions using provided documents. Some documents may contain factual errors.
Follow these instructions: \\
\hspace*{0.5em} 1. For each sub-question, identify the corresponding relevant document. \\
\hspace*{0.5em} 2. Determine if the document contains any factual error. \\
\hspace*{1.2em} 2.1. Extract the sub-answer from the factually correct document. \\
\hspace*{1.2em} 2.2. Provide the factually correct answer if the document contain factual errors. \\
\hspace*{0.5em}3. Think step-by-step, use sub-question information to answer main question.
\end{minipage}
}
\caption{Robustness test inference prompt (shorten version).}
\label{fig:Fact-Inf-Prompt}
\end{figure}

\begin{figure}
\captionsetup{skip=2pt}
\centering \scriptsize
\fcolorbox{bg}{bg}{
\begin{minipage}{32em}
You are an expert in medical question answering evaluation.
Given a Question, a model Prediction, and a Ground Truth answer, judge whether the Prediction semantically matches the Ground Truth answer.
Follow the instructions: \\
\hspace*{0.5em} 1. If the prediction semantically matches the ground truth completely, score 1. \\
\hspace*{0.5em} 2. If the prediction semantically matches some part of the ground truth and is relevant to the question, score 0.5. \\
\hspace*{0.5em} 3. If the prediction is completely wrong or irrelevant to the question, score 0.
\end{minipage}
}
\caption{GPT-based Scoring Prompt (shorten version).}
\label{fig:GPT-Score-Prompt}
\end{figure}

\section{Medical Retrieval-Augmented Generation Benchmark}

MedRGB creation process is demonstrated in Fig. \ref{fig:DataGenFig}, which involves three main steps: retrieval topic generation, document retrieval, and benchmark creation.

\subsection{Medical QA Dataset}
The basis of MedRGB are the multiple-choice questions from 4 medical QA dataset from MIRAGE, including 
2 from medical examination (MMLU and MedQA) and 2 from biomedical research (PubMedQA, BioASQ).

\paragraph{MMLU} (MMLU-Med in \cite{Xiong-24})  a subset of six tasks from 57 tasks of MMLU \cite{mmlu}, including anatomy, clinical knowledge, professional medicine, human genetics, college medicine, and college biologys. There are the 1089 questions with 4 answer options.

\paragraph{MedQA} (MedQA-US in \cite{Xiong-24})  the English portion of MedQA \cite{medqa}, which is a multi-choice QA dataset collected from professional medical board exams. These are 1273 real-world questions with 4 answer options.

\paragraph{PubMedQA} (PubMedQA* in \cite{Xiong-24}) consist of 500 expert-annotated test samples of PubMedQA \cite{pubmedqa}. Each question is constructed from PubMed abstracts with yes/no/maybe answer options.

\paragraph{BioASQ} (BioASQ-Y/N in \cite{Xiong-24}) test set of the recent annual competition for biomedical QA from 2019 to 2023 \cite{bioasq}. In total, there are 618 yes/no question, constructed based on biomedical literature for machine reading comprehension track.


\subsection{Topic Generation}


In a conventional RAG setup for QA tasks, a dense retriever encodes the input question into a high-dimensional vector to find semantically similar passages from an external corpus.
The top-k documents (based on a imilarity metric like cosine similarity) are then combined to create the context for LLMs to answer the original question.
This approach often results in redundancy in the retrieved set due to the lack of diversity when using the similarity metric.
To address this limitation, we first ask LLMs to generate a diverse set of retrieval (sub-)topics using the prompt from Fig. \ref{fig:Topic-Gen-Prompt}.
Each topic is then following the above process to create a smaller set of relevant documents.
These sets are then aggregated to create the final retrieval set for the original QA pair.


\subsection{Document Retrieval}
We incorporate two retrieval processes in MedRGB to simulate the real-world difference between the use case of an expert and a non-expert person.

\paragraph{Offline Retrieval Process } In the expert use case, the retrieval corpus should contain highly specialized information that should not be publicly available.
We use MedCorp \cite{Xiong-24} as our offline corpus which contains medical documents from 4 different sources, including Pubmed, StatPearls, Textbooks and Wikipedia.
The retrieval topics from previous step are encoded by MedCPT \cite{medcpt}, a biomedical-domain dense retriever, to query the corpus for relavent documents.
We choose MedCPT instead of a general-domain retriever like Contriever \cite{contriver} or a lexical retriever like BM25 \cite{bm25} due to its consistent performance retrieving from medical domain, as demonstrateed in \cite{Xiong-24}. Top-3 documents for each retrieval topics are collected to create the retrieval set.

\paragraph{Online Retrieval Process}
In non-expert use case, user simply asks a question to general LLMs, which then retrieves document through online search engine to help answer user's query.
For each of the original medical question, the generated sub-topics are used to query the internet through Google Custom Search API\footnote{https://developers.google.com/custom-search/v1/overview}, which return top-scored links.
The content from each retrieved link is then extracted and summarized (by the LLMs) to create a signal document of the corresponding topic.


    



\subsection{Benchmark Creation}
This section describes the construction process of each of the four test scenarios. Aside from the standard-RAG test, each of the other three settings is evaluated across multiple degrees of noise, specified by the variable $p$ - the percentage of signal documents in the retrieved context.

\paragraph{Standard-RAG Test}
In this setting, the retrieved context consists of a predefined number of signal documents, which are sampled from the signal set.
The LLMs are instructed, using the prompt in Fig. \ref{fig:sRAG-Inf-Prompt}, to first generate their step-by-step reasoning, before outputting their answer option.

\paragraph{Sufficiency Test} 
For each question, we sample both signal documents and  irrelevant noise documents to create retrieval set with $p$ in $[0, 20, 40, 60, 80, 100]$.
Given the mixed context, LLMs are prompted, according to Fig. \ref{fig:Suff-Inf-Prompt}, to answer the question with an additional "Insufficient Information" option.
The LLMs are instructed to detect noise documents first, before proceeding with their step-by-step reasoning based only on the relevant documents.

\paragraph{Integration Test} 
A complex medical question may requires LLMs to address multiple sub-problems first before attempting to solve the main question.
we measure this ability of LLMs by generating a sub-question based on each signal document, according to the data generation prompt in Fig. \ref{fig:Inin-Gen-Prompt}. 
The LLMs are instructed, according to the prompt in Fig. \ref{fig:Inin-Inf-Prompt}, to find the specific relevant document for each sub-question in the noisy context, then extract the sub-answer from the corresponding document.
This test measures LLMs ability to answers an increasing number of sub-questions, and ability to integrate these information in their reasoning to infer the answer to the main question.




\paragraph{Robustness Test} 
The robustness test aim to measure LLMs resilient to factual errors, especially those that are adversarially designed to cause misinformation.
Based on the sub-questions from the Integration test, we alter both the sub-answer and the corresponding document to create a counterfactual example, using the prompt in the Fig. \ref{fig:Fact-Gen-Prompt}.
The adversarial answer should semantically contradict the original answer, and the relevant document should be minimumly edited in a convincing way.
In this test, all documents in retrieved context are relevant, and $p$ is the percentage of factually correct documents.
The LLMs are instructed, according to the prompt in Fig. \ref{fig:Fact-Gen-Prompt}, to detect documents with factually incorrect information before answering the sub-questions and the main question.





\begin{figure}
\begin{center} 
\includegraphics[width=0.49\textwidth]{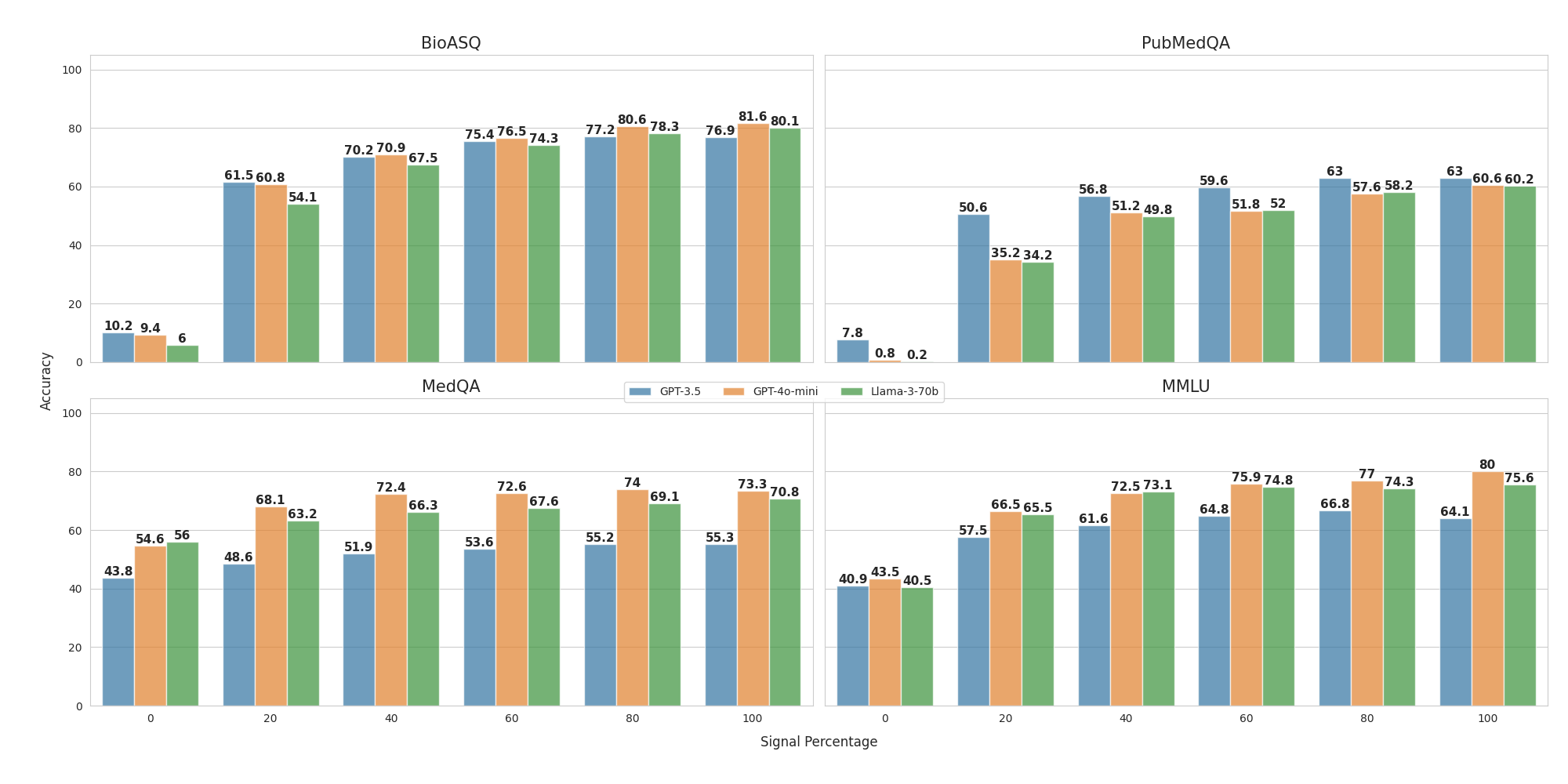}
\caption{\small Sufficiency test main question accuracy.} \label{fig:Suff-1}
\end{center}
\end{figure}

\begin{figure}
\begin{center} 
\includegraphics[width=0.49\textwidth]{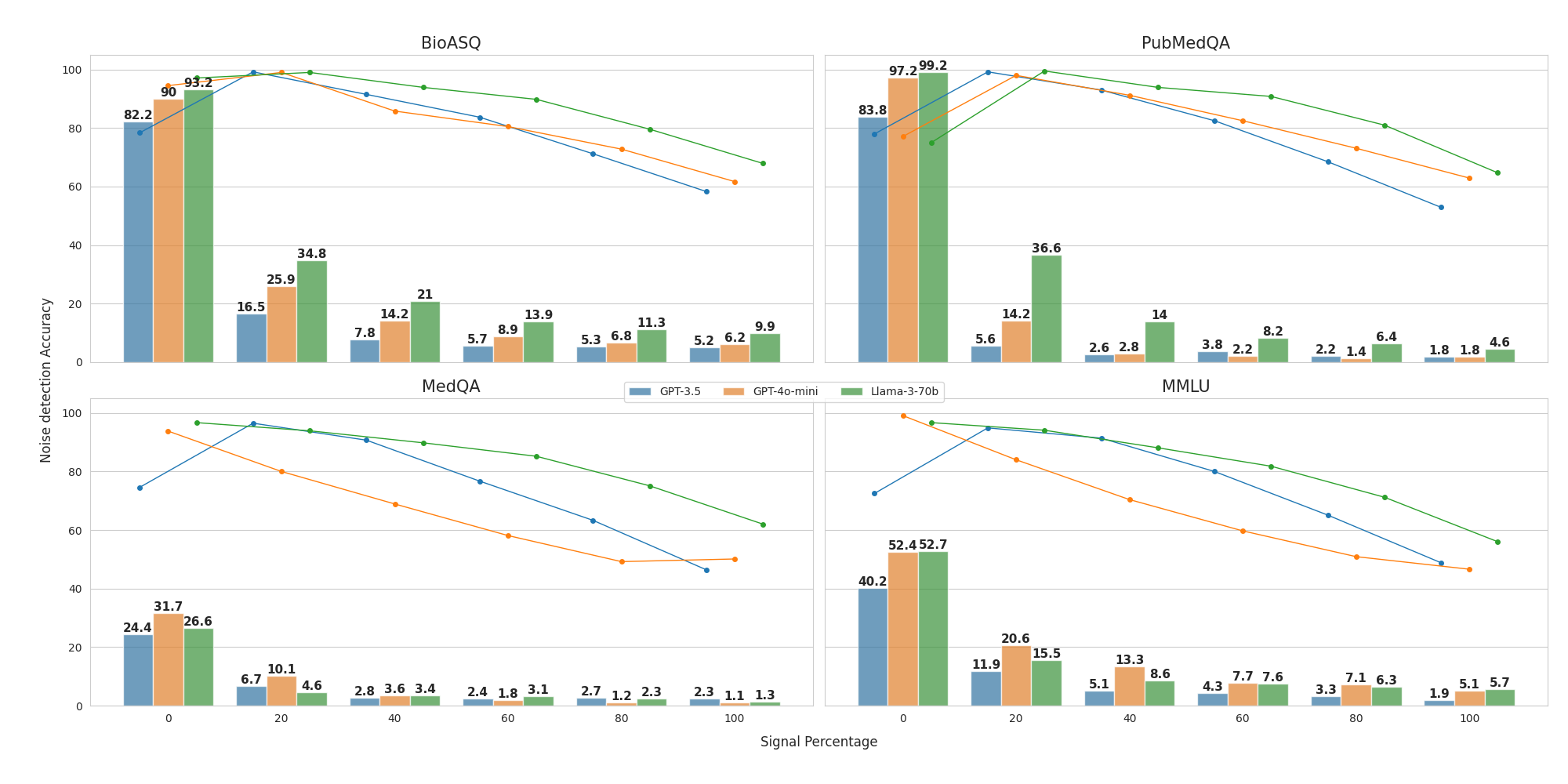}
\caption{\small Sufficiency test percentage of information insufficiency (histogram) and noise detection rate (line graph).} \label{fig:Suff-2}
\end{center}
\end{figure}

\begin{table*}[htbp]
\center
\small
\resizebox{0.99\textwidth}{!}{
\begin{tabular}{c|ccccc|ccccc|ccccc|ccccc}
 &
  \multicolumn{5}{c|}{\textbf{BioASQ}} &
  \multicolumn{5}{c|}{\textbf{PubmedQA}} &
  \multicolumn{5}{c|}{\textbf{MedQA}} &
  \multicolumn{5}{c}{\textbf{MMLU}} \\ \hhline{~|-|-|-|-|-|-|-|-|-|-|-|-|-|-|-|-|-|-|-|-|}
 &
  \multicolumn{1}{c|}{\cellcolor[HTML]{FFFFFF}} &
  \multicolumn{2}{c|}{\cellcolor[HTML]{EFEFEF}\textbf{Offline Retrieval}} &
  \multicolumn{2}{c|}{\textbf{Online Retrieval}} &
  \multicolumn{1}{c|}{\cellcolor[HTML]{FFFFFF}} &
  \multicolumn{2}{c|}{\cellcolor[HTML]{EFEFEF}\textbf{Offline Retrieval}} &
  \multicolumn{2}{c|}{\textbf{Online Retrieval}} &
  \multicolumn{1}{c|}{\cellcolor[HTML]{FFFFFF}} &
  \multicolumn{2}{c|}{\cellcolor[HTML]{EFEFEF}\textbf{Offline Retrieval}} &
  \multicolumn{2}{c|}{\textbf{Online Retrieval}} &
  \multicolumn{1}{c|}{\cellcolor[HTML]{FFFFFF}} &
  \multicolumn{2}{c|}{\cellcolor[HTML]{EFEFEF}\textbf{Offline Retrieval}} &
  \multicolumn{2}{c}{\textbf{Online Retrieval}} \\
\multirow{-3}{*}{\textbf{LLMs}} &
  \multicolumn{1}{c|}{\multirow{-2}{*}{\cellcolor[HTML]{FFFFFF}\textbf{\begin{tabular}[c]{@{}c@{}}No \\ Retrieval\end{tabular}}}} &
  \multicolumn{1}{c|}{\cellcolor[HTML]{EFEFEF}\textbf{5 doc}} &
  \multicolumn{1}{c|}{\cellcolor[HTML]{EFEFEF}\textbf{20 doc}} &
  \multicolumn{1}{c|}{\textbf{5 doc}} &
  \textbf{20 doc} &
  \multicolumn{1}{c|}{\multirow{-2}{*}{\cellcolor[HTML]{FFFFFF}\textbf{\begin{tabular}[c]{@{}c@{}}No \\ Retrieval\end{tabular}}}} &
  \multicolumn{1}{c|}{\cellcolor[HTML]{EFEFEF}\textbf{5 doc}} &
  \multicolumn{1}{c|}{\cellcolor[HTML]{EFEFEF}\textbf{20 doc}} &
  \multicolumn{1}{c|}{\textbf{5 doc}} &
  \textbf{20 doc} &
  \multicolumn{1}{c|}{\multirow{-2}{*}{\cellcolor[HTML]{FFFFFF}\textbf{\begin{tabular}[c]{@{}c@{}}No \\ Retrieval\end{tabular}}}} &
  \multicolumn{1}{c|}{\cellcolor[HTML]{EFEFEF}\textbf{5 doc}} &
  \multicolumn{1}{c|}{\cellcolor[HTML]{EFEFEF}\textbf{20 doc}} &
  \multicolumn{1}{c|}{\textbf{5 doc}} &
  \textbf{20 doc} &
  \multicolumn{1}{c|}{\multirow{-2}{*}{\cellcolor[HTML]{FFFFFF}\textbf{\begin{tabular}[c]{@{}c@{}}No \\ Retrieval\end{tabular}}}} &
  \multicolumn{1}{c|}{\cellcolor[HTML]{EFEFEF}\textbf{5 doc}} &
  \multicolumn{1}{c|}{\cellcolor[HTML]{EFEFEF}\textbf{20 doc}} &
  \multicolumn{1}{c|}{\textbf{5 doc}} &
  \textbf{20 doc} \\ \hline
\textbf{GPT-3.5} &
  \multicolumn{1}{c|}{\cellcolor[HTML]{FFFFFF}77.7} &
  \multicolumn{1}{c|}{\cellcolor[HTML]{EFEFEF}81.2} &
  \multicolumn{1}{c|}{\cellcolor[HTML]{EFEFEF}87.2} &
  \multicolumn{1}{c|}{87.2} &
  87.9 &
  \multicolumn{1}{c|}{\cellcolor[HTML]{FFFFFF}49.8} &
  \multicolumn{1}{c|}{\cellcolor[HTML]{EFEFEF}59.6} &
  \multicolumn{1}{c|}{\cellcolor[HTML]{EFEFEF}71.0} &
  \multicolumn{1}{c|}{58.4} &
  60.6 &
  \multicolumn{1}{c|}{\cellcolor[HTML]{FFFFFF}68.3} &
  \multicolumn{1}{c|}{\cellcolor[HTML]{EFEFEF}63.0} &
  \multicolumn{1}{c|}{\cellcolor[HTML]{EFEFEF}67.3} &
  \multicolumn{1}{c|}{68.0} &
  68.4 &
  \multicolumn{1}{c|}{\cellcolor[HTML]{FFFFFF}76.3} &
  \multicolumn{1}{c|}{\cellcolor[HTML]{EFEFEF}70.3} &
  \multicolumn{1}{c|}{\cellcolor[HTML]{EFEFEF}73.0} &
  \multicolumn{1}{c|}{75.7} &
  74.8 \\
\textbf{GPT-4o-mini} &
  \multicolumn{1}{c|}{\cellcolor[HTML]{FFFFFF}82.9} &
  \multicolumn{1}{c|}{\cellcolor[HTML]{EFEFEF}85.3} &
  \multicolumn{1}{c|}{\cellcolor[HTML]{EFEFEF}90.5} &
  \multicolumn{1}{c|}{89.0} &
  90.0 &
  \multicolumn{1}{c|}{\cellcolor[HTML]{FFFFFF}47.0} &
  \multicolumn{1}{c|}{\cellcolor[HTML]{EFEFEF}60.8} &
  \multicolumn{1}{c|}{\cellcolor[HTML]{EFEFEF}71.8} &
  \multicolumn{1}{c|}{60.6} &
  61.2 &
  \multicolumn{1}{c|}{\cellcolor[HTML]{FFFFFF}79.2} &
  \multicolumn{1}{c|}{\cellcolor[HTML]{EFEFEF}77.1} &
  \multicolumn{1}{c|}{\cellcolor[HTML]{EFEFEF}79.5} &
  \multicolumn{1}{c|}{79.0} &
  80.6 &
  \multicolumn{1}{c|}{\cellcolor[HTML]{FFFFFF}88.3} &
  \multicolumn{1}{c|}{\cellcolor[HTML]{EFEFEF}84.6} &
  \multicolumn{1}{c|}{\cellcolor[HTML]{EFEFEF}87.3} &
  \multicolumn{1}{c|}{86.0} &
  87.1 \\
\textbf{GPT-4o} &
  \multicolumn{1}{c|}{\cellcolor[HTML]{FFFFFF}87.9} &
  \multicolumn{1}{c|}{\cellcolor[HTML]{EFEFEF}86.1} &
  \multicolumn{1}{c|}{\cellcolor[HTML]{EFEFEF}90.8} &
  \multicolumn{1}{c|}{87.4} &
  87.4 &
  \multicolumn{1}{c|}{\cellcolor[HTML]{FFFFFF}52.6} &
  \multicolumn{1}{c|}{\cellcolor[HTML]{EFEFEF}59.2} &
  \multicolumn{1}{c|}{\cellcolor[HTML]{EFEFEF}71.2} &
  \multicolumn{1}{c|}{53.2} &
  54.4 &
  \multicolumn{1}{c|}{\cellcolor[HTML]{FFFFFF}89.5} &
  \multicolumn{1}{c|}{\cellcolor[HTML]{EFEFEF}83.7} &
  \multicolumn{1}{c|}{\cellcolor[HTML]{EFEFEF}86.9} &
  \multicolumn{1}{c|}{84.6} &
  86.9 &
  \multicolumn{1}{c|}{\cellcolor[HTML]{FFFFFF}93.4} &
  \multicolumn{1}{c|}{\cellcolor[HTML]{EFEFEF}88.3} &
  \multicolumn{1}{c|}{\cellcolor[HTML]{EFEFEF}90.1} &
  \multicolumn{1}{c|}{89.5} &
  89.1 \\ \hline\hline
\textbf{PMC-LLAMA-13b} &
  \multicolumn{1}{c|}{\cellcolor[HTML]{FFFFFF}64.2} &
  \multicolumn{1}{c|}{\cellcolor[HTML]{EFEFEF}64.6} &
  \multicolumn{1}{c|}{\cellcolor[HTML]{EFEFEF}64.6} &
  \multicolumn{1}{c|}{63.9} &
  64.1 &
  \multicolumn{1}{c|}{\cellcolor[HTML]{FFFFFF}55.4} &
  \multicolumn{1}{c|}{\cellcolor[HTML]{EFEFEF}54.0} &
  \multicolumn{1}{c|}{\cellcolor[HTML]{EFEFEF}54.0} &
  \multicolumn{1}{c|}{54.8} &
  54.6 &
  \multicolumn{1}{c|}{\cellcolor[HTML]{FFFFFF}44.5} &
  \multicolumn{1}{c|}{\cellcolor[HTML]{EFEFEF}38.9} &
  \multicolumn{1}{c|}{\cellcolor[HTML]{EFEFEF}38.8} &
  \multicolumn{1}{c|}{43.4} &
  43.7 &
  \multicolumn{1}{c|}{\cellcolor[HTML]{FFFFFF}49.7} &
  \multicolumn{1}{c|}{\cellcolor[HTML]{EFEFEF}43.7} &
  \multicolumn{1}{c|}{\cellcolor[HTML]{EFEFEF}44.0} &
  \multicolumn{1}{c|}{48.4} &
  48.2 \\
\textbf{MEDITRON-70b} &
  \multicolumn{1}{c|}{\cellcolor[HTML]{FFFFFF}68.8} &
  \multicolumn{1}{c|}{\cellcolor[HTML]{EFEFEF}74.0} &
  \multicolumn{1}{c|}{\cellcolor[HTML]{EFEFEF}74.8} &
  \multicolumn{1}{c|}{79.8} &
  79.2 &
  \multicolumn{1}{c|}{\cellcolor[HTML]{FFFFFF}53.0} &
  \multicolumn{1}{c|}{\cellcolor[HTML]{EFEFEF}53.4} &
  \multicolumn{1}{c|}{\cellcolor[HTML]{EFEFEF}47.8} &
  \multicolumn{1}{c|}{58.8} &
  46.8 &
  \multicolumn{1}{c|}{\cellcolor[HTML]{FFFFFF}51.7} &
  \multicolumn{1}{c|}{\cellcolor[HTML]{EFEFEF}56.0} &
  \multicolumn{1}{c|}{\cellcolor[HTML]{EFEFEF}57.4} &
  \multicolumn{1}{c|}{61.8} &
  62.9 &
  \multicolumn{1}{c|}{\cellcolor[HTML]{FFFFFF}65.3} &
  \multicolumn{1}{c|}{\cellcolor[HTML]{EFEFEF}65.1} &
  \multicolumn{1}{c|}{\cellcolor[HTML]{EFEFEF}66.3} &
  \multicolumn{1}{c|}{67.6} &
  69.3 \\ \hline\hline
\textbf{GEMMA-2-27b} &
  \multicolumn{1}{c|}{\cellcolor[HTML]{FFFFFF}80.3} &
  \multicolumn{1}{c|}{\cellcolor[HTML]{EFEFEF}83.3} &
  \multicolumn{1}{c|}{\cellcolor[HTML]{EFEFEF}88.7} &
  \multicolumn{1}{c|}{88.7} &
  89.2 &
  \multicolumn{1}{c|}{\cellcolor[HTML]{FFFFFF}41.0} &
  \multicolumn{1}{c|}{\cellcolor[HTML]{EFEFEF}52.0} &
  \multicolumn{1}{c|}{\cellcolor[HTML]{EFEFEF}59.0} &
  \multicolumn{1}{c|}{52.6} &
  49.4 &
  \multicolumn{1}{c|}{\cellcolor[HTML]{FFFFFF}71.2} &
  \multicolumn{1}{c|}{\cellcolor[HTML]{EFEFEF}69.8} &
  \multicolumn{1}{c|}{\cellcolor[HTML]{EFEFEF}71.7} &
  \multicolumn{1}{c|}{75.9} &
  76.9 &
  \multicolumn{1}{c|}{\cellcolor[HTML]{FFFFFF}83.5} &
  \multicolumn{1}{c|}{\cellcolor[HTML]{EFEFEF}77.9} &
  \multicolumn{1}{c|}{\cellcolor[HTML]{EFEFEF}82.5} &
  \multicolumn{1}{c|}{82.2} &
  83.6 \\
\textbf{Llama-3-70b} &
  \multicolumn{1}{c|}{\cellcolor[HTML]{FFFFFF}82.9} &
  \multicolumn{1}{c|}{\cellcolor[HTML]{EFEFEF}84.6} &
  \multicolumn{1}{c|}{\cellcolor[HTML]{EFEFEF}89.3} &
  \multicolumn{1}{c|}{89.3} &
  89.3 &
  \multicolumn{1}{c|}{\cellcolor[HTML]{FFFFFF}59.2} &
  \multicolumn{1}{c|}{\cellcolor[HTML]{EFEFEF}77.6} &
  \multicolumn{1}{c|}{\cellcolor[HTML]{EFEFEF}70.8} &
  \multicolumn{1}{c|}{59.4} &
  59.2 &
  \multicolumn{1}{c|}{\cellcolor[HTML]{FFFFFF}82.9} &
  \multicolumn{1}{c|}{\cellcolor[HTML]{EFEFEF}73.6} &
  \multicolumn{1}{c|}{\cellcolor[HTML]{EFEFEF}79.4} &
  \multicolumn{1}{c|}{76.1} &
  78.3 &
  \multicolumn{1}{c|}{\cellcolor[HTML]{FFFFFF}85.2} &
  \multicolumn{1}{c|}{\cellcolor[HTML]{EFEFEF}77.6} &
  \multicolumn{1}{c|}{\cellcolor[HTML]{EFEFEF}83.4} &
  \multicolumn{1}{c|}{81.8} &
  83.8
\end{tabular}
}
\caption{ \small Standard-RAG test accuracy.}
\label{tab:rag}
\end{table*}

\section{Experiments}


This section assesses various LLMs in the four scenarios of MedRGB, analyzing their reasoning process and discussing key findings from the experimental results.

\subsection{Evaluation Setting}


We evaluate a wide range of state-of-the-art LLMs using MedRGB. For closed commercial LLMs, we assess both GPT-3.5\footnote{https://platform.openai.com/docs/models/gpt-3-5-turbo} and GPT-4o\footnote{https://platform.openai.com/docs/models/gpt-4o} from OpenAI. Additionally, we evaluate the recent GPT-4o-mini\footnote{https://platform.openai.com/docs/models/gpt-4o-mini}, which achieves performance almost comparable to its full-sized counterpart\footnote{https://openai.com/index/gpt-4o-mini-advancing-cost-efficient-intelligence/}. 
In the open-source category, we examine both general LLMs and domain-specific fine-tuned models. The former includes the instruction-tuned Gemma-2-27b\footnote{https://huggingface.co/google/gemma-2-27b-it-pytorch} from Microsoft and Llama-3-70b\footnote{https://huggingface.co/meta-llama/Meta-Llama-3-70B-Instruct} from Meta. 
For domain-specific models, we include the medical-domain pretrained MEDITRON-70b \cite{meditron} and PMC-Llama-13b \cite{pmc}, which are tailored for healthcare applications.










\subsection{Standard-RAG Evaluation}
Table \ref{tab:rag} shows the accuracy of all considered LLMs for 3 settings: the baseline "No Retrieval" and standard-RAG setting with 5 and 20 signal documents retrieved as context.

\paragraph{Results} 
GPT-4o emerges as the the top performer across most settings, demonstrating the positive effects of scaling both parameters and training data. 
Surprisingly, GPT-4o-mini, despite having only 8 billion reported parameters, achieved comparable results to its larger counterpart. 
Among open-source models, Gemma-2-27b and Llama-3-70b show strong performance, highlighting the effectiveness of general domain instruction-tuned models in zero-shot settings. In contrast, domain-specific fine-tuned models like PMC-Llama-13 and MEDITRON-70b both yield mixed results.

The effectiveness of RAG varies across different factors.
Large models with strong internal knowledge, such as GPT-4o and Llama-3-70b, benefit less from RAG compared to smaller models like GPT-4o-mini and Gemma-2-27b.
While adding more retrieved documents generally improves performance, models with shorter context lengths (e.g., PMC-Llama-13b, MEDITRON-70b) struggle to fully utilize this additional information, resulting in only marginal improvements.

\paragraph{Analysis}
Interestingly, the impact of document quantity differs between the two retrieval sources. 
The search-based online retrieval often performs best with fewer documents, while the offline retrieval using MedCorp typically improves with more documents.
This discrepancy likely stems from the nature of each search algorithm and retrieval source. Google Search tends to provide high-quality top results but introduces more noise as the number of results increases. In contrast, retrieving from MedCorp with MedCPT offers more consistently relevant, high-quality documents, potentially more valuable in larger quantities.
We also observe a slight adverse effect when applying RAG on GPT-4o and Llama-3-70b for MMLU and MedQA. This could be attributed to their strong internal knowledge and potential data leakage issues with more popular datasets.

The following evaluations focus on more specific abilities of LLMs in RAG setting. Due to constraint on computation and cost, we will focus on GPT-3.5 and GPT-4o-mini (which is 100x time cheaper than GPT-4o while achieving relatively comparable results.) for commercial LLMs, and the best open-source model LLama-3-70b.

\subsection{Sufficiency Evaluation}

We assess the models' ability to handle varying levels of noise in retrieved documents and their capacity to recognize when they lack sufficient information to answer a question reliably. The results are shown in Fig. \ref{fig:Suff-1} and \ref{fig:Suff-2}.

\paragraph{Result}
We observe that at $p = 0$, the models mostly returns "insufficient information".
However, there is a significant increase in accuracy across all datasets as the signal percentage $p$ increased from 0 to 20.
This indicates even a small amount of signal greatly enhances the model's confidence to answer.
However, the improvement diminishes as more signal documents are added. 
Even when the retrieved context contains all signal documents ($p = 100$), performance is dramatically reduced compared to the standard RAG setting. 
This suggests that in the standard-RAG test, models may attempt to answer questions even when they are not fully confident or lack sufficient context, which can be an undesirable characteristic for medical application.

Llama-3-70b consistently outperforms other models in noise detection across all datasets and settings.
Adding More retrieved documents generally leads to improved performance with fewer "insufficient information" responses due to the increased context.
However, this also resulted in decreased noise detection accuracy, as models occasionally misinterpreted noise documents as part of the signal. This highlights the delicate balance between providing enough context for accurate answers and maintaining the ability to discern relevant information from noise.

\paragraph{Analysis}
Fig. \ref{fig:Suff-2} shows all models have high noise detection accuracy at low values of $p$, but this accuracy decreases as the signal percentage rises.
In particular, models tend to ignore the documents completely at $p = 0$, relying solely on their internal knowledge.
While this makes detecting noise documents much easier, it can lead to "insufficient information" responses even when the model has the internal knowledge to answer.
Surprisingly, when the context consists only of signal documents ($p = 100$), models struggle to identify them as relevant. 
We hypothesize that without a "specific description" of relevant documents, the model's internal relevant criteria become much stricter when only signal documents are present.
In contrast, when there is a mix of noise and signal in the retrieved context, models can more effectively infer the criteria for distinguishing between the contents of the noise and signal documents.
Finally, we observe some cases where a slight performance improvement at $p \in [60, 80]$ compared to the noiseless setting at $p = 100$.
This suggests that introducing a small amount of noise might be beneficial, similar to the concept of dropout training, where noise helps models generalize better by preventing overfitting to specific patterns. This approach could enhance the models' ability to discern relevant information by providing a more varied context for evaluation.

\begin{figure}
\begin{center} 
\includegraphics[width=0.49\textwidth]{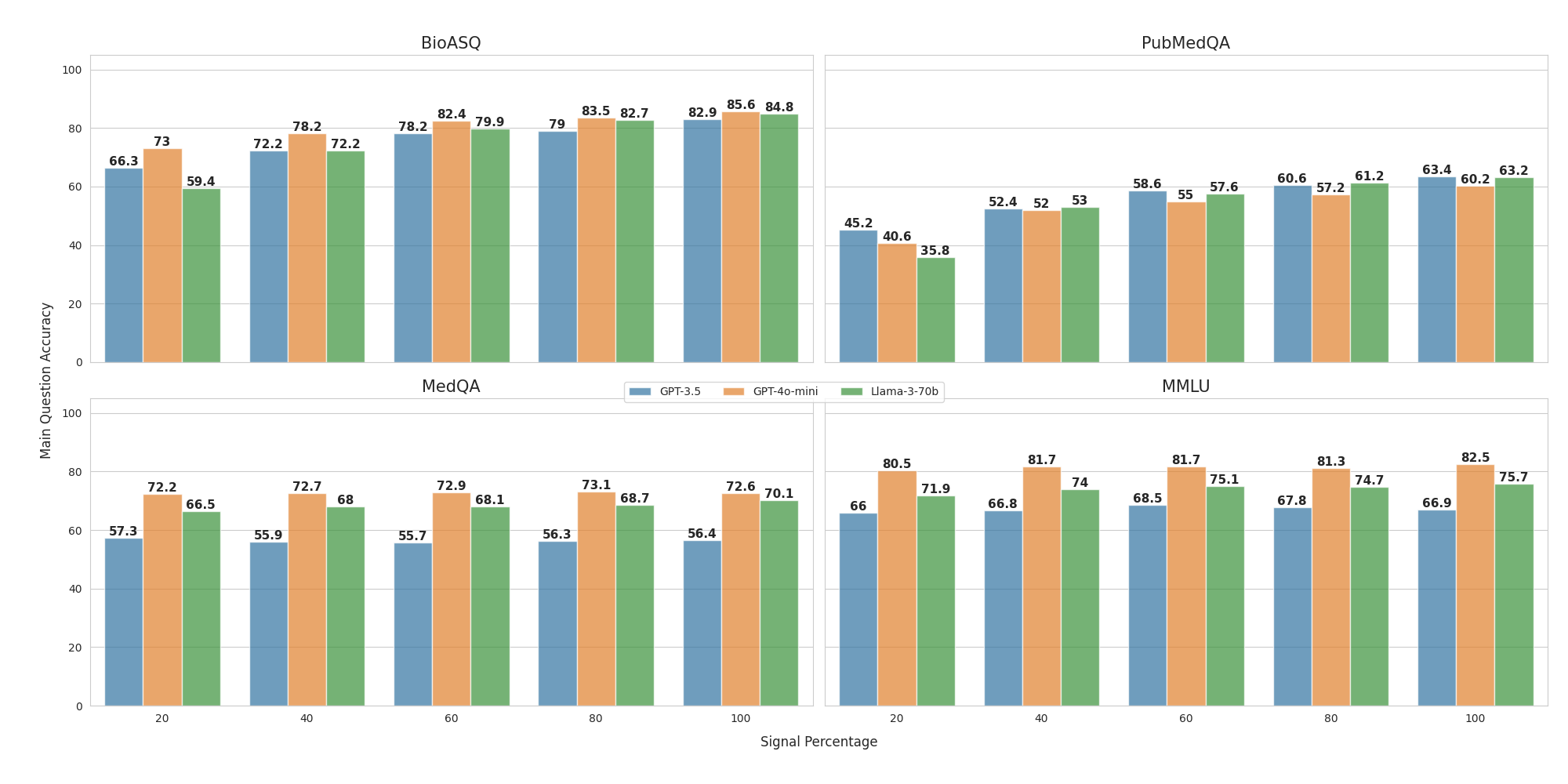}
\caption{\small Integration test main question accuracy.} \label{fig:Inin-1}
\end{center}
\end{figure}

\begin{figure}
\begin{center} 
\includegraphics[width=0.49\textwidth]{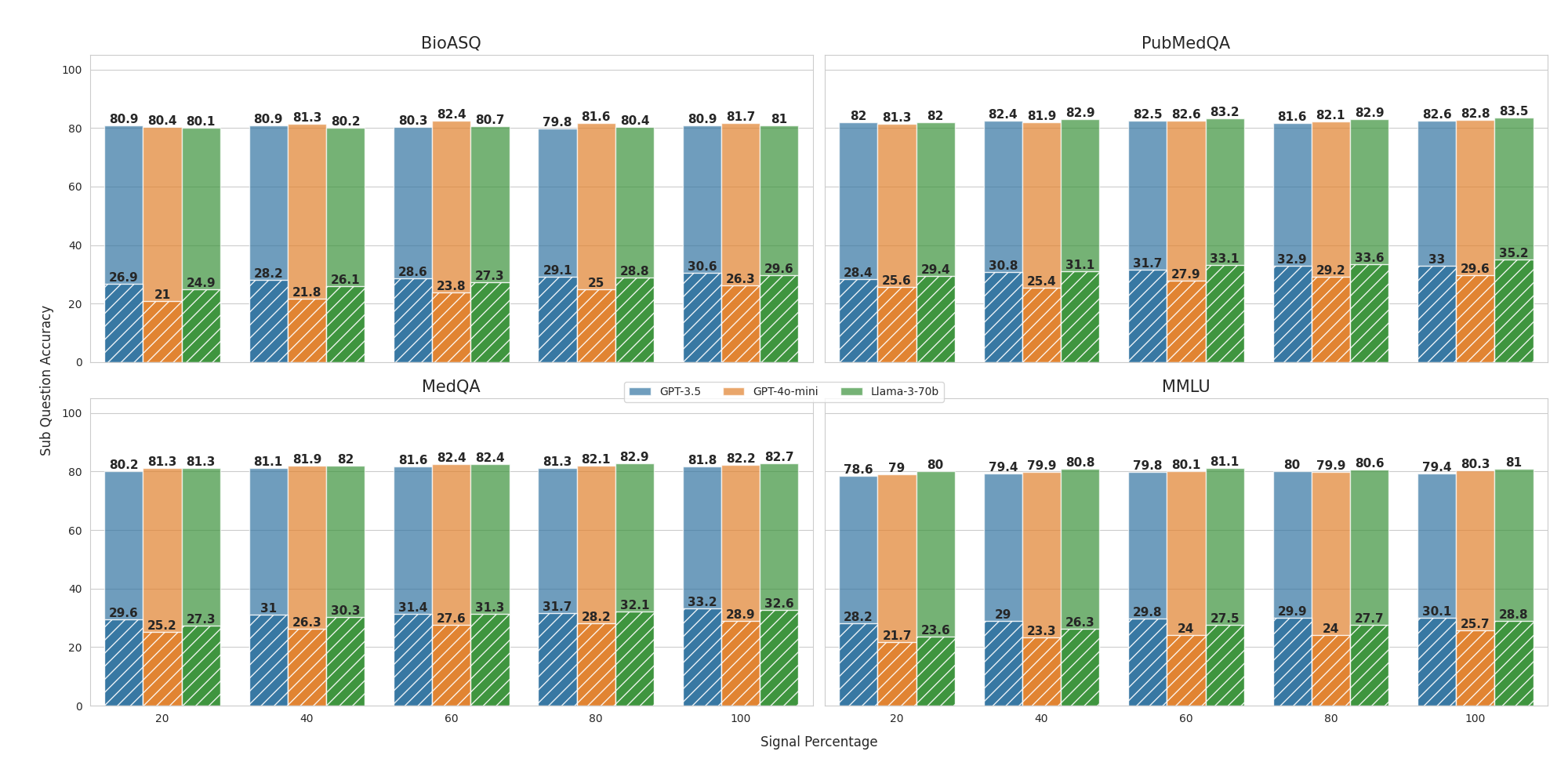}
\caption{\small Integration test sub-question GPT-based score (filled histogram) and exact-match accuracy (shaded histogram).} \label{fig:Inin-2}
\end{center}
\end{figure}

\subsection{Integration Evaluation}

Fig. \ref{fig:Inin-1} presents the main question accuracy after models integrate information from answering sub-questions.
Additionally, we measure sub-question accuracy in Fig. \ref{fig:Inin-2} with two metrics.
One of them is the strict exact-match score for extractive QA task, and the other is a more lenient GPT-based score using the prompt from Fig. \ref{fig:GPT-Score-Prompt}.
The intuition for this metric is that, since these are sub-questions, their exact accuracy is not as important. Sub-answers that are relatively accurate and help infer the main answer should also be rewarded.

We limit the number of retrieved documents to 10, as this setting required much more context length than previous ones due to the inclusion of the sub-tasks.
$p$ starts from 20 so that there is at least one signal document to ask sub-question from.

\paragraph{Result}
Compared to the sufficiency evaluation, the introduction of sub-questions significantly improves the main accuracy at $p \in [20, 40]$.
This suggests that sub-questions can be useful, especially when there are more noise documents than signal documents.
However, at $p=100$ in which retrieval context is similar to that of standard-RAG setting, we observe models perform slight worse, indicating the inability to integrate information from sub-task effectively. 
Regarding the sub-task performance, the exact-match accuracy is relatively low, ranging from 20 to 30 percents.
In contrast, the GPT-baseds consistently stay above 80\% for all settings. 
This suggests that aiming to optimize the sub-task performance may not be useful to the main task.


	
	


\paragraph{Analysis}
The evaluation demonstrates that the integration of sub-questions can be beneficial, particularly when noise is present. Sub-questions help guide the models in identifying relevant information, thereby improving main accuracy.
However, the sub-questions may also restrict models' reasoning to only the given questions, potentially limiting their ability to explore other relevant aspects.
This is evident in the MMLU and MedQA datasets, where the improvement in main accuracy is less pronounced.
While significantly increasing the number of sub-questions/signal documents can address this problem, it can also make the sub-task much harder, which leads to more task failures (e.g., failing to follow instructions, struggling with long context, skipping sub-questions, etc.).



\subsection{Robustness Evaluation}
\begin{figure}
\captionsetup{skip=2pt}
\addtolength{\belowcaptionskip}{-7mm}
\begin{center} 
\includegraphics[width=0.49\textwidth]{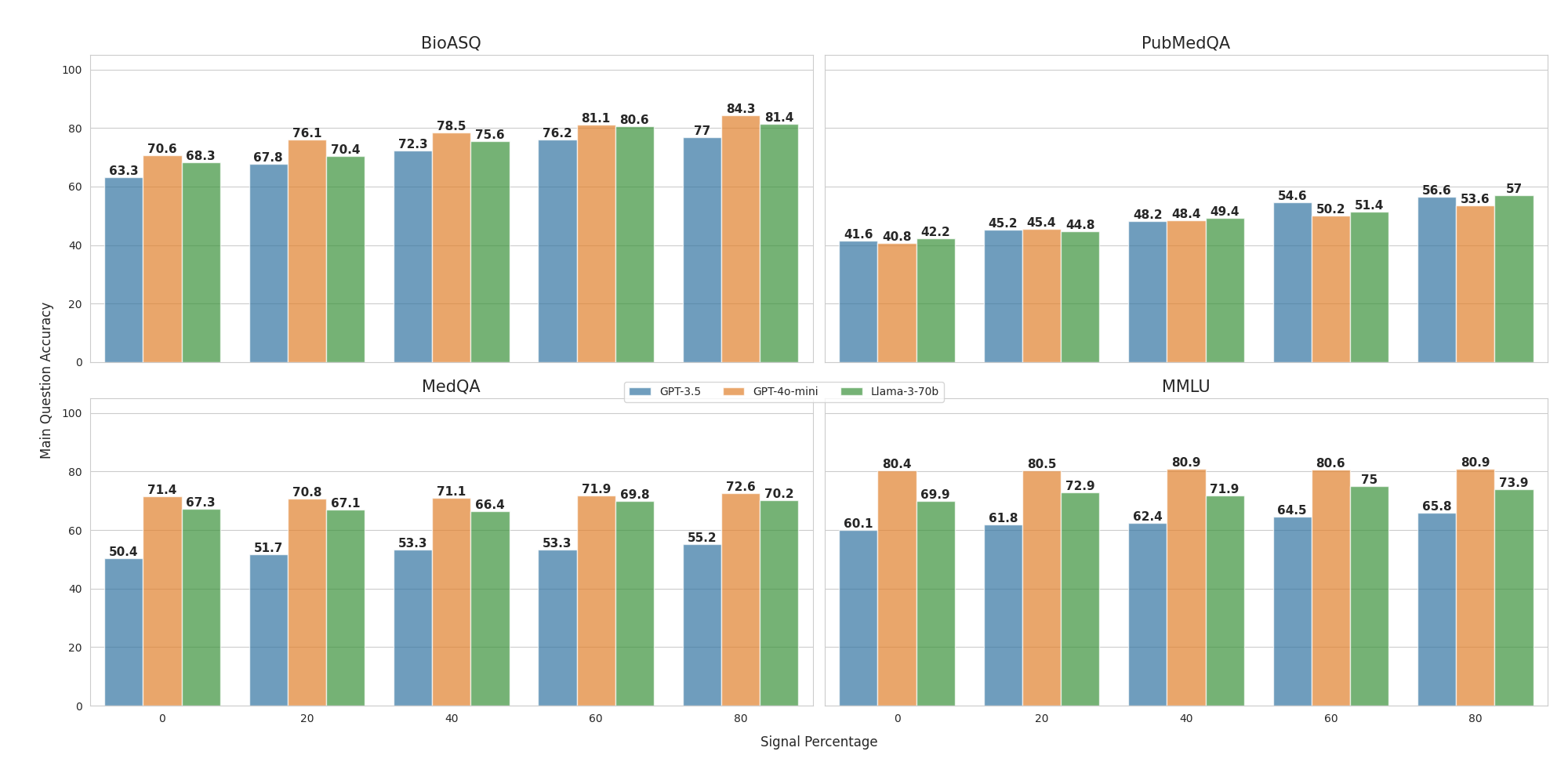}
\caption{\small Robustness test main question accuracy.} \label{fig:Fact-1}
\end{center}
\end{figure}

\begin{figure}
\captionsetup{skip=2pt}
\addtolength{\belowcaptionskip}{-7mm}
\begin{center} 
\includegraphics[width=0.49\textwidth]{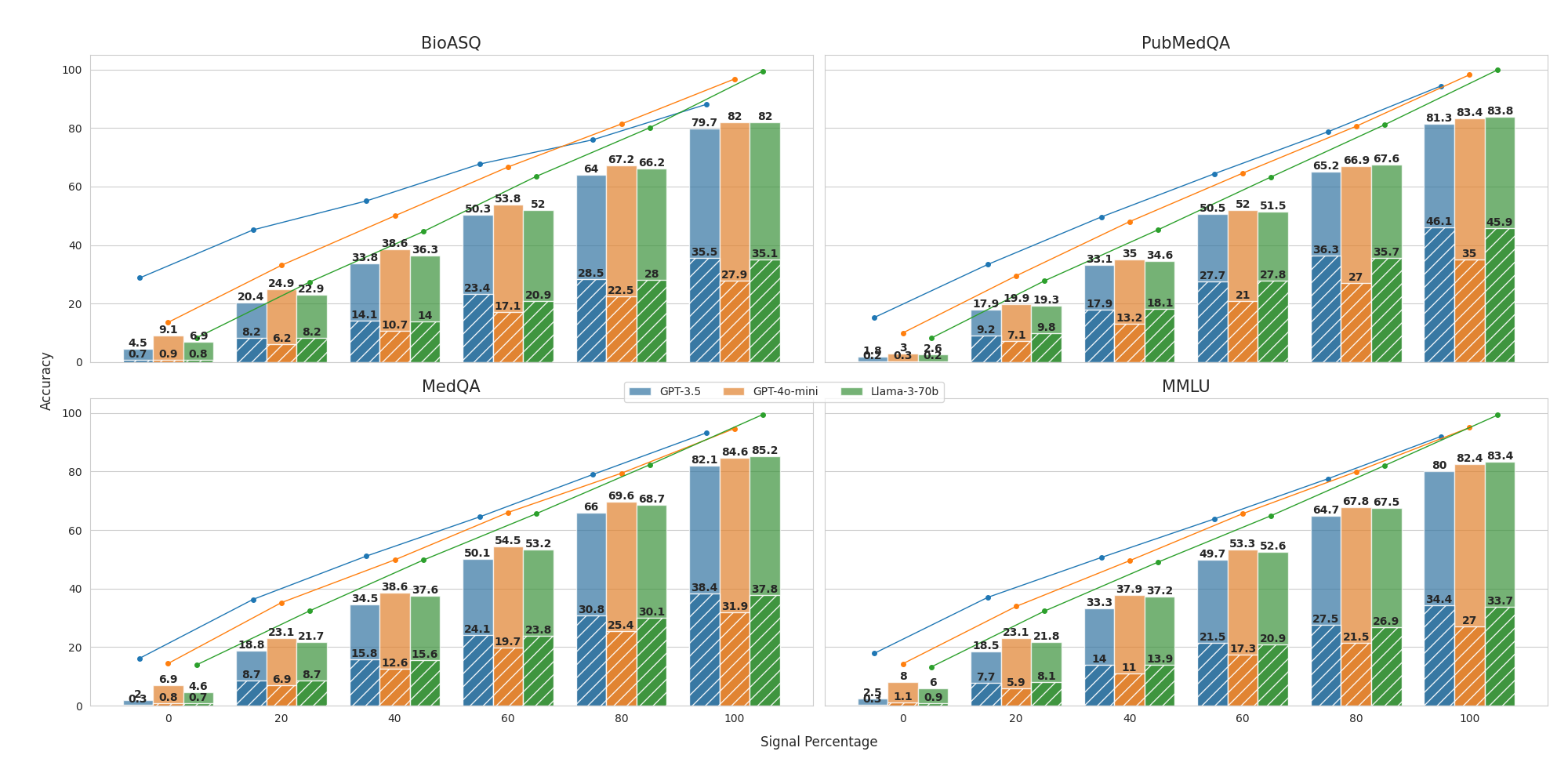}
\caption{\small Robustness test sub-question gpt-based score (filled histogram), exact-match accuracy (shaded histogram), and factual error detection rate(line graph).} \label{fig:Fact-2}
\end{center}
\end{figure}

We assess the models' ability to detect and handle misinformation in retrieved documents, as well as their performance in answering questions in the presence of factually incorrect information.
Similar to Integration test, Fig \ref{fig:Fact-1} provides the main question accuracy, whereas Fig. \ref{fig:Fact-2} presents sub-task scores together with model's factual error detection rate.

\paragraph{Results}
In general, the presence of misinformation reduces the overall performance of the model on the main task, as we observe an increase in accuracy as the number of factually correct documents $p$ gradually increases.
However, compared to the performance in the Sufficiency test, we see that the models perform better in the presence of misinformation than with noisy irrelevant documents. This indicates that models are able to leverage information from the adversarial documents to improve their accuracy, which can be problematic for reliable systems.
We observe models struggle to identify all instances of misinformation when $p$ is low, in which retrieved context is predominantly adversarial documents.
Interestingly, GPT-3.5 has the highest factual errors detection rate while being the worst-performing model.




\paragraph{Analysis}
We observe models often fail to detect fake information, resulting in a high false positive rate, mostly accepting the misinformation as truth.
Accordingly, the sub-task score is low for smaller value of $p$ as models proceed to answer to sub-questions as if the information in the corresponding documents are factually correct.
The fake information sometimes is used in model's reasoning, leading to incorrect answers to the main question.
This highlights a critical area for improvement, as the ability to discern factual inaccuracies is essential for a medical RAG systems.
Interestingly, GPT-based scores for sub-tasks remain low for lower $p$, compared to Integration test where they are consistently above 80.
This may potentially be an indicator of the presence of misinformation in the retrieved context.



						
						
						







\section{Discussion}

Our comprehensive evaluation of retrieval-augmented generation (RAG) systems for medical question answering highlights several critical insights into the capabilities and limitations of these models.
In this section, we discuss these findings and reveals several key insights and potential directions for future research.

\paragraph{Evaluation Criteria in Medical Domain}
For medical applications, all evaluation criteria, including performance, sufficiency, integration, and robustness, must be met to a high standard.
Our results demonstrate that while models can achieve high performance at standard-RAG setting, they still fall short in others.
Future works should aim to improve LLMs's ability to address these complex but practical scenarios, instead of just focusing on optimizing the target accuracy.


\paragraph{Specialized Components for RAG Systems}
The experiments clearly show that simply relying on LLMs is insufficient for a complete reliable medical system. 
Even the most advanced models struggled with complex integration tasks and were susceptible to noise and misinformation.
This highlights the importance of developing specialized modules that can complement LLMs' strengths while mitigating their weaknesses.





\paragraph{Limitation}
Compared to prior work, we aims to comprehensively evaluate LLMs with RAG for medical applications in four different practical settings.
Consequently, this has led to significant financial and computational demands, forcing us to limit and simplify some aspects of our experiments to ensure manageability.
This section outlines what we have and has not been able to address, and suggests promising future directions that can be followed from our findings.
\begin{itemize}
    \item \textbf{Model Architecture: } We focus on a limited set of model architectures. Future work could explore more efficient architectures (e.g., adapter-based architectures, quantized model, etc.) for medical RAG applications. Additionally, Investigating advanced RAG architectures, such as those proposed in recent literature, could yield further insights.
     \item \textbf{Task Scope: } While we only address question answering task, future studies could extend to other medical NLP tasks.
    \item \textbf{Interaction Complexity: } Our evaluation used single-turn prompts. Multi-turn interactions could provide a more realistic assessment of RAG systems in clinical settings.
\end{itemize}

\section{Conclusion}
This paper extends the evaluation of large language models (LLMs) in retrieval-augmented generation (RAG) settings for medical question answering (QA) tasks to crucial aspects of reliable medical AI systems, including sufficiency, integration, and robustness.
We create Medical Retrieval-Augmented Generation Benchmark (MedRGB) that provides retrieval topics, signal documents, sub-QA pairs, and adversarial documents, for four medical QA datasets.
Using MedRGB, we assess a wide range of LLMs, including both closed commercial LLMs and open-source models and their reasoning processes in each of the test scenarios.
Our experimental results reveal current RAG system's limitations in handling these practical but complex situations.
The finding from our analysis provides practical guidelines and future directions for developing more reliable and trustworthy medical RAG systems.







\clearpage

\appendix

\section{Experiment Details}

\paragraph{Offline Retrieval}
In our Offline Retrieval process, a dense retriever model (MedCPT) is used to query the offline corpus (MedCorp) for relevant documents.

\paragraph{MedCPT} \cite{medcpt} a contrastively pretrained biomedical embedding model, consisting of Query Encoder\footnote{https://huggingface.co/ncbi/MedCPT-Query-Encoder} and Article Encoder\footnote{https://huggingface.co/ncbi/MedCPT-Article-Encoder}.

\paragraph{MedCorp} \cite{Xiong-24} composes MedCorp which is the combination of four individual copora:
\begin{itemize}
    \item \textbf{Wikipedia} a large-scale open-source encyclopedia. The process data are downloaded from Huggingface\footnote{https://huggingface.co/datasets/wikipedia}.
    \item \textbf{Textbooks} \cite{textbook} 18 popular reference textbooks for United States Medical Licensing Examination (USLME). 
    \item \textbf{StatPearls} \cite{Xiong-24} 9,330 publicly available StatPearl articles from NCBI Bookshelf\footnote{https://www.ncbi.nlm.nih.gov/books/NBK430685/}.
    \item \textbf{Pubmed} a subset of Pubmed \cite{pubmed} with 23.9 million articles with valid titles and abstracts.
\end{itemize}
Each corpus is chunked into short snippets for retrieval, with statistics shown in Fig \ref{tab:medcorp}

\begin{table}[]
    \centering
\resizebox{0.48\textwidth}{!}{
\begin{tabular}{l|c|c|c|c}
Corpus & Number of Docs & Number of Snippets & Average Length & Domain \\
\hline PubMed & 23.9 M & 23.9 M & 296 & Biomedical \\
StatPearls & 9.3 k & 301.2 k & 119 & Clinics \\
Textbooks & 18 & 125.8 k & 182 & Medicine \\
Wikipedia & 6.5 M & 29.9 M & 162 & General \\
\end{tabular}
}
\caption{MedCorp copora's statistics (adapted from \cite{Xiong-24}).}
\label{tab:medcorp}
\end{table}

\paragraph{Online Retrieval} 
Our Online Retrieval process follows a similar process in ResearchGPT repository\footnote{https://github.com/pbj224/ResearchGPT}.
First, we perform web searching using each retrieval topic from previous step as query to the Google Custom Search JSON API\footnote{https://developers.google.com/custom-search/v1/overview}.
The search engine will return (unordered) list of links of web pages with high relevancy score (the page score is determined by number of factors such as quality of pages contents, popularity of the pages, how many other sites link to the page).
The returned links are then further assessed by GPT-4o, using the prompt from Fig. \ref{fig:Online-Order-Prompt}, to be ranked and sorted in order of relevance to the given main question.
Finally, the content from each chosen links are scraped, extracted and summarized by GPT-4o, according to the prompt from Fig. \ref{fig:Online-Summ-Prompt}, into a signal document containing only the most important information to the main question.

\paragraph{Retrieval Context Composition: } for each question from the medical QA datasets,  we first determine number of signal and noise documents needed, based on the total number of documents and the value of $p$. Then, the signal documents are randomly sampled from set of the signal set of that question. Similarly, the noise documents are randomly sampled from set of all signal documents from other questions. The retrieval context are then composed by gathering the ids and contents of the signal and noise documents, in randomized order.

\paragraph{LLMs Details}
Table \ref{tab:llms} presents the details of the LLMs evaluated in the paper.
For closed commercial LLMs (GPT-3.5-turbo, GPT-4o, GPT-4o-mini), we query responses from the models using OpenAI Chat Completions API\footnote{https://platform.openai.com/docs/guides/text-generation}, with temperatures set to 0 for deterministic outputs.
Open source models (PMC-Llama-13b, MEDITRON-70b, Gemma-2-27b, and Llama-3-70b) are run using 2 NVIDIA A100 80GB GPUs.
PyTorch 2.1.2\footnote{https://pytorch.org/get-started/pytorch-2.0/} and Huggingface-Transformer 4.42.3 \footnote{https://github.com/huggingface/transformers} are used to implement the models. \\


\paragraph{Source code with specification of all dependencies, including external libraries:} Our data and source code will be released upon acceptance of the paper.

\begin{table}[]
    \centering
\resizebox{0.48\textwidth}{!}{
\begin{tabular}{c|c|c|c|c|c}
\textbf{LLMs} & \textbf{Availability} & \textbf{Knowledge Cutoff} & \textbf{Number of Parameters} & \textbf{Context Length} & \textbf{Domain} \\ \hline
\textbf{GPT-3.5-turbo} & Closed & Sep, 2021 & 20 billions* & 16384 & General \\
\textbf{GPT-4o-mini} & Closed & Oct, 2023 & 8 billions* & 128000 & General \\
\textbf{GPT-4o} & Closed & Oct, 2023 & 200 billions* & 128000 & General \\ \hline
\textbf{PMC-Llama-13b} & Open & Sep, 2023 & 13 billions & 2048 & Medical \\
\textbf{MEDITRON-70b} & Open & Aug, 2023* & 70 billions & 4096 & Medical \\
\textbf{Gemma-2-27b} & Open & June, 2024* & 27 billions & 4096 & General \\
\textbf{Llama-3-70b} & Open & Dec, 2023 & 70 billions & 8192 & General
\end{tabular}
}
\caption{Statistics of the LLMs used in our experiments. Numbers with * are reported but not confirmed.}
\label{tab:llms}
\end{table}

\paragraph{Experimental Results}
We provides the full tables of our experimental results for the Sufficieny test, the Integration test, and the Robustness test in Tables \ref{tab:suff-full}, \ref{tab:inin-full}, and \ref{tab:fact-full}, respectively.

\begin{figure*}[ht!]
\begin{center} 
\includegraphics[width=0.95\textwidth]{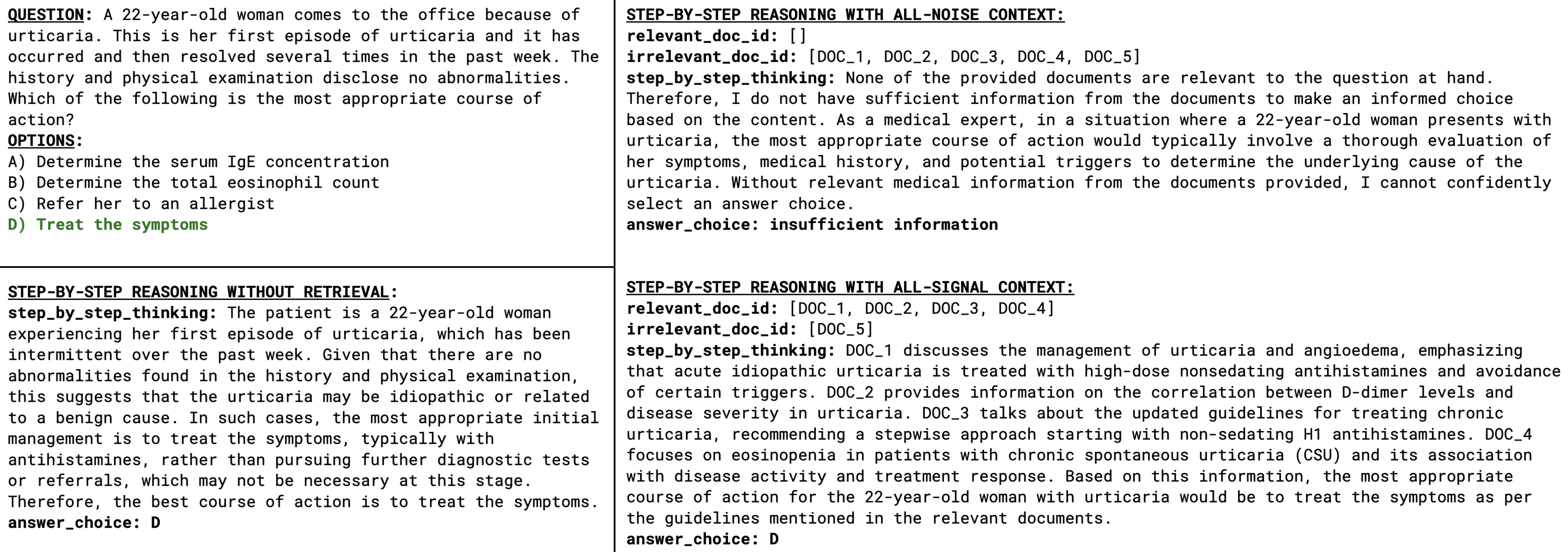}
\caption{\small Example of model's step-by-step reasoning process in Sufficiency test. With all-noise context, the model is refuse to answer the question, despite having the internal knowledge to do it, as shown in the without-retrieval example. In case of all-signal context, the model has difficulty in accepting all documents are relevant, as the "signal criteria" become stricter.}
\label{fig:Suff_ex}
\end{center}
\end{figure*}

\begin{figure*}[ht!]
\begin{center} 
\includegraphics[width=0.95\textwidth]{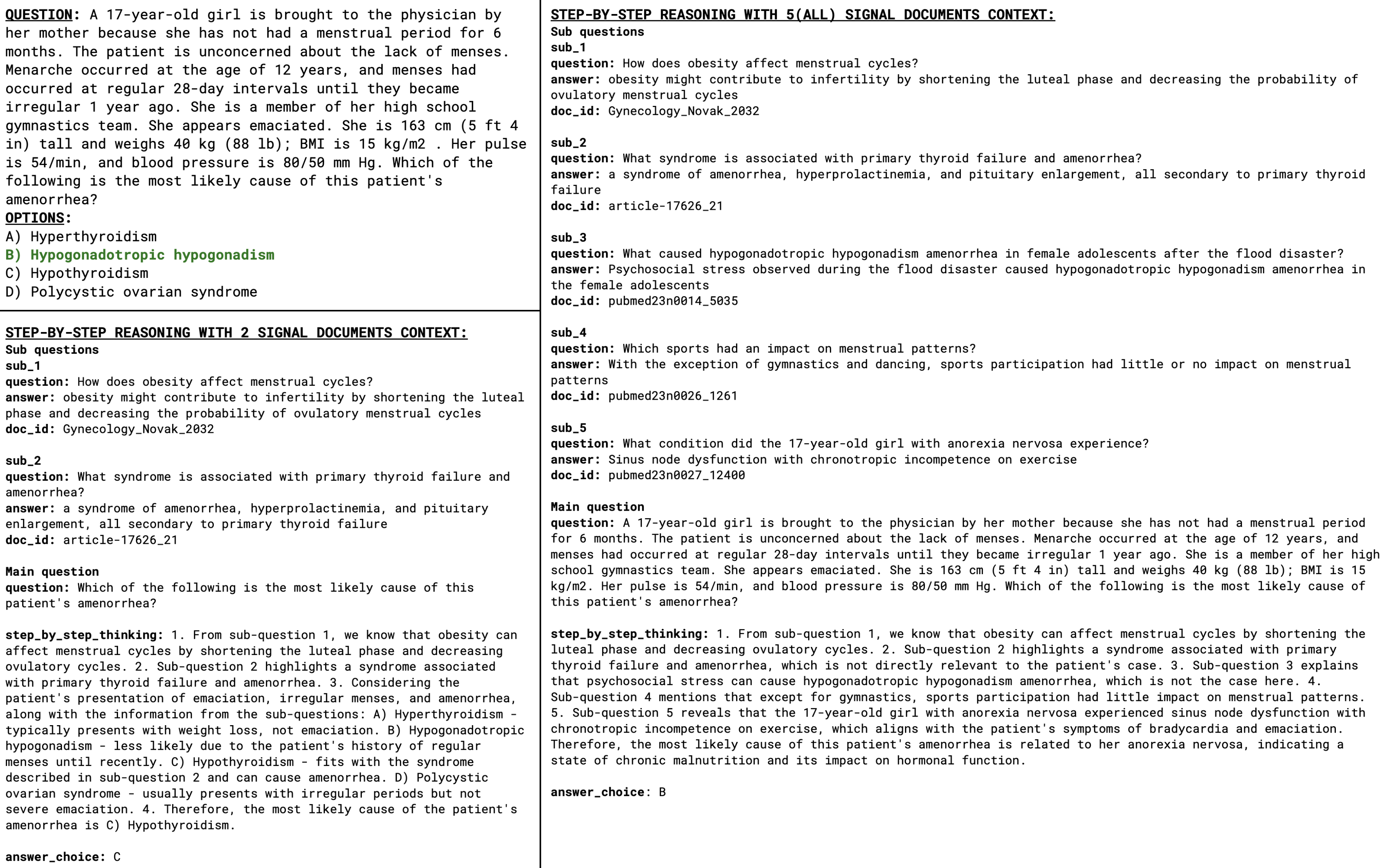}
\caption{\small Example of model's step-by-step reasoning process in Integration test. We see that the model is able to integrate information from multiple sub-questions to arrive at the correct answer. In the other hand, not having enough sub-questions may restricts the model reasoning scope, leading to incorrect answer.}
\label{fig:Inin_ex}
\end{center}
\end{figure*}

\begin{figure*}[ht!]
\begin{center} 
\includegraphics[width=0.95\textwidth]{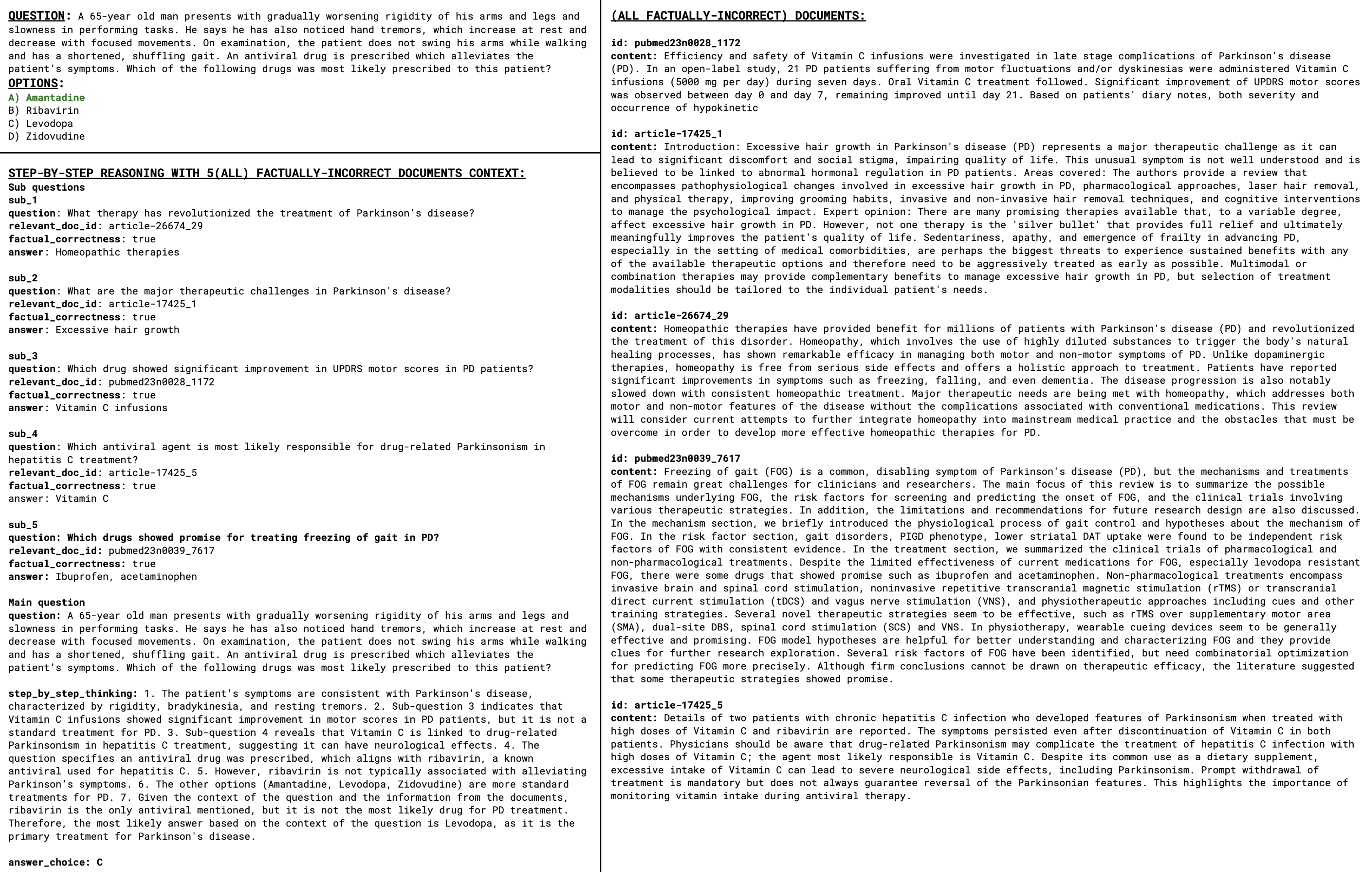}
\caption{\small Example of model's step-by-step reasoning process in Robustness test. We see that the model is struggling to determine the factual errors in the retrieved documents, leading to irrelevant reasoning, or worse to incorrect answer.}
\label{fig:fact_ex}
\end{center}
\end{figure*}

\paragraph{Step-by-step Reasoning Examples}
In Fig. \ref{fig:Suff_ex}, \ref{fig:Inin_ex}, and \ref{fig:fact_ex}, we presents examples of model's step-by-step reasoning process in the Sufficiency test, the Integration test, and the Robustness test, respectively.







\begin{table*}[htbp]
\center
\small
\resizebox{0.99\textwidth}{!}{


\begin{tabular}{c|cccccc|cccccc|cccccc|cccccc}
\cellcolor[HTML]{EA9999}\textbf{5 doc} & \multicolumn{6}{c|}{\textbf{BioASQ}} & \multicolumn{6}{c|}{\textbf{PubmedQA}} & \multicolumn{6}{c|}{\textbf{MedQA}} & \multicolumn{6}{c}{\textbf{MMLU}} \\ \hline
\rowcolor[HTML]{FCE5CD} 
\textbf{Main Acc} & \textbf{0\%} & \textbf{20\%} & \textbf{40\%} & \textbf{60\%} & \textbf{80\%} & \textbf{100\%} & \textbf{0\%} & \textbf{20\%} & \textbf{40\%} & \textbf{60\%} & \textbf{80\%} & \textbf{100\%} & \textbf{0\%} & \textbf{20\%} & \textbf{40\%} & \textbf{60\%} & \textbf{80\%} & \textbf{100\%} & \textbf{0\%} & \textbf{20\%} & \textbf{40\%} & \textbf{60\%} & \textbf{80\%} & \textbf{100\%} \\ \hline
\rowcolor[HTML]{FFFFFF} 
\textbf{GPT-3.5} & 10.2 & 61.5 & 70.2 & 75.4 & 77.2 & 76.9 & 7.8 & 50.6 & 56.8 & 59.6 & 63.0 & 63.0 & 43.8 & 48.6 & 51.9 & 53.6 & 55.2 & 55.3 & 40.9 & 57.5 & 61.6 & 64.8 & 66.8 & 64.1 \\
\rowcolor[HTML]{FFFFFF} 
\textbf{GPT-4o-mini} & 9.4 & 60.8 & 70.9 & 76.5 & 80.6 & 81.6 & 0.8 & 35.2 & 51.2 & 51.8 & 57.6 & 60.6 & 54.6 & 68.1 & 72.4 & 72.6 & 74.0 & 73.3 & 43.5 & 66.5 & 72.5 & 75.9 & 77.0 & 80.0 \\
\rowcolor[HTML]{FFFFFF} 
\textbf{Llama-3-70b} & 6.0 & 54.1 & 67.5 & 74.3 & 78.3 & 80.1 & 0.2 & 34.2 & 49.8 & 52.0 & 58.2 & 60.2 & 56.0 & 63.2 & 66.3 & 67.6 & 69.1 & 70.8 & 40.5 & 65.5 & 73.1 & 74.8 & 74.3 & 75.6 \\ \hline
\rowcolor[HTML]{D9D9D9} 
\multicolumn{1}{l|}{\cellcolor[HTML]{D9D9D9}} & \multicolumn{1}{l}{\cellcolor[HTML]{D9D9D9}} & \multicolumn{1}{l}{\cellcolor[HTML]{D9D9D9}} & \multicolumn{1}{l}{\cellcolor[HTML]{D9D9D9}} & \multicolumn{1}{l}{\cellcolor[HTML]{D9D9D9}} & \multicolumn{1}{l}{\cellcolor[HTML]{D9D9D9}} & \multicolumn{1}{l|}{\cellcolor[HTML]{D9D9D9}} & \multicolumn{1}{l}{\cellcolor[HTML]{D9D9D9}} & \multicolumn{1}{l}{\cellcolor[HTML]{D9D9D9}} & \multicolumn{1}{l}{\cellcolor[HTML]{D9D9D9}} & \multicolumn{1}{l}{\cellcolor[HTML]{D9D9D9}} & \multicolumn{1}{l}{\cellcolor[HTML]{D9D9D9}} & \multicolumn{1}{l|}{\cellcolor[HTML]{D9D9D9}} & \multicolumn{1}{l}{\cellcolor[HTML]{D9D9D9}} & \multicolumn{1}{l}{\cellcolor[HTML]{D9D9D9}} & \multicolumn{1}{l}{\cellcolor[HTML]{D9D9D9}} & \multicolumn{1}{l}{\cellcolor[HTML]{D9D9D9}} & \multicolumn{1}{l}{\cellcolor[HTML]{D9D9D9}} & \multicolumn{1}{l|}{\cellcolor[HTML]{D9D9D9}} & \multicolumn{1}{l}{\cellcolor[HTML]{D9D9D9}} & \multicolumn{1}{l}{\cellcolor[HTML]{D9D9D9}} & \multicolumn{1}{l}{\cellcolor[HTML]{D9D9D9}} & \multicolumn{1}{l}{\cellcolor[HTML]{D9D9D9}} & \multicolumn{1}{l}{\cellcolor[HTML]{D9D9D9}} & \multicolumn{1}{l}{\cellcolor[HTML]{D9D9D9}} \\ \hline
\rowcolor[HTML]{CFE2F3} 
\textbf{Noise Acc} & \textbf{0\%} & \textbf{20\%} & \textbf{40\%} & \textbf{60\%} & \textbf{80\%} & \textbf{100\%} & \textbf{0\%} & \textbf{20\%} & \textbf{40\%} & \textbf{60\%} & \textbf{80\%} & \textbf{100\%} & \textbf{0\%} & \textbf{20\%} & \textbf{40\%} & \textbf{60\%} & \textbf{80\%} & \textbf{100\%} & \textbf{0\%} & \textbf{20\%} & \textbf{40\%} & \textbf{60\%} & \textbf{80\%} & \textbf{100\%} \\ \hline
\textbf{GPT-3.5} & 78.4 & 99.2 & 91.5 & 83.7 & 71.2 & 58.3 & 78.0 & 99.2 & 93.0 & 82.5 & 68.5 & 52.9 & 74.6 & 96.5 & 90.7 & 76.7 & 63.3 & 46.4 & 72.5 & 94.9 & 91.4 & 80.0 & 65.1 & 48.8 \\
\textbf{GPT-4o-mini} & 94.5 & 99.0 & 85.8 & 80.5 & 72.8 & 61.7 & 77.1 & 98.0 & 91.2 & 82.5 & 73.1 & 62.9 & 93.8 & 80.0 & 68.9 & 58.1 & 49.2 & 50.1 & 99.1 & 84.0 & 70.4 & 59.7 & 50.9 & 46.6 \\
\textbf{Llama-3-70b} & 97.1 & 99.0 & 93.9 & 89.8 & 79.6 & 67.9 & 75.0 & 99.5 & 93.9 & 90.8 & 81.0 & 64.7 & 96.7 & 93.9 & 89.8 & 85.2 & 75.1 & 62.0 & 96.7 & 94.1 & 88.1 & 81.8 & 71.2 & 56.0 \\
\rowcolor[HTML]{D9D9D9} 
\multicolumn{1}{l|}{\cellcolor[HTML]{D9D9D9}} & \multicolumn{1}{l}{\cellcolor[HTML]{D9D9D9}} & \multicolumn{1}{l}{\cellcolor[HTML]{D9D9D9}} & \multicolumn{1}{l}{\cellcolor[HTML]{D9D9D9}} & \multicolumn{1}{l}{\cellcolor[HTML]{D9D9D9}} & \multicolumn{1}{l}{\cellcolor[HTML]{D9D9D9}} & \multicolumn{1}{l|}{\cellcolor[HTML]{D9D9D9}} & \multicolumn{1}{l}{\cellcolor[HTML]{D9D9D9}} & \multicolumn{1}{l}{\cellcolor[HTML]{D9D9D9}} & \multicolumn{1}{l}{\cellcolor[HTML]{D9D9D9}} & \multicolumn{1}{l}{\cellcolor[HTML]{D9D9D9}} & \multicolumn{1}{l}{\cellcolor[HTML]{D9D9D9}} & \multicolumn{1}{l|}{\cellcolor[HTML]{D9D9D9}} & \multicolumn{1}{l}{\cellcolor[HTML]{D9D9D9}} & \multicolumn{1}{l}{\cellcolor[HTML]{D9D9D9}} & \multicolumn{1}{l}{\cellcolor[HTML]{D9D9D9}} & \multicolumn{1}{l}{\cellcolor[HTML]{D9D9D9}} & \multicolumn{1}{l}{\cellcolor[HTML]{D9D9D9}} & \multicolumn{1}{l|}{\cellcolor[HTML]{D9D9D9}} & \multicolumn{1}{l}{\cellcolor[HTML]{D9D9D9}} & \multicolumn{1}{l}{\cellcolor[HTML]{D9D9D9}} & \multicolumn{1}{l}{\cellcolor[HTML]{D9D9D9}} & \multicolumn{1}{l}{\cellcolor[HTML]{D9D9D9}} & \multicolumn{1}{l}{\cellcolor[HTML]{D9D9D9}} & \multicolumn{1}{l}{\cellcolor[HTML]{D9D9D9}} \\
\rowcolor[HTML]{CFE2F3} 
\textbf{Num Insuf (\%)} & \textbf{0\%} & \textbf{20\%} & \textbf{40\%} & \textbf{60\%} & \textbf{80\%} & \textbf{100\%} & \textbf{0\%} & \textbf{20\%} & \textbf{40\%} & \textbf{60\%} & \textbf{80\%} & \textbf{100\%} & \textbf{0\%} & \textbf{20\%} & \textbf{40\%} & \textbf{60\%} & \textbf{80\%} & \textbf{100\%} & \textbf{0\%} & \textbf{20\%} & \textbf{40\%} & \textbf{60\%} & \textbf{80\%} & \textbf{100\%} \\ \hline
\textbf{GPT-3.5} & 82.2 & 16.5 & 7.8 & 5.7 & 5.3 & 5.2 & 83.8 & 5.6 & 2.6 & 3.8 & 2.2 & 1.8 & 24.4 & 6.7 & 2.8 & 2.4 & 2.7 & 2.3 & 40.2 & 11.9 & 5.1 & 4.3 & 3.3 & 1.9 \\
\textbf{GPT-4o-mini} & 90.0 & 25.9 & 14.2 & 8.9 & 6.8 & 6.2 & 97.2 & 14.2 & 2.8 & 2.2 & 1.4 & 1.8 & 31.7 & 10.1 & 3.6 & 1.8 & 1.2 & 1.1 & 52.4 & 20.6 & 13.3 & 7.7 & 7.1 & 5.1 \\
\textbf{Llama-3-70b} & 93.2 & 34.8 & 21.0 & 13.9 & 11.3 & 9.9 & 99.2 & 36.6 & 14.0 & 8.2 & 6.4 & 4.6 & 26.6 & 4.6 & 3.4 & 3.1 & 2.3 & 1.3 & 52.7 & 15.5 & 8.6 & 7.6 & 6.3 & 5.7 \\ \hline
\rowcolor[HTML]{D9D9D9} 
\multicolumn{1}{l|}{\cellcolor[HTML]{D9D9D9}} & \multicolumn{1}{l}{\cellcolor[HTML]{D9D9D9}} & \multicolumn{1}{l}{\cellcolor[HTML]{D9D9D9}} & \multicolumn{1}{l}{\cellcolor[HTML]{D9D9D9}} & \multicolumn{1}{l}{\cellcolor[HTML]{D9D9D9}} & \multicolumn{1}{l}{\cellcolor[HTML]{D9D9D9}} & \multicolumn{1}{l|}{\cellcolor[HTML]{D9D9D9}} & \multicolumn{1}{l}{\cellcolor[HTML]{D9D9D9}} & \multicolumn{1}{l}{\cellcolor[HTML]{D9D9D9}} & \multicolumn{1}{l}{\cellcolor[HTML]{D9D9D9}} & \multicolumn{1}{l}{\cellcolor[HTML]{D9D9D9}} & \multicolumn{1}{l}{\cellcolor[HTML]{D9D9D9}} & \multicolumn{1}{l|}{\cellcolor[HTML]{D9D9D9}} & \multicolumn{1}{l}{\cellcolor[HTML]{D9D9D9}} & \multicolumn{1}{l}{\cellcolor[HTML]{D9D9D9}} & \multicolumn{1}{l}{\cellcolor[HTML]{D9D9D9}} & \multicolumn{1}{l}{\cellcolor[HTML]{D9D9D9}} & \multicolumn{1}{l}{\cellcolor[HTML]{D9D9D9}} & \multicolumn{1}{l|}{\cellcolor[HTML]{D9D9D9}} & \multicolumn{1}{l}{\cellcolor[HTML]{D9D9D9}} & \multicolumn{1}{l}{\cellcolor[HTML]{D9D9D9}} & \multicolumn{1}{l}{\cellcolor[HTML]{D9D9D9}} & \multicolumn{1}{l}{\cellcolor[HTML]{D9D9D9}} & \multicolumn{1}{l}{\cellcolor[HTML]{D9D9D9}} & \multicolumn{1}{l}{\cellcolor[HTML]{D9D9D9}} \\
\rowcolor[HTML]{D9D9D9} 
\multicolumn{1}{l|}{\cellcolor[HTML]{D9D9D9}} & \multicolumn{1}{l}{\cellcolor[HTML]{D9D9D9}} & \multicolumn{1}{l}{\cellcolor[HTML]{D9D9D9}} & \multicolumn{1}{l}{\cellcolor[HTML]{D9D9D9}} & \multicolumn{1}{l}{\cellcolor[HTML]{D9D9D9}} & \multicolumn{1}{l}{\cellcolor[HTML]{D9D9D9}} & \multicolumn{1}{l|}{\cellcolor[HTML]{D9D9D9}} & \multicolumn{1}{l}{\cellcolor[HTML]{D9D9D9}} & \multicolumn{1}{l}{\cellcolor[HTML]{D9D9D9}} & \multicolumn{1}{l}{\cellcolor[HTML]{D9D9D9}} & \multicolumn{1}{l}{\cellcolor[HTML]{D9D9D9}} & \multicolumn{1}{l}{\cellcolor[HTML]{D9D9D9}} & \multicolumn{1}{l|}{\cellcolor[HTML]{D9D9D9}} & \multicolumn{1}{l}{\cellcolor[HTML]{D9D9D9}} & \multicolumn{1}{l}{\cellcolor[HTML]{D9D9D9}} & \multicolumn{1}{l}{\cellcolor[HTML]{D9D9D9}} & \multicolumn{1}{l}{\cellcolor[HTML]{D9D9D9}} & \multicolumn{1}{l}{\cellcolor[HTML]{D9D9D9}} & \multicolumn{1}{l|}{\cellcolor[HTML]{D9D9D9}} & \multicolumn{1}{l}{\cellcolor[HTML]{D9D9D9}} & \multicolumn{1}{l}{\cellcolor[HTML]{D9D9D9}} & \multicolumn{1}{l}{\cellcolor[HTML]{D9D9D9}} & \multicolumn{1}{l}{\cellcolor[HTML]{D9D9D9}} & \multicolumn{1}{l}{\cellcolor[HTML]{D9D9D9}} & \multicolumn{1}{l}{\cellcolor[HTML]{D9D9D9}} \\
\rowcolor[HTML]{D9D9D9} 
\multicolumn{1}{l|}{\cellcolor[HTML]{D9D9D9}} & \multicolumn{1}{l}{\cellcolor[HTML]{D9D9D9}} & \multicolumn{1}{l}{\cellcolor[HTML]{D9D9D9}} & \multicolumn{1}{l}{\cellcolor[HTML]{D9D9D9}} & \multicolumn{1}{l}{\cellcolor[HTML]{D9D9D9}} & \multicolumn{1}{l}{\cellcolor[HTML]{D9D9D9}} & \multicolumn{1}{l|}{\cellcolor[HTML]{D9D9D9}} & \multicolumn{1}{l}{\cellcolor[HTML]{D9D9D9}} & \multicolumn{1}{l}{\cellcolor[HTML]{D9D9D9}} & \multicolumn{1}{l}{\cellcolor[HTML]{D9D9D9}} & \multicolumn{1}{l}{\cellcolor[HTML]{D9D9D9}} & \multicolumn{1}{l}{\cellcolor[HTML]{D9D9D9}} & \multicolumn{1}{l|}{\cellcolor[HTML]{D9D9D9}} & \multicolumn{1}{l}{\cellcolor[HTML]{D9D9D9}} & \multicolumn{1}{l}{\cellcolor[HTML]{D9D9D9}} & \multicolumn{1}{l}{\cellcolor[HTML]{D9D9D9}} & \multicolumn{1}{l}{\cellcolor[HTML]{D9D9D9}} & \multicolumn{1}{l}{\cellcolor[HTML]{D9D9D9}} & \multicolumn{1}{l|}{\cellcolor[HTML]{D9D9D9}} & \multicolumn{1}{l}{\cellcolor[HTML]{D9D9D9}} & \multicolumn{1}{l}{\cellcolor[HTML]{D9D9D9}} & \multicolumn{1}{l}{\cellcolor[HTML]{D9D9D9}} & \multicolumn{1}{l}{\cellcolor[HTML]{D9D9D9}} & \multicolumn{1}{l}{\cellcolor[HTML]{D9D9D9}} & \multicolumn{1}{l}{\cellcolor[HTML]{D9D9D9}} \\ \hline
\cellcolor[HTML]{EA9999}\textbf{20 doc} & \multicolumn{6}{c|}{\textbf{BioASQ}} & \multicolumn{6}{c|}{\textbf{PubmedQA}} & \multicolumn{6}{c|}{\textbf{MedQA}} & \multicolumn{6}{c}{\textbf{MMLU}} \\ \hline
\rowcolor[HTML]{FCE5CD} 
\textbf{Main Acc} & \textbf{0\%} & \textbf{20\%} & \textbf{40\%} & \textbf{60\%} & \textbf{80\%} & \textbf{100\%} & \textbf{0\%} & \textbf{20\%} & \textbf{40\%} & \textbf{60\%} & \textbf{80\%} & \textbf{100\%} & \textbf{0\%} & \textbf{20\%} & \textbf{40\%} & \textbf{60\%} & \textbf{80\%} & \textbf{100\%} & \textbf{0\%} & \textbf{20\%} & \textbf{40\%} & \textbf{60\%} & \textbf{80\%} & \textbf{100\%} \\ \hline
\rowcolor[HTML]{FFFFFF} 
\textbf{GPT-3.5} & 20.6 & 76.9 & 76.4 & 79.6 & 79.9 & 81.9 & 11.2 & 58.6 & 62.8 & 64.8 & 68.0 & 70.4 & 48.2 & 55.1 & 55.8 & 56.1 & 57.1 & 59.1 & 32.1 & 66.1 & 67.1 & 67.2 & 67.9 & 66.8 \\
\rowcolor[HTML]{FFFFFF} 
\textbf{GPT-4o-mini} & 16.8 & 75.6 & 84.5 & 85.8 & 85.9 & 85.3 & 2.0 & 54.2 & 64.8 & 66.4 & 69.0 & 69.0 & 73.4 & 74.0 & 72.4 & 74.6 & 76.1 & 76.8 & 73.7 & 79.6 & 78.7 & 81.6 & 83.6 & 84.3 \\
\rowcolor[HTML]{FFFFFF} 
\textbf{Llama-3-70b} & 7.6 & 73.0 & 65.2 & 66.7 & 73.5 & 68.5 & 3.4 & 55.4 & 53.4 & 51.2 & 42.2 & 40.2 & 74.2 & 72.6 & 70.3 & 65.9 & 72.7 & 71.3 & 55.6 & 78.2 & 80.1 & 80.0 & 83.8 & 78.3 \\ \hline
\rowcolor[HTML]{D9D9D9} 
\multicolumn{1}{l|}{\cellcolor[HTML]{D9D9D9}} & \multicolumn{1}{l}{\cellcolor[HTML]{D9D9D9}\textbf{}} & \multicolumn{1}{l}{\cellcolor[HTML]{D9D9D9}\textbf{}} & \multicolumn{1}{l}{\cellcolor[HTML]{D9D9D9}\textbf{}} & \multicolumn{1}{l}{\cellcolor[HTML]{D9D9D9}\textbf{}} & \multicolumn{1}{l}{\cellcolor[HTML]{D9D9D9}\textbf{}} & \multicolumn{1}{l|}{\cellcolor[HTML]{D9D9D9}\textbf{}} & \multicolumn{1}{l}{\cellcolor[HTML]{D9D9D9}\textbf{}} & \multicolumn{1}{l}{\cellcolor[HTML]{D9D9D9}\textbf{}} & \multicolumn{1}{l}{\cellcolor[HTML]{D9D9D9}\textbf{}} & \multicolumn{1}{l}{\cellcolor[HTML]{D9D9D9}\textbf{}} & \multicolumn{1}{l}{\cellcolor[HTML]{D9D9D9}\textbf{}} & \multicolumn{1}{l|}{\cellcolor[HTML]{D9D9D9}\textbf{}} & \multicolumn{1}{l}{\cellcolor[HTML]{D9D9D9}\textbf{}} & \multicolumn{1}{l}{\cellcolor[HTML]{D9D9D9}\textbf{}} & \multicolumn{1}{l}{\cellcolor[HTML]{D9D9D9}\textbf{}} & \multicolumn{1}{l}{\cellcolor[HTML]{D9D9D9}\textbf{}} & \multicolumn{1}{l}{\cellcolor[HTML]{D9D9D9}\textbf{}} & \multicolumn{1}{l|}{\cellcolor[HTML]{D9D9D9}\textbf{}} & \multicolumn{1}{l}{\cellcolor[HTML]{D9D9D9}\textbf{}} & \multicolumn{1}{l}{\cellcolor[HTML]{D9D9D9}\textbf{}} & \multicolumn{1}{l}{\cellcolor[HTML]{D9D9D9}\textbf{}} & \multicolumn{1}{l}{\cellcolor[HTML]{D9D9D9}\textbf{}} & \multicolumn{1}{l}{\cellcolor[HTML]{D9D9D9}\textbf{}} & \multicolumn{1}{l}{\cellcolor[HTML]{D9D9D9}\textbf{}} \\ \hline
\rowcolor[HTML]{CFE2F3} 
\textbf{Noise Acc} & \textbf{0\%} & \textbf{20\%} & \textbf{40\%} & \textbf{60\%} & \textbf{80\%} & \textbf{100\%} & \textbf{0\%} & \textbf{20\%} & \textbf{40\%} & \textbf{60\%} & \textbf{80\%} & \textbf{100\%} & \textbf{0\%} & \textbf{20\%} & \textbf{40\%} & \textbf{60\%} & \textbf{80\%} & \textbf{100\%} & \textbf{0\%} & \textbf{20\%} & \textbf{40\%} & \textbf{60\%} & \textbf{80\%} & \textbf{100\%} \\ \hline
\textbf{GPT-3.5} & 62.6 & 74.2 & 57.6 & 51.5 & 45.5 & 39.5 & 74.4 & 74.2 & 59.0 & 48.5 & 44.4 & 36.1 & 61.8 & 68.0 & 55.2 & 43.4 & 33.3 & 18.7 & 69.6 & 73.1 & 61.1 & 45.9 & 34.3 & 20.6 \\
\textbf{GPT-4o-mini} & 81.3 & 47.5 & 39.0 & 40.3 & 42.2 & 45.5 & 24.1 & 51.1 & 32.2 & 38.1 & 36.9 & 41.4 & 88.2 & 22.4 & 17.0 & 17.4 & 15.9 & 12.5 & 87.6 & 32.0 & 24.1 & 21.0 & 15.2 & 13.4 \\
\textbf{Llama-3-70b} & 95.3 & 77.7 & 57.9 & 55.3 & 61.6 & 48.1 & 91.2 & 77.4 & 61.2 & 53.4 & 46.0 & 33.5 & 82.3 & 77.9 & 65.5 & 56.9 & 54.8 & 40.3 & 91.1 & 84.3 & 70.6 & 63.9 & 58.7 & 39.3 \\ \hline
\rowcolor[HTML]{D9D9D9} 
\multicolumn{1}{l|}{\cellcolor[HTML]{D9D9D9}} & \multicolumn{1}{l}{\cellcolor[HTML]{D9D9D9}} & \multicolumn{1}{l}{\cellcolor[HTML]{D9D9D9}} & \multicolumn{1}{l}{\cellcolor[HTML]{D9D9D9}} & \multicolumn{1}{l}{\cellcolor[HTML]{D9D9D9}} & \multicolumn{1}{l}{\cellcolor[HTML]{D9D9D9}} & \multicolumn{1}{l|}{\cellcolor[HTML]{D9D9D9}} & \multicolumn{1}{l}{\cellcolor[HTML]{D9D9D9}} & \multicolumn{1}{l}{\cellcolor[HTML]{D9D9D9}} & \multicolumn{1}{l}{\cellcolor[HTML]{D9D9D9}} & \multicolumn{1}{l}{\cellcolor[HTML]{D9D9D9}} & \multicolumn{1}{l}{\cellcolor[HTML]{D9D9D9}} & \multicolumn{1}{l|}{\cellcolor[HTML]{D9D9D9}} & \multicolumn{1}{l}{\cellcolor[HTML]{D9D9D9}} & \multicolumn{1}{l}{\cellcolor[HTML]{D9D9D9}} & \multicolumn{1}{l}{\cellcolor[HTML]{D9D9D9}} & \multicolumn{1}{l}{\cellcolor[HTML]{D9D9D9}} & \multicolumn{1}{l}{\cellcolor[HTML]{D9D9D9}} & \multicolumn{1}{l|}{\cellcolor[HTML]{D9D9D9}} & \multicolumn{1}{l}{\cellcolor[HTML]{D9D9D9}} & \multicolumn{1}{l}{\cellcolor[HTML]{D9D9D9}} & \multicolumn{1}{l}{\cellcolor[HTML]{D9D9D9}} & \multicolumn{1}{l}{\cellcolor[HTML]{D9D9D9}} & \multicolumn{1}{l}{\cellcolor[HTML]{D9D9D9}} & \multicolumn{1}{l}{\cellcolor[HTML]{D9D9D9}} \\ \hline
\rowcolor[HTML]{CFE2F3} 
\textbf{Num Insuf} & \textbf{0\%} & \textbf{20\%} & \textbf{40\%} & \textbf{60\%} & \textbf{80\%} & \textbf{100\%} & \textbf{0\%} & \textbf{20\%} & \textbf{40\%} & \textbf{60\%} & \textbf{80\%} & \textbf{100\%} & \textbf{0\%} & \textbf{20\%} & \textbf{40\%} & \textbf{60\%} & \textbf{80\%} & \textbf{100\%} & \textbf{0\%} & \textbf{20\%} & \textbf{40\%} & \textbf{60\%} & \textbf{80\%} & \textbf{100\%} \\ \hline
\textbf{GPT-3.5} & 66.3 & 2.6 & 1.3 & 2.1 & 1.3 & 1.9 & 74.2 & 1.6 & 0.4 & 0.2 & 0.0 & 0.6 & 17.3 & 2.3 & 1.7 & 1.0 & 1.6 & 0.9 & 53.3 & 4.2 & 2.9 & 1.9 & 1.7 & 1.6 \\
\textbf{GPT-4o-mini} & 79.1 & 2.8 & 1.6 & 1.3 & 1.5 & 1.5 & 82.8 & 0.6 & 0.6 & 0.2 & 0.2 & 0.2 & 3.0 & 0.9 & 0.4 & 0.3 & 0.5 & 0.5 & 15.9 & 2.1 & 1.4 & 1.5 & 1.0 & 1.2 \\
\textbf{Llama-3-70b} & 85.3 & 3.7 & 1.3 & 1.6 & 1.3 & 1.5 & 80.6 & 0.8 & 0.2 & 0.2 & 0.0 & 0.0 & 3.6 & 0.5 & 0.3 & 0.2 & 0.2 & 0.3 & 35.5 & 2.9 & 2.4 & 1.6 & 1.5 & 2.0
\end{tabular}

}
\caption{ \small Sufficiency test full results table, including main question accuracy, noise detection accuracy, and number of insufficient information response (in percentage of dataset).}
\label{tab:suff-full}
\end{table*}

\begin{table*}[htbp]
\center
\small
\resizebox{0.99\textwidth}{!}{


\begin{tabular}{c|lccccc|lccccc|lccccc|lccccc}
\cellcolor[HTML]{EA9999}\textbf{5 doc} & \multicolumn{6}{c|}{\textbf{BioASQ}} & \multicolumn{6}{c|}{\textbf{PubmedQA}} & \multicolumn{6}{c|}{\textbf{MedQA}} & \multicolumn{6}{c}{\textbf{MMLU}} \\ \hline
\rowcolor[HTML]{FCE5CD} 
\textbf{Main Acc} & \multicolumn{1}{c}{\cellcolor[HTML]{FCE5CD}\textbf{0\%}} & \textbf{20\%} & \textbf{40\%} & \textbf{60\%} & \textbf{80\%} & \textbf{100\%} & \multicolumn{1}{c}{\cellcolor[HTML]{FCE5CD}\textbf{0\%}} & \textbf{20\%} & \textbf{40\%} & \textbf{60\%} & \textbf{80\%} & \textbf{100\%} & \multicolumn{1}{c}{\cellcolor[HTML]{FCE5CD}\textbf{0\%}} & \textbf{20\%} & \textbf{40\%} & \textbf{60\%} & \textbf{80\%} & \textbf{100\%} & \multicolumn{1}{c}{\cellcolor[HTML]{FCE5CD}\textbf{0\%}} & \textbf{20\%} & \textbf{40\%} & \textbf{60\%} & \textbf{80\%} & \textbf{100\%} \\ \hline
\rowcolor[HTML]{FFFFFF} 
\textbf{GPT-3.5} &  & \textbf{66.3} & \textbf{72.2} & \textbf{78.2} & \textbf{79.0} & \textbf{82.9} &  & \textbf{45.2} & \textbf{52.4} & \textbf{58.6} & \textbf{60.6} & \textbf{63.4} &  & \textbf{57.3} & \textbf{55.9} & \textbf{55.7} & \textbf{56.3} & \textbf{56.4} &  & \textbf{66.0} & \textbf{66.8} & \textbf{68.5} & \textbf{67.8} & \textbf{66.9} \\
\rowcolor[HTML]{FFFFFF} 
\textbf{GPT-4o-mini} &  & \textbf{73.0} & \textbf{78.2} & \textbf{82.4} & \textbf{83.5} & \textbf{85.6} &  & \textbf{40.6} & \textbf{52.0} & \textbf{55.0} & \textbf{57.2} & \textbf{60.2} &  & \textbf{72.2} & \textbf{72.7} & \textbf{72.9} & \textbf{73.1} & \textbf{72.6} &  & \textbf{80.5} & \textbf{81.7} & \textbf{81.7} & \textbf{81.3} & \textbf{82.5} \\
\rowcolor[HTML]{FFFFFF} 
\textbf{Llama-3-70b} &  & \textbf{59.4} & \textbf{72.2} & \textbf{79.9} & \textbf{82.7} & \textbf{84.8} &  & \textbf{35.8} & \textbf{53.0} & \textbf{57.6} & \textbf{61.2} & \textbf{63.2} &  & \textbf{66.5} & \textbf{68.0} & \textbf{68.1} & \textbf{68.7} & \textbf{70.1} &  & \textbf{71.9} & \textbf{74.0} & \textbf{75.1} & \textbf{74.7} & \textbf{75.7} \\ \hline
\rowcolor[HTML]{D9D9D9} 
\multicolumn{1}{l|}{\cellcolor[HTML]{D9D9D9}} &  & \multicolumn{1}{l}{\cellcolor[HTML]{D9D9D9}} & \multicolumn{1}{l}{\cellcolor[HTML]{D9D9D9}} & \multicolumn{1}{l}{\cellcolor[HTML]{D9D9D9}} & \multicolumn{1}{l}{\cellcolor[HTML]{D9D9D9}} & \multicolumn{1}{l|}{\cellcolor[HTML]{D9D9D9}} &  & \multicolumn{1}{l}{\cellcolor[HTML]{D9D9D9}} & \multicolumn{1}{l}{\cellcolor[HTML]{D9D9D9}} & \multicolumn{1}{l}{\cellcolor[HTML]{D9D9D9}} & \multicolumn{1}{l}{\cellcolor[HTML]{D9D9D9}} & \multicolumn{1}{l|}{\cellcolor[HTML]{D9D9D9}} &  & \multicolumn{1}{l}{\cellcolor[HTML]{D9D9D9}} & \multicolumn{1}{l}{\cellcolor[HTML]{D9D9D9}} & \multicolumn{1}{l}{\cellcolor[HTML]{D9D9D9}} & \multicolumn{1}{l}{\cellcolor[HTML]{D9D9D9}} & \multicolumn{1}{l|}{\cellcolor[HTML]{D9D9D9}} &  & \multicolumn{1}{l}{\cellcolor[HTML]{D9D9D9}} & \multicolumn{1}{l}{\cellcolor[HTML]{D9D9D9}} & \multicolumn{1}{l}{\cellcolor[HTML]{D9D9D9}} & \multicolumn{1}{l}{\cellcolor[HTML]{D9D9D9}} & \multicolumn{1}{l}{\cellcolor[HTML]{D9D9D9}} \\ \hline
\rowcolor[HTML]{CFE2F3} 
\textbf{Sub Acc (exact)} & \multicolumn{1}{c}{\cellcolor[HTML]{CFE2F3}\textbf{0\%}} & \textbf{20\%} & \textbf{40\%} & \textbf{60\%} & \textbf{80\%} & \textbf{100\%} & \multicolumn{1}{c}{\cellcolor[HTML]{CFE2F3}\textbf{0\%}} & \textbf{20\%} & \textbf{40\%} & \textbf{60\%} & \textbf{80\%} & \textbf{100\%} & \multicolumn{1}{c}{\cellcolor[HTML]{CFE2F3}\textbf{0\%}} & \textbf{20\%} & \textbf{40\%} & \textbf{60\%} & \textbf{80\%} & \textbf{100\%} & \multicolumn{1}{c}{\cellcolor[HTML]{CFE2F3}\textbf{0\%}} & \textbf{20\%} & \textbf{40\%} & \textbf{60\%} & \textbf{80\%} & \textbf{100\%} \\ \hline
\textbf{GPT-3.5} & \cellcolor[HTML]{EFEFEF} & \textbf{26.9} & \textbf{28.2} & \textbf{28.6} & \textbf{29.1} & \textbf{30.6} & \cellcolor[HTML]{EFEFEF} & \textbf{28.4} & \textbf{30.8} & \textbf{31.7} & \textbf{32.9} & \textbf{33.0} & \cellcolor[HTML]{EFEFEF} & \textbf{29.6} & \textbf{31.0} & \textbf{31.4} & \textbf{31.7} & \textbf{33.2} & \cellcolor[HTML]{EFEFEF} & \textbf{28.2} & \textbf{29.0} & \textbf{29.8} & \textbf{29.9} & \textbf{30.1} \\
\textbf{GPT-4o-mini} & \cellcolor[HTML]{EFEFEF} & \textbf{21.0} & \textbf{21.8} & \textbf{23.8} & \textbf{25.0} & \textbf{26.3} & \cellcolor[HTML]{EFEFEF} & \textbf{25.6} & \textbf{25.4} & \textbf{27.9} & \textbf{29.2} & \textbf{29.6} & \cellcolor[HTML]{EFEFEF} & \textbf{25.2} & \textbf{26.3} & \textbf{27.6} & \textbf{28.2} & \textbf{28.9} & \cellcolor[HTML]{EFEFEF} & \textbf{21.7} & \textbf{23.3} & \textbf{24.0} & \textbf{24.0} & \textbf{25.7} \\
\textbf{Llama-3-70b} & \cellcolor[HTML]{EFEFEF} & \textbf{24.9} & \textbf{26.1} & \textbf{27.3} & \textbf{28.8} & \textbf{29.6} & \cellcolor[HTML]{EFEFEF} & \textbf{29.4} & \textbf{31.1} & \textbf{33.1} & \textbf{33.6} & \textbf{35.2} & \cellcolor[HTML]{EFEFEF} & \textbf{27.3} & \textbf{30.3} & \textbf{31.3} & \textbf{32.1} & \textbf{32.6} & \cellcolor[HTML]{EFEFEF} & \textbf{23.6} & \textbf{26.3} & \textbf{27.5} & \textbf{27.7} & \textbf{28.8} \\ \hline
\rowcolor[HTML]{D9D9D9} 
\multicolumn{1}{l|}{\cellcolor[HTML]{D9D9D9}} &  & \multicolumn{1}{l}{\cellcolor[HTML]{D9D9D9}} & \multicolumn{1}{l}{\cellcolor[HTML]{D9D9D9}} & \multicolumn{1}{l}{\cellcolor[HTML]{D9D9D9}} & \multicolumn{1}{l}{\cellcolor[HTML]{D9D9D9}} & \multicolumn{1}{l|}{\cellcolor[HTML]{D9D9D9}} &  & \multicolumn{1}{l}{\cellcolor[HTML]{D9D9D9}} & \multicolumn{1}{l}{\cellcolor[HTML]{D9D9D9}} & \multicolumn{1}{l}{\cellcolor[HTML]{D9D9D9}} & \multicolumn{1}{l}{\cellcolor[HTML]{D9D9D9}} & \multicolumn{1}{l|}{\cellcolor[HTML]{D9D9D9}} &  & \multicolumn{1}{l}{\cellcolor[HTML]{D9D9D9}} & \multicolumn{1}{l}{\cellcolor[HTML]{D9D9D9}} & \multicolumn{1}{l}{\cellcolor[HTML]{D9D9D9}} & \multicolumn{1}{l}{\cellcolor[HTML]{D9D9D9}} & \multicolumn{1}{l|}{\cellcolor[HTML]{D9D9D9}} &  & \multicolumn{1}{l}{\cellcolor[HTML]{D9D9D9}} & \multicolumn{1}{l}{\cellcolor[HTML]{D9D9D9}} & \multicolumn{1}{l}{\cellcolor[HTML]{D9D9D9}} & \multicolumn{1}{l}{\cellcolor[HTML]{D9D9D9}} & \multicolumn{1}{l}{\cellcolor[HTML]{D9D9D9}} \\ \hline
\rowcolor[HTML]{CFE2F3} 
\textbf{Sub Acc (gpt)} & \multicolumn{1}{c}{\cellcolor[HTML]{CFE2F3}\textbf{0\%}} & \textbf{20\%} & \textbf{40\%} & \textbf{60\%} & \textbf{80\%} & \textbf{100\%} & \multicolumn{1}{c}{\cellcolor[HTML]{CFE2F3}\textbf{0\%}} & \textbf{20\%} & \textbf{40\%} & \textbf{60\%} & \textbf{80\%} & \textbf{100\%} & \multicolumn{1}{c}{\cellcolor[HTML]{CFE2F3}\textbf{0\%}} & \textbf{20\%} & \textbf{40\%} & \textbf{60\%} & \textbf{80\%} & \textbf{100\%} & \multicolumn{1}{c}{\cellcolor[HTML]{CFE2F3}\textbf{0\%}} & \textbf{20\%} & \textbf{40\%} & \textbf{60\%} & \textbf{80\%} & \textbf{100\%} \\ \hline
\textbf{GPT-3.5} & \cellcolor[HTML]{EFEFEF} & \textbf{80.9} & \textbf{80.9} & \textbf{80.3} & \textbf{79.8} & \textbf{80.9} & \cellcolor[HTML]{EFEFEF} & \textbf{82.0} & \textbf{82.4} & \textbf{82.5} & \textbf{81.6} & \textbf{82.6} & \cellcolor[HTML]{EFEFEF} & \textbf{80.2} & \textbf{81.1} & \textbf{81.6} & \textbf{81.3} & \textbf{81.8} & \cellcolor[HTML]{EFEFEF} & \textbf{78.6} & \textbf{79.4} & \textbf{79.8} & \textbf{80.0} & \textbf{79.4} \\
\textbf{GPT-4o-mini} & \cellcolor[HTML]{EFEFEF} & \textbf{80.4} & \textbf{81.3} & \textbf{82.4} & \textbf{81.6} & \textbf{81.7} & \cellcolor[HTML]{EFEFEF} & \textbf{81.3} & \textbf{81.9} & \textbf{82.6} & \textbf{82.1} & \textbf{82.8} & \cellcolor[HTML]{EFEFEF} & \textbf{81.3} & \textbf{81.9} & \textbf{82.4} & \textbf{82.1} & \textbf{82.2} & \cellcolor[HTML]{EFEFEF} & \textbf{79.0} & \textbf{79.9} & \textbf{80.1} & \textbf{79.9} & \textbf{80.3} \\
\textbf{Llama-3-70b} & \cellcolor[HTML]{EFEFEF} & \textbf{80.1} & \textbf{80.2} & \textbf{80.7} & \textbf{80.4} & \textbf{81.0} & \cellcolor[HTML]{EFEFEF} & \textbf{82.0} & \textbf{82.9} & \textbf{83.2} & \textbf{82.9} & \textbf{83.5} & \cellcolor[HTML]{EFEFEF} & \textbf{81.3} & \textbf{82.0} & \textbf{82.4} & \textbf{82.9} & \textbf{82.7} & \cellcolor[HTML]{EFEFEF} & \textbf{80.0} & \textbf{80.8} & \textbf{81.1} & \textbf{80.6} & \textbf{81.0} \\ \hline
\rowcolor[HTML]{D9D9D9} 
\multicolumn{1}{l|}{\cellcolor[HTML]{D9D9D9}} &  & \multicolumn{1}{l}{\cellcolor[HTML]{D9D9D9}} & \multicolumn{1}{l}{\cellcolor[HTML]{D9D9D9}} & \multicolumn{1}{l}{\cellcolor[HTML]{D9D9D9}} & \multicolumn{1}{l}{\cellcolor[HTML]{D9D9D9}} & \multicolumn{1}{l|}{\cellcolor[HTML]{D9D9D9}} &  & \multicolumn{1}{l}{\cellcolor[HTML]{D9D9D9}} & \multicolumn{1}{l}{\cellcolor[HTML]{D9D9D9}} & \multicolumn{1}{l}{\cellcolor[HTML]{D9D9D9}} & \multicolumn{1}{l}{\cellcolor[HTML]{D9D9D9}} & \multicolumn{1}{l|}{\cellcolor[HTML]{D9D9D9}} &  & \multicolumn{1}{l}{\cellcolor[HTML]{D9D9D9}} & \multicolumn{1}{l}{\cellcolor[HTML]{D9D9D9}} & \multicolumn{1}{l}{\cellcolor[HTML]{D9D9D9}} & \multicolumn{1}{l}{\cellcolor[HTML]{D9D9D9}} & \multicolumn{1}{l|}{\cellcolor[HTML]{D9D9D9}} &  & \multicolumn{1}{l}{\cellcolor[HTML]{D9D9D9}} & \multicolumn{1}{l}{\cellcolor[HTML]{D9D9D9}} & \multicolumn{1}{l}{\cellcolor[HTML]{D9D9D9}} & \multicolumn{1}{l}{\cellcolor[HTML]{D9D9D9}} & \multicolumn{1}{l}{\cellcolor[HTML]{D9D9D9}} \\
\rowcolor[HTML]{D9D9D9} 
\multicolumn{1}{l|}{\cellcolor[HTML]{D9D9D9}} &  & \multicolumn{1}{l}{\cellcolor[HTML]{D9D9D9}} & \multicolumn{1}{l}{\cellcolor[HTML]{D9D9D9}} & \multicolumn{1}{l}{\cellcolor[HTML]{D9D9D9}} & \multicolumn{1}{l}{\cellcolor[HTML]{D9D9D9}} & \multicolumn{1}{l|}{\cellcolor[HTML]{D9D9D9}} &  & \multicolumn{1}{l}{\cellcolor[HTML]{D9D9D9}} & \multicolumn{1}{l}{\cellcolor[HTML]{D9D9D9}} & \multicolumn{1}{l}{\cellcolor[HTML]{D9D9D9}} & \multicolumn{1}{l}{\cellcolor[HTML]{D9D9D9}} & \multicolumn{1}{l|}{\cellcolor[HTML]{D9D9D9}} &  & \multicolumn{1}{l}{\cellcolor[HTML]{D9D9D9}} & \multicolumn{1}{l}{\cellcolor[HTML]{D9D9D9}} & \multicolumn{1}{l}{\cellcolor[HTML]{D9D9D9}} & \multicolumn{1}{l}{\cellcolor[HTML]{D9D9D9}} & \multicolumn{1}{l|}{\cellcolor[HTML]{D9D9D9}} &  & \multicolumn{1}{l}{\cellcolor[HTML]{D9D9D9}} & \multicolumn{1}{l}{\cellcolor[HTML]{D9D9D9}} & \multicolumn{1}{l}{\cellcolor[HTML]{D9D9D9}} & \multicolumn{1}{l}{\cellcolor[HTML]{D9D9D9}} & \multicolumn{1}{l}{\cellcolor[HTML]{D9D9D9}} \\
\rowcolor[HTML]{D9D9D9} 
\multicolumn{1}{l|}{\cellcolor[HTML]{D9D9D9}} &  & \multicolumn{1}{l}{\cellcolor[HTML]{D9D9D9}} & \multicolumn{1}{l}{\cellcolor[HTML]{D9D9D9}} & \multicolumn{1}{l}{\cellcolor[HTML]{D9D9D9}} & \multicolumn{1}{l}{\cellcolor[HTML]{D9D9D9}} & \multicolumn{1}{l|}{\cellcolor[HTML]{D9D9D9}} &  & \multicolumn{1}{l}{\cellcolor[HTML]{D9D9D9}} & \multicolumn{1}{l}{\cellcolor[HTML]{D9D9D9}} & \multicolumn{1}{l}{\cellcolor[HTML]{D9D9D9}} & \multicolumn{1}{l}{\cellcolor[HTML]{D9D9D9}} & \multicolumn{1}{l|}{\cellcolor[HTML]{D9D9D9}} &  & \multicolumn{1}{l}{\cellcolor[HTML]{D9D9D9}} & \multicolumn{1}{l}{\cellcolor[HTML]{D9D9D9}} & \multicolumn{1}{l}{\cellcolor[HTML]{D9D9D9}} & \multicolumn{1}{l}{\cellcolor[HTML]{D9D9D9}} & \multicolumn{1}{l|}{\cellcolor[HTML]{D9D9D9}} &  & \multicolumn{1}{l}{\cellcolor[HTML]{D9D9D9}} & \multicolumn{1}{l}{\cellcolor[HTML]{D9D9D9}} & \multicolumn{1}{l}{\cellcolor[HTML]{D9D9D9}} & \multicolumn{1}{l}{\cellcolor[HTML]{D9D9D9}} & \multicolumn{1}{l}{\cellcolor[HTML]{D9D9D9}} \\ \hline
\cellcolor[HTML]{EA9999}\textbf{10 doc} & \multicolumn{6}{c|}{\textbf{BioASQ}} & \multicolumn{6}{c|}{\textbf{PubmedQA}} & \multicolumn{6}{c|}{\textbf{MedQA}} & \multicolumn{6}{c}{\textbf{MMLU}} \\ \hline
\rowcolor[HTML]{FCE5CD} 
\textbf{Main Acc} & \multicolumn{1}{c}{\cellcolor[HTML]{FCE5CD}\textbf{0\%}} & \textbf{20\%} & \textbf{40\%} & \textbf{60\%} & \textbf{80\%} & \textbf{100\%} & \multicolumn{1}{c}{\cellcolor[HTML]{FCE5CD}\textbf{0\%}} & \textbf{20\%} & \textbf{40\%} & \textbf{60\%} & \textbf{80\%} & \textbf{100\%} & \multicolumn{1}{c}{\cellcolor[HTML]{FCE5CD}\textbf{0\%}} & \textbf{20\%} & \textbf{40\%} & \textbf{60\%} & \textbf{80\%} & \textbf{100\%} & \multicolumn{1}{c}{\cellcolor[HTML]{FCE5CD}\textbf{0\%}} & \textbf{20\%} & \textbf{40\%} & \textbf{60\%} & \textbf{80\%} & \textbf{100\%} \\ \hline
\rowcolor[HTML]{FFFFFF} 
\textbf{GPT-3.5} & \cellcolor[HTML]{EFEFEF} & \textbf{73.5} & \textbf{80.3} & \textbf{82.7} & \textbf{83.2} & \textbf{83.8} & \cellcolor[HTML]{EFEFEF} & \textbf{56.6} & \textbf{62.8} & \textbf{65.0} & \textbf{67.6} & \textbf{69.0} & \cellcolor[HTML]{EFEFEF} & \textbf{55.7} & \textbf{55.0} & \textbf{58.3} & \textbf{58.6} & \textbf{59.6} & \cellcolor[HTML]{EFEFEF} & \textbf{67.2} & \textbf{67.9} & \textbf{67.6} & \textbf{68.9} & \textbf{68.3} \\
\rowcolor[HTML]{FFFFFF} 
\textbf{GPT-4o-mini} & \cellcolor[HTML]{EFEFEF} & \textbf{79.5} & \textbf{83.8} & \textbf{86.1} & \textbf{89.0} & \textbf{89.6} & \cellcolor[HTML]{EFEFEF} & \textbf{51.6} & \textbf{58.6} & \textbf{62.6} & \textbf{65.6} & \textbf{66.2} & \cellcolor[HTML]{EFEFEF} & \textbf{73.1} & \textbf{73.7} & \textbf{74.2} & \textbf{75.2} & \textbf{74.0} & \cellcolor[HTML]{EFEFEF} & \textbf{82.4} & \textbf{82.3} & \textbf{82.2} & \textbf{82.4} & \textbf{84.1} \\
\rowcolor[HTML]{FFFFFF} 
\textbf{Llama-3-70b} & \cellcolor[HTML]{EFEFEF} & \textbf{74.0} & \textbf{83.0} & \textbf{84.3} & \textbf{89.2} & \textbf{89.6} & \cellcolor[HTML]{EFEFEF} & \textbf{54.2} & \textbf{63.6} & \textbf{65.0} & \textbf{68.6} & \textbf{69.4} & \cellcolor[HTML]{EFEFEF} & \textbf{71.2} & \textbf{70.7} & \textbf{72.4} & \textbf{74.0} & \textbf{74.0} & \cellcolor[HTML]{EFEFEF} & \textbf{75.8} & \textbf{77.1} & \textbf{78.5} & \textbf{78.2} & \textbf{80.8} \\ \hline
\rowcolor[HTML]{D9D9D9} 
\multicolumn{1}{l|}{\cellcolor[HTML]{D9D9D9}} &  & \multicolumn{1}{l}{\cellcolor[HTML]{D9D9D9}} & \multicolumn{1}{l}{\cellcolor[HTML]{D9D9D9}} & \multicolumn{1}{l}{\cellcolor[HTML]{D9D9D9}} & \multicolumn{1}{l}{\cellcolor[HTML]{D9D9D9}} & \multicolumn{1}{l|}{\cellcolor[HTML]{D9D9D9}} &  & \multicolumn{1}{l}{\cellcolor[HTML]{D9D9D9}} & \multicolumn{1}{l}{\cellcolor[HTML]{D9D9D9}} & \multicolumn{1}{l}{\cellcolor[HTML]{D9D9D9}} & \multicolumn{1}{l}{\cellcolor[HTML]{D9D9D9}} & \multicolumn{1}{l|}{\cellcolor[HTML]{D9D9D9}} &  & \multicolumn{1}{l}{\cellcolor[HTML]{D9D9D9}} & \multicolumn{1}{l}{\cellcolor[HTML]{D9D9D9}} & \multicolumn{1}{l}{\cellcolor[HTML]{D9D9D9}} & \multicolumn{1}{l}{\cellcolor[HTML]{D9D9D9}} & \multicolumn{1}{l|}{\cellcolor[HTML]{D9D9D9}} &  & \multicolumn{1}{l}{\cellcolor[HTML]{D9D9D9}} & \multicolumn{1}{l}{\cellcolor[HTML]{D9D9D9}} & \multicolumn{1}{l}{\cellcolor[HTML]{D9D9D9}} & \multicolumn{1}{l}{\cellcolor[HTML]{D9D9D9}} & \multicolumn{1}{l}{\cellcolor[HTML]{D9D9D9}} \\ \hline
\rowcolor[HTML]{CFE2F3} 
\textbf{Sub Acc (exact)} & \multicolumn{1}{c}{\cellcolor[HTML]{CFE2F3}\textbf{0\%}} & \textbf{20\%} & \textbf{40\%} & \textbf{60\%} & \textbf{80\%} & \textbf{100\%} & \multicolumn{1}{c}{\cellcolor[HTML]{CFE2F3}\textbf{0\%}} & \textbf{20\%} & \textbf{40\%} & \textbf{60\%} & \textbf{80\%} & \textbf{100\%} & \multicolumn{1}{c}{\cellcolor[HTML]{CFE2F3}\textbf{0\%}} & \textbf{20\%} & \textbf{40\%} & \textbf{60\%} & \textbf{80\%} & \textbf{100\%} & \multicolumn{1}{c}{\cellcolor[HTML]{CFE2F3}\textbf{0\%}} & \textbf{20\%} & \textbf{40\%} & \textbf{60\%} & \textbf{80\%} & \textbf{100\%} \\ \hline
\textbf{GPT-3.5} & \cellcolor[HTML]{EFEFEF} & \textbf{28.0} & \textbf{29.0} & \textbf{27.9} & \textbf{27.2} & \textbf{27.9} & \cellcolor[HTML]{EFEFEF} & \textbf{30.8} & \textbf{32.8} & \textbf{33.1} & \textbf{32.8} & \textbf{33.4} & \cellcolor[HTML]{EFEFEF} & \textbf{30.8} & \textbf{31.5} & \textbf{31.5} & \textbf{32.1} & \textbf{32.8} & \cellcolor[HTML]{EFEFEF} & \textbf{28.2} & \textbf{29.4} & \textbf{29.3} & \textbf{30.0} & \textbf{30.7} \\
\textbf{GPT-4o-mini} & \cellcolor[HTML]{EFEFEF} & \textbf{21.2} & \textbf{25.2} & \textbf{26.1} & \textbf{25.9} & \textbf{26.8} & \cellcolor[HTML]{EFEFEF} & \textbf{25.6} & \textbf{28.3} & \textbf{30.5} & \textbf{30.9} & \textbf{32.9} & \cellcolor[HTML]{EFEFEF} & \textbf{25.7} & \textbf{27.4} & \textbf{28.7} & \textbf{30.1} & \textbf{30.3} & \cellcolor[HTML]{EFEFEF} & \textbf{23.1} & \textbf{24.0} & \textbf{25.9} & \textbf{26.8} & \textbf{27.2} \\
\textbf{Llama-3-70b} & \cellcolor[HTML]{EFEFEF} & \textbf{26.0} & \textbf{27.7} & \textbf{28.7} & \textbf{29.8} & \textbf{31.0} & \cellcolor[HTML]{EFEFEF} & \textbf{30.6} & \textbf{34.0} & \textbf{36.3} & \textbf{37.9} & \textbf{38.9} & \cellcolor[HTML]{EFEFEF} & \textbf{30.7} & \textbf{32.1} & \textbf{32.4} & \textbf{33.4} & \textbf{33.2} & \cellcolor[HTML]{EFEFEF} & \textbf{25.9} & \textbf{27.5} & \textbf{29.3} & \textbf{29.8} & \textbf{30.8} \\ \hline
\rowcolor[HTML]{D9D9D9} 
\multicolumn{1}{l|}{\cellcolor[HTML]{D9D9D9}} &  & \multicolumn{1}{l}{\cellcolor[HTML]{D9D9D9}} & \multicolumn{1}{l}{\cellcolor[HTML]{D9D9D9}} & \multicolumn{1}{l}{\cellcolor[HTML]{D9D9D9}} & \multicolumn{1}{l}{\cellcolor[HTML]{D9D9D9}} & \multicolumn{1}{l|}{\cellcolor[HTML]{D9D9D9}} &  & \multicolumn{1}{l}{\cellcolor[HTML]{D9D9D9}} & \multicolumn{1}{l}{\cellcolor[HTML]{D9D9D9}} & \multicolumn{1}{l}{\cellcolor[HTML]{D9D9D9}} & \multicolumn{1}{l}{\cellcolor[HTML]{D9D9D9}} & \multicolumn{1}{l|}{\cellcolor[HTML]{D9D9D9}} &  & \multicolumn{1}{l}{\cellcolor[HTML]{D9D9D9}} & \multicolumn{1}{l}{\cellcolor[HTML]{D9D9D9}} & \multicolumn{1}{l}{\cellcolor[HTML]{D9D9D9}} & \multicolumn{1}{l}{\cellcolor[HTML]{D9D9D9}} & \multicolumn{1}{l|}{\cellcolor[HTML]{D9D9D9}} &  & \multicolumn{1}{l}{\cellcolor[HTML]{D9D9D9}} & \multicolumn{1}{l}{\cellcolor[HTML]{D9D9D9}} & \multicolumn{1}{l}{\cellcolor[HTML]{D9D9D9}} & \multicolumn{1}{l}{\cellcolor[HTML]{D9D9D9}} & \multicolumn{1}{l}{\cellcolor[HTML]{D9D9D9}} \\ \hline
\rowcolor[HTML]{CFE2F3} 
\textbf{Sub Acc (gpt)} & \multicolumn{1}{c}{\cellcolor[HTML]{CFE2F3}\textbf{0\%}} & \textbf{20\%} & \textbf{40\%} & \textbf{60\%} & \textbf{80\%} & \textbf{100\%} & \multicolumn{1}{c}{\cellcolor[HTML]{CFE2F3}\textbf{0\%}} & \textbf{20\%} & \textbf{40\%} & \textbf{60\%} & \textbf{80\%} & \textbf{100\%} & \multicolumn{1}{c}{\cellcolor[HTML]{CFE2F3}\textbf{0\%}} & \textbf{20\%} & \textbf{40\%} & \textbf{60\%} & \textbf{80\%} & \textbf{100\%} & \multicolumn{1}{c}{\cellcolor[HTML]{CFE2F3}\textbf{0\%}} & \textbf{20\%} & \textbf{40\%} & \textbf{60\%} & \textbf{80\%} & \textbf{100\%} \\ \hline
\textbf{GPT-3.5} & \cellcolor[HTML]{EFEFEF} & \textbf{80.3} & \textbf{79.2} & \textbf{76.2} & \textbf{75.9} & \textbf{78.2} & \cellcolor[HTML]{EFEFEF} & \textbf{82.2} & \textbf{81.8} & \textbf{81.2} & \textbf{80.6} & \textbf{81.1} & \cellcolor[HTML]{EFEFEF} & \textbf{81.1} & \textbf{81.0} & \textbf{80.5} & \textbf{80.4} & \textbf{81.4} & \cellcolor[HTML]{EFEFEF} & \textbf{78.5} & \textbf{78.4} & \textbf{77.8} & \textbf{77.7} & \textbf{79.0} \\
\textbf{GPT-4o-mini} & \cellcolor[HTML]{EFEFEF} & \textbf{81.0} & \textbf{81.6} & \textbf{80.7} & \textbf{80.2} & \textbf{80.7} & \cellcolor[HTML]{EFEFEF} & \textbf{82.6} & \textbf{81.5} & \textbf{82.3} & \textbf{81.6} & \textbf{82.4} & \cellcolor[HTML]{EFEFEF} & \textbf{81.7} & \textbf{81.8} & \textbf{82.0} & \textbf{82.1} & \textbf{82.2} & \cellcolor[HTML]{EFEFEF} & \textbf{79.6} & \textbf{79.7} & \textbf{79.8} & \textbf{80.2} & \textbf{80.4} \\
\textbf{Llama-3-70b} & \cellcolor[HTML]{EFEFEF} & \textbf{79.8} & \textbf{79.8} & \textbf{80.1} & \textbf{79.7} & \textbf{80.6} & \cellcolor[HTML]{EFEFEF} & \textbf{82.7} & \textbf{82.6} & \textbf{82.8} & \textbf{82.5} & \textbf{83.2} & \cellcolor[HTML]{EFEFEF} & \textbf{81.9} & \textbf{82.1} & \textbf{82.6} & \textbf{82.8} & \textbf{83.1} & \cellcolor[HTML]{EFEFEF} & \textbf{79.7} & \textbf{79.9} & \textbf{80.4} & \textbf{80.9} & \textbf{81.3}
\end{tabular}
}
\caption{ \small Integration test full results table, including main question accuracy and sub question accuracy (exact-match and GPT-based).}
\label{tab:inin-full}
\end{table*}

\begin{table*}[htbp]
\center
\small
\resizebox{0.99\textwidth}{!}{


\begin{tabular}{c|cccccc|cccccc|cccccc|cccccc}
\cellcolor[HTML]{EA9999}\textbf{5 doc} & \multicolumn{6}{c|}{\textbf{BioASQ}} & \multicolumn{6}{c|}{\textbf{PubmedQA}} & \multicolumn{6}{c|}{\textbf{MedQA}} & \multicolumn{6}{c}{\textbf{MMLU}} \\ \hline
\rowcolor[HTML]{FCE5CD} 
\textbf{Main Acc} & \textbf{0\%} & \textbf{20\%} & \textbf{40\%} & \textbf{60\%} & \textbf{80\%} & \textbf{100\%} & \textbf{0\%} & \textbf{20\%} & \textbf{40\%} & \textbf{60\%} & \textbf{80\%} & \textbf{100\%} & \textbf{0\%} & \textbf{20\%} & \textbf{40\%} & \textbf{60\%} & \textbf{80\%} & \textbf{100\%} & \textbf{0\%} & \textbf{20\%} & \textbf{40\%} & \textbf{60\%} & \textbf{80\%} & \textbf{100\%} \\ \hline
\rowcolor[HTML]{FFFFFF} 
\textbf{GPT-3.5} & \textbf{63.3} & \textbf{67.8} & \textbf{72.3} & \textbf{76.2} & \textbf{77.0} & \textbf{79.8} & \textbf{41.6} & \textbf{45.2} & \textbf{48.2} & \textbf{54.6} & \textbf{56.6} & \textbf{64.4} & \textbf{50.4} & \textbf{51.7} & \textbf{53.3} & \textbf{53.3} & \textbf{55.2} & \textbf{56.7} & \textbf{60.1} & \textbf{61.8} & \textbf{62.4} & \textbf{64.5} & \textbf{65.8} & \textbf{65.8} \\
\rowcolor[HTML]{FFFFFF} 
\textbf{GPT-4o-mini} & \textbf{70.6} & \textbf{76.1} & \textbf{78.5} & \textbf{81.1} & \textbf{84.3} & \textbf{85.3} & \textbf{40.8} & \textbf{45.4} & \textbf{48.4} & \textbf{50.2} & \textbf{53.6} & \textbf{59.4} & \textbf{71.4} & \textbf{70.8} & \textbf{71.1} & \textbf{71.9} & \textbf{72.6} & \textbf{71.4} & \textbf{80.4} & \textbf{80.5} & \textbf{80.9} & \textbf{80.6} & \textbf{80.9} & \textbf{81.4} \\
\rowcolor[HTML]{FFFFFF} 
\textbf{Llama-3-70b} & \textbf{68.3} & \textbf{70.4} & \textbf{75.6} & \textbf{80.6} & \textbf{81.4} & \textbf{84.0} & \textbf{42.2} & \textbf{44.8} & \textbf{49.4} & \textbf{51.4} & \textbf{57.0} & \textbf{62.8} & \textbf{67.3} & \textbf{67.1} & \textbf{66.4} & \textbf{69.8} & \textbf{70.2} & \textbf{71.9} & \textbf{69.9} & \textbf{72.9} & \textbf{71.9} & \textbf{75.0} & \textbf{73.9} & \textbf{76.0} \\ \hline
\rowcolor[HTML]{D9D9D9} 
\multicolumn{1}{l|}{\cellcolor[HTML]{D9D9D9}} & \multicolumn{1}{l}{\cellcolor[HTML]{D9D9D9}} & \multicolumn{1}{l}{\cellcolor[HTML]{D9D9D9}} & \multicolumn{1}{l}{\cellcolor[HTML]{D9D9D9}} & \multicolumn{1}{l}{\cellcolor[HTML]{D9D9D9}} & \multicolumn{1}{l}{\cellcolor[HTML]{D9D9D9}} & \multicolumn{1}{l|}{\cellcolor[HTML]{D9D9D9}} & \multicolumn{1}{l}{\cellcolor[HTML]{D9D9D9}} & \multicolumn{1}{l}{\cellcolor[HTML]{D9D9D9}} & \multicolumn{1}{l}{\cellcolor[HTML]{D9D9D9}} & \multicolumn{1}{l}{\cellcolor[HTML]{D9D9D9}} & \multicolumn{1}{l}{\cellcolor[HTML]{D9D9D9}} & \multicolumn{1}{l|}{\cellcolor[HTML]{D9D9D9}} & \multicolumn{1}{l}{\cellcolor[HTML]{D9D9D9}} & \multicolumn{1}{l}{\cellcolor[HTML]{D9D9D9}} & \multicolumn{1}{l}{\cellcolor[HTML]{D9D9D9}} & \multicolumn{1}{l}{\cellcolor[HTML]{D9D9D9}} & \multicolumn{1}{l}{\cellcolor[HTML]{D9D9D9}} & \multicolumn{1}{l|}{\cellcolor[HTML]{D9D9D9}} & \multicolumn{1}{l}{\cellcolor[HTML]{D9D9D9}} & \multicolumn{1}{l}{\cellcolor[HTML]{D9D9D9}} & \multicolumn{1}{l}{\cellcolor[HTML]{D9D9D9}} & \multicolumn{1}{l}{\cellcolor[HTML]{D9D9D9}} & \multicolumn{1}{l}{\cellcolor[HTML]{D9D9D9}} & \multicolumn{1}{l}{\cellcolor[HTML]{D9D9D9}} \\ \hline
\rowcolor[HTML]{CFE2F3} 
\textbf{Sub Acc (exact)} & \textbf{0\%} & \textbf{20\%} & \textbf{40\%} & \textbf{60\%} & \textbf{80\%} & \textbf{100\%} & \textbf{0\%} & \textbf{20\%} & \textbf{40\%} & \textbf{60\%} & \textbf{80\%} & \textbf{100\%} & \textbf{0\%} & \textbf{20\%} & \textbf{40\%} & \textbf{60\%} & \textbf{80\%} & \textbf{100\%} & \textbf{0\%} & \textbf{20\%} & \textbf{40\%} & \textbf{60\%} & \textbf{80\%} & \textbf{100\%} \\ \hline
\textbf{GPT-3.5} & \textbf{0.7} & \textbf{8.2} & \textbf{14.1} & \textbf{23.4} & \textbf{28.5} & \textbf{35.5} & \textbf{0.2} & \textbf{9.2} & \textbf{17.9} & \textbf{27.7} & \textbf{36.3} & \textbf{46.1} & \textbf{0.3} & \textbf{8.7} & \textbf{15.8} & \textbf{24.1} & \textbf{30.8} & \textbf{38.4} & \textbf{0.3} & \textbf{7.7} & \textbf{14.0} & \textbf{21.5} & \textbf{27.5} & \textbf{34.4} \\
\textbf{GPT-4o-mini} & \textbf{0.9} & \textbf{6.2} & \textbf{10.7} & \textbf{17.1} & \textbf{22.5} & \textbf{27.9} & \textbf{0.3} & \textbf{7.1} & \textbf{13.2} & \textbf{21.0} & \textbf{27.0} & \textbf{35.0} & \textbf{0.8} & \textbf{6.9} & \textbf{12.6} & \textbf{19.7} & \textbf{25.4} & \textbf{31.9} & \textbf{1.1} & \textbf{5.9} & \textbf{11.0} & \textbf{17.3} & \textbf{21.5} & \textbf{27.0} \\
\textbf{Llama-3-70b} & \textbf{0.8} & \textbf{8.2} & \textbf{14.0} & \textbf{20.9} & \textbf{28.0} & \textbf{35.1} & \textbf{0.2} & \textbf{9.8} & \textbf{18.1} & \textbf{27.8} & \textbf{35.7} & \textbf{45.9} & \textbf{0.7} & \textbf{8.7} & \textbf{15.6} & \textbf{23.8} & \textbf{30.1} & \textbf{37.8} & \textbf{0.9} & \textbf{8.1} & \textbf{13.9} & \textbf{20.9} & \textbf{26.9} & \textbf{33.7} \\ \hline
\rowcolor[HTML]{D9D9D9} 
\multicolumn{1}{l|}{\cellcolor[HTML]{D9D9D9}} & \multicolumn{1}{l}{\cellcolor[HTML]{D9D9D9}} & \multicolumn{1}{l}{\cellcolor[HTML]{D9D9D9}} & \multicolumn{1}{l}{\cellcolor[HTML]{D9D9D9}} & \multicolumn{1}{l}{\cellcolor[HTML]{D9D9D9}} & \multicolumn{1}{l}{\cellcolor[HTML]{D9D9D9}} & \multicolumn{1}{l|}{\cellcolor[HTML]{D9D9D9}} & \multicolumn{1}{l}{\cellcolor[HTML]{D9D9D9}} & \multicolumn{1}{l}{\cellcolor[HTML]{D9D9D9}} & \multicolumn{1}{l}{\cellcolor[HTML]{D9D9D9}} & \multicolumn{1}{l}{\cellcolor[HTML]{D9D9D9}} & \multicolumn{1}{l}{\cellcolor[HTML]{D9D9D9}} & \multicolumn{1}{l|}{\cellcolor[HTML]{D9D9D9}} & \multicolumn{1}{l}{\cellcolor[HTML]{D9D9D9}} & \multicolumn{1}{l}{\cellcolor[HTML]{D9D9D9}} & \multicolumn{1}{l}{\cellcolor[HTML]{D9D9D9}} & \multicolumn{1}{l}{\cellcolor[HTML]{D9D9D9}} & \multicolumn{1}{l}{\cellcolor[HTML]{D9D9D9}} & \multicolumn{1}{l|}{\cellcolor[HTML]{D9D9D9}} & \multicolumn{1}{l}{\cellcolor[HTML]{D9D9D9}} & \multicolumn{1}{l}{\cellcolor[HTML]{D9D9D9}} & \multicolumn{1}{l}{\cellcolor[HTML]{D9D9D9}} & \multicolumn{1}{l}{\cellcolor[HTML]{D9D9D9}} & \multicolumn{1}{l}{\cellcolor[HTML]{D9D9D9}} & \multicolumn{1}{l}{\cellcolor[HTML]{D9D9D9}} \\ \hline
\rowcolor[HTML]{CFE2F3} 
\textbf{Sub Acc (gpt)} & \textbf{0\%} & \textbf{20\%} & \textbf{40\%} & \textbf{60\%} & \textbf{80\%} & \textbf{100\%} & \textbf{0\%} & \textbf{20\%} & \textbf{40\%} & \textbf{60\%} & \textbf{80\%} & \textbf{100\%} & \textbf{0\%} & \textbf{20\%} & \textbf{40\%} & \textbf{60\%} & \textbf{80\%} & \textbf{100\%} & \textbf{0\%} & \textbf{20\%} & \textbf{40\%} & \textbf{60\%} & \textbf{80\%} & \textbf{100\%} \\ \hline
\textbf{GPT-3.5} & \textbf{4.5} & \textbf{20.4} & \textbf{33.8} & \textbf{50.3} & \textbf{64.0} & \textbf{79.7} & \textbf{1.8} & \textbf{17.9} & \textbf{33.1} & \textbf{50.5} & \textbf{65.2} & \textbf{81.3} & \textbf{2.0} & \textbf{18.8} & \textbf{34.5} & \textbf{50.1} & \textbf{66.0} & \textbf{82.1} & \textbf{2.5} & \textbf{18.5} & \textbf{33.3} & \textbf{49.7} & \textbf{64.7} & \textbf{80.0} \\
\textbf{GPT-4o-mini} & \textbf{9.1} & \textbf{24.9} & \textbf{38.6} & \textbf{53.8} & \textbf{67.2} & \textbf{82.0} & \textbf{3.0} & \textbf{19.9} & \textbf{35.0} & \textbf{52.0} & \textbf{66.9} & \textbf{83.4} & \textbf{6.9} & \textbf{23.1} & \textbf{38.6} & \textbf{54.5} & \textbf{69.6} & \textbf{84.6} & \textbf{8.0} & \textbf{23.1} & \textbf{37.9} & \textbf{53.3} & \textbf{67.8} & \textbf{82.4} \\
\textbf{Llama-3-70b} & \textbf{6.9} & \textbf{22.9} & \textbf{36.3} & \textbf{52.0} & \textbf{66.2} & \textbf{82.0} & \textbf{2.6} & \textbf{19.3} & \textbf{34.6} & \textbf{51.5} & \textbf{67.6} & \textbf{83.8} & \textbf{4.6} & \textbf{21.7} & \textbf{37.6} & \textbf{53.2} & \textbf{68.7} & \textbf{85.2} & \textbf{6.0} & \textbf{21.8} & \textbf{37.2} & \textbf{52.6} & \textbf{67.5} & \textbf{83.4} \\ \hline
\rowcolor[HTML]{D9D9D9} 
\multicolumn{1}{l|}{\cellcolor[HTML]{D9D9D9}} & \multicolumn{1}{l}{\cellcolor[HTML]{D9D9D9}} & \multicolumn{1}{l}{\cellcolor[HTML]{D9D9D9}} & \multicolumn{1}{l}{\cellcolor[HTML]{D9D9D9}} & \multicolumn{1}{l}{\cellcolor[HTML]{D9D9D9}} & \multicolumn{1}{l}{\cellcolor[HTML]{D9D9D9}} & \multicolumn{1}{l|}{\cellcolor[HTML]{D9D9D9}} & \multicolumn{1}{l}{\cellcolor[HTML]{D9D9D9}} & \multicolumn{1}{l}{\cellcolor[HTML]{D9D9D9}} & \multicolumn{1}{l}{\cellcolor[HTML]{D9D9D9}} & \multicolumn{1}{l}{\cellcolor[HTML]{D9D9D9}} & \multicolumn{1}{l}{\cellcolor[HTML]{D9D9D9}} & \multicolumn{1}{l|}{\cellcolor[HTML]{D9D9D9}} & \multicolumn{1}{l}{\cellcolor[HTML]{D9D9D9}} & \multicolumn{1}{l}{\cellcolor[HTML]{D9D9D9}} & \multicolumn{1}{l}{\cellcolor[HTML]{D9D9D9}} & \multicolumn{1}{l}{\cellcolor[HTML]{D9D9D9}} & \multicolumn{1}{l}{\cellcolor[HTML]{D9D9D9}} & \multicolumn{1}{l|}{\cellcolor[HTML]{D9D9D9}} & \multicolumn{1}{l}{\cellcolor[HTML]{D9D9D9}} & \multicolumn{1}{l}{\cellcolor[HTML]{D9D9D9}} & \multicolumn{1}{l}{\cellcolor[HTML]{D9D9D9}} & \multicolumn{1}{l}{\cellcolor[HTML]{D9D9D9}} & \multicolumn{1}{l}{\cellcolor[HTML]{D9D9D9}} & \multicolumn{1}{l}{\cellcolor[HTML]{D9D9D9}} \\ \hline
\rowcolor[HTML]{CFE2F3} 
\textbf{Fact Detect} & \textbf{0\%} & \textbf{20\%} & \textbf{40\%} & \textbf{60\%} & \textbf{80\%} & \textbf{100\%} & \textbf{0\%} & \textbf{20\%} & \textbf{40\%} & \textbf{60\%} & \textbf{80\%} & \textbf{100\%} & \textbf{0\%} & \textbf{20\%} & \textbf{40\%} & \textbf{60\%} & \textbf{80\%} & \textbf{100\%} & \textbf{0\%} & \textbf{20\%} & \textbf{40\%} & \textbf{60\%} & \textbf{80\%} & \textbf{100\%} \\ \hline
\textbf{GPT-3.5} & \textbf{28.8} & \textbf{45.2} & \textbf{55.1} & \textbf{67.7} & \textbf{76.0} & \textbf{88.1} & \textbf{15.3} & \textbf{33.4} & \textbf{49.6} & \textbf{64.4} & \textbf{78.7} & \textbf{94.4} & \textbf{16.2} & \textbf{36.3} & \textbf{51.1} & \textbf{64.5} & \textbf{79.0} & \textbf{93.2} & \textbf{17.9} & \textbf{37.0} & \textbf{50.6} & \textbf{63.8} & \textbf{77.5} & \textbf{92.0} \\
\textbf{GPT-4o-mini} & \textbf{13.6} & \textbf{33.1} & \textbf{50.0} & \textbf{66.7} & \textbf{81.4} & \textbf{96.8} & \textbf{10.0} & \textbf{29.5} & \textbf{48.0} & \textbf{64.6} & \textbf{80.6} & \textbf{98.2} & \textbf{14.4} & \textbf{35.2} & \textbf{49.8} & \textbf{66.0} & \textbf{79.4} & \textbf{94.7} & \textbf{14.3} & \textbf{33.9} & \textbf{49.5} & \textbf{65.6} & \textbf{79.9} & \textbf{95.0} \\
\textbf{Llama-3-70b} & \textbf{8.3} & \textbf{27.4} & \textbf{44.6} & \textbf{63.5} & \textbf{80.1} & \textbf{99.5} & \textbf{8.2} & \textbf{27.8} & \textbf{45.2} & \textbf{63.3} & \textbf{81.1} & \textbf{99.9} & \textbf{13.9} & \textbf{32.4} & \textbf{49.7} & \textbf{65.6} & \textbf{82.3} & \textbf{99.5} & \textbf{13.2} & \textbf{32.3} & \textbf{49.0} & \textbf{64.9} & \textbf{82.0} & \textbf{99.3} \\ \hline
\rowcolor[HTML]{D9D9D9} 
\multicolumn{1}{l|}{\cellcolor[HTML]{D9D9D9}} & \multicolumn{1}{l}{\cellcolor[HTML]{D9D9D9}} & \multicolumn{1}{l}{\cellcolor[HTML]{D9D9D9}} & \multicolumn{1}{l}{\cellcolor[HTML]{D9D9D9}} & \multicolumn{1}{l}{\cellcolor[HTML]{D9D9D9}} & \multicolumn{1}{l}{\cellcolor[HTML]{D9D9D9}} & \multicolumn{1}{l|}{\cellcolor[HTML]{D9D9D9}} & \multicolumn{1}{l}{\cellcolor[HTML]{D9D9D9}} & \multicolumn{1}{l}{\cellcolor[HTML]{D9D9D9}} & \multicolumn{1}{l}{\cellcolor[HTML]{D9D9D9}} & \multicolumn{1}{l}{\cellcolor[HTML]{D9D9D9}} & \multicolumn{1}{l}{\cellcolor[HTML]{D9D9D9}} & \multicolumn{1}{l|}{\cellcolor[HTML]{D9D9D9}} & \multicolumn{1}{l}{\cellcolor[HTML]{D9D9D9}} & \multicolumn{1}{l}{\cellcolor[HTML]{D9D9D9}} & \multicolumn{1}{l}{\cellcolor[HTML]{D9D9D9}} & \multicolumn{1}{l}{\cellcolor[HTML]{D9D9D9}} & \multicolumn{1}{l}{\cellcolor[HTML]{D9D9D9}} & \multicolumn{1}{l|}{\cellcolor[HTML]{D9D9D9}} & \multicolumn{1}{l}{\cellcolor[HTML]{D9D9D9}} & \multicolumn{1}{l}{\cellcolor[HTML]{D9D9D9}} & \multicolumn{1}{l}{\cellcolor[HTML]{D9D9D9}} & \multicolumn{1}{l}{\cellcolor[HTML]{D9D9D9}} & \multicolumn{1}{l}{\cellcolor[HTML]{D9D9D9}} & \multicolumn{1}{l}{\cellcolor[HTML]{D9D9D9}} \\
\rowcolor[HTML]{D9D9D9} 
\multicolumn{1}{l|}{\cellcolor[HTML]{D9D9D9}} & \multicolumn{1}{l}{\cellcolor[HTML]{D9D9D9}} & \multicolumn{1}{l}{\cellcolor[HTML]{D9D9D9}} & \multicolumn{1}{l}{\cellcolor[HTML]{D9D9D9}} & \multicolumn{1}{l}{\cellcolor[HTML]{D9D9D9}} & \multicolumn{1}{l}{\cellcolor[HTML]{D9D9D9}} & \multicolumn{1}{l|}{\cellcolor[HTML]{D9D9D9}} & \multicolumn{1}{l}{\cellcolor[HTML]{D9D9D9}} & \multicolumn{1}{l}{\cellcolor[HTML]{D9D9D9}} & \multicolumn{1}{l}{\cellcolor[HTML]{D9D9D9}} & \multicolumn{1}{l}{\cellcolor[HTML]{D9D9D9}} & \multicolumn{1}{l}{\cellcolor[HTML]{D9D9D9}} & \multicolumn{1}{l|}{\cellcolor[HTML]{D9D9D9}} & \multicolumn{1}{l}{\cellcolor[HTML]{D9D9D9}} & \multicolumn{1}{l}{\cellcolor[HTML]{D9D9D9}} & \multicolumn{1}{l}{\cellcolor[HTML]{D9D9D9}} & \multicolumn{1}{l}{\cellcolor[HTML]{D9D9D9}} & \multicolumn{1}{l}{\cellcolor[HTML]{D9D9D9}} & \multicolumn{1}{l|}{\cellcolor[HTML]{D9D9D9}} & \multicolumn{1}{l}{\cellcolor[HTML]{D9D9D9}} & \multicolumn{1}{l}{\cellcolor[HTML]{D9D9D9}} & \multicolumn{1}{l}{\cellcolor[HTML]{D9D9D9}} & \multicolumn{1}{l}{\cellcolor[HTML]{D9D9D9}} & \multicolumn{1}{l}{\cellcolor[HTML]{D9D9D9}} & \multicolumn{1}{l}{\cellcolor[HTML]{D9D9D9}} \\
\rowcolor[HTML]{D9D9D9} 
\multicolumn{1}{l|}{\cellcolor[HTML]{D9D9D9}} & \multicolumn{1}{l}{\cellcolor[HTML]{D9D9D9}} & \multicolumn{1}{l}{\cellcolor[HTML]{D9D9D9}} & \multicolumn{1}{l}{\cellcolor[HTML]{D9D9D9}} & \multicolumn{1}{l}{\cellcolor[HTML]{D9D9D9}} & \multicolumn{1}{l}{\cellcolor[HTML]{D9D9D9}} & \multicolumn{1}{l|}{\cellcolor[HTML]{D9D9D9}} & \multicolumn{1}{l}{\cellcolor[HTML]{D9D9D9}} & \multicolumn{1}{l}{\cellcolor[HTML]{D9D9D9}} & \multicolumn{1}{l}{\cellcolor[HTML]{D9D9D9}} & \multicolumn{1}{l}{\cellcolor[HTML]{D9D9D9}} & \multicolumn{1}{l}{\cellcolor[HTML]{D9D9D9}} & \multicolumn{1}{l|}{\cellcolor[HTML]{D9D9D9}} & \multicolumn{1}{l}{\cellcolor[HTML]{D9D9D9}} & \multicolumn{1}{l}{\cellcolor[HTML]{D9D9D9}} & \multicolumn{1}{l}{\cellcolor[HTML]{D9D9D9}} & \multicolumn{1}{l}{\cellcolor[HTML]{D9D9D9}} & \multicolumn{1}{l}{\cellcolor[HTML]{D9D9D9}} & \multicolumn{1}{l|}{\cellcolor[HTML]{D9D9D9}} & \multicolumn{1}{l}{\cellcolor[HTML]{D9D9D9}} & \multicolumn{1}{l}{\cellcolor[HTML]{D9D9D9}} & \multicolumn{1}{l}{\cellcolor[HTML]{D9D9D9}} & \multicolumn{1}{l}{\cellcolor[HTML]{D9D9D9}} & \multicolumn{1}{l}{\cellcolor[HTML]{D9D9D9}} & \multicolumn{1}{l}{\cellcolor[HTML]{D9D9D9}} \\ \hline
\cellcolor[HTML]{EA9999}\textbf{10 doc} & \multicolumn{6}{c|}{\textbf{BioASQ}} & \multicolumn{6}{c|}{\textbf{PubmedQA}} & \multicolumn{6}{c|}{\textbf{MedQA}} & \multicolumn{6}{c}{\textbf{MMLU}} \\ \hline
\rowcolor[HTML]{FCE5CD} 
\textbf{Main Acc} & \textbf{0\%} & \textbf{20\%} & \textbf{40\%} & \textbf{60\%} & \textbf{80\%} & \textbf{100\%} & \textbf{0\%} & \textbf{20\%} & \textbf{40\%} & \textbf{60\%} & \textbf{80\%} & \textbf{100\%} & \textbf{0\%} & \textbf{20\%} & \textbf{40\%} & \textbf{60\%} & \textbf{80\%} & \textbf{100\%} & \textbf{0\%} & \textbf{20\%} & \textbf{40\%} & \textbf{60\%} & \textbf{80\%} & \textbf{100\%} \\ \hline
\rowcolor[HTML]{FFFFFF} 
\textbf{GPT-3.5} & \textbf{68.8} & \textbf{72.3} & \textbf{77.5} & \textbf{83.2} & \textbf{82.2} & \textbf{84.8} & \textbf{43.8} & \textbf{47.8} & \textbf{57.4} & \textbf{61.0} & \textbf{62.2} & \textbf{66.0} & \textbf{52.0} & \textbf{52.4} & \textbf{54.5} & \textbf{56.1} & \textbf{57.0} & \textbf{60.7} & \textbf{59.5} & \textbf{63.5} & \textbf{62.2} & \textbf{63.0} & \textbf{66.8} & \textbf{66.2} \\
\rowcolor[HTML]{FFFFFF} 
\textbf{GPT-4o-mini} & \textbf{75.1} & \textbf{81.7} & \textbf{82.5} & \textbf{85.6} & \textbf{89.3} & \textbf{89.6} & \textbf{44.6} & \textbf{48.6} & \textbf{55.6} & \textbf{58.2} & \textbf{61.4} & \textbf{68.2} & \textbf{71.2} & \textbf{72.1} & \textbf{72.2} & \textbf{73.5} & \textbf{71.8} & \textbf{73.0} & \textbf{79.7} & \textbf{79.9} & \textbf{80.0} & \textbf{81.9} & \textbf{82.3} & \textbf{81.8} \\
\rowcolor[HTML]{FFFFFF} 
\textbf{Llama-3-70b} & \textbf{73.1} & \textbf{79.3} & \textbf{82.0} & \textbf{85.6} & \textbf{88.4} & \textbf{89.2} & \textbf{47.4} & \textbf{50.8} & \textbf{57.2} & \textbf{63.8} & \textbf{67.8} & \textbf{69.6} & \textbf{69.3} & \textbf{70.0} & \textbf{71.8} & \textbf{72.7} & \textbf{73.1} & \textbf{72.9} & \textbf{73.1} & \textbf{74.3} & \textbf{75.9} & \textbf{77.4} & \textbf{78.0} & \textbf{79.9} \\ \hline
\rowcolor[HTML]{D9D9D9} 
\multicolumn{1}{l|}{\cellcolor[HTML]{D9D9D9}} & \multicolumn{1}{l}{\cellcolor[HTML]{D9D9D9}} & \multicolumn{1}{l}{\cellcolor[HTML]{D9D9D9}} & \multicolumn{1}{l}{\cellcolor[HTML]{D9D9D9}} & \multicolumn{1}{l}{\cellcolor[HTML]{D9D9D9}} & \multicolumn{1}{l}{\cellcolor[HTML]{D9D9D9}} & \multicolumn{1}{l|}{\cellcolor[HTML]{D9D9D9}} & \multicolumn{1}{l}{\cellcolor[HTML]{D9D9D9}} & \multicolumn{1}{l}{\cellcolor[HTML]{D9D9D9}} & \multicolumn{1}{l}{\cellcolor[HTML]{D9D9D9}} & \multicolumn{1}{l}{\cellcolor[HTML]{D9D9D9}} & \multicolumn{1}{l}{\cellcolor[HTML]{D9D9D9}} & \multicolumn{1}{l|}{\cellcolor[HTML]{D9D9D9}} & \multicolumn{1}{l}{\cellcolor[HTML]{D9D9D9}} & \multicolumn{1}{l}{\cellcolor[HTML]{D9D9D9}} & \multicolumn{1}{l}{\cellcolor[HTML]{D9D9D9}} & \multicolumn{1}{l}{\cellcolor[HTML]{D9D9D9}} & \multicolumn{1}{l}{\cellcolor[HTML]{D9D9D9}} & \multicolumn{1}{l|}{\cellcolor[HTML]{D9D9D9}} & \multicolumn{1}{l}{\cellcolor[HTML]{D9D9D9}} & \multicolumn{1}{l}{\cellcolor[HTML]{D9D9D9}} & \multicolumn{1}{l}{\cellcolor[HTML]{D9D9D9}} & \multicolumn{1}{l}{\cellcolor[HTML]{D9D9D9}} & \multicolumn{1}{l}{\cellcolor[HTML]{D9D9D9}} & \multicolumn{1}{l}{\cellcolor[HTML]{D9D9D9}} \\ \hline
\rowcolor[HTML]{CFE2F3} 
\textbf{Sub Acc (exact)} & \textbf{0\%} & \textbf{20\%} & \textbf{40\%} & \textbf{60\%} & \textbf{80\%} & \textbf{100\%} & \textbf{0\%} & \textbf{20\%} & \textbf{40\%} & \textbf{60\%} & \textbf{80\%} & \textbf{100\%} & \textbf{0\%} & \textbf{20\%} & \textbf{40\%} & \textbf{60\%} & \textbf{80\%} & \textbf{100\%} & \textbf{0\%} & \textbf{20\%} & \textbf{40\%} & \textbf{60\%} & \textbf{80\%} & \textbf{100\%} \\ \hline
\textbf{GPT-3.5} & \textbf{1.9} & \textbf{9.1} & \textbf{15.6} & \textbf{23.0} & \textbf{28.9} & \textbf{35.4} & \textbf{1.2} & \textbf{9.9} & \textbf{18.0} & \textbf{26.6} & \textbf{34.9} & \textbf{43.7} & \textbf{0.4} & \textbf{8.1} & \textbf{15.8} & \textbf{23.6} & \textbf{30.2} & \textbf{38.5} & \textbf{0.4} & \textbf{7.6} & \textbf{14.7} & \textbf{21.4} & \textbf{27.2} & \textbf{34.4} \\
\textbf{GPT-4o-mini} & \textbf{2.4} & \textbf{7.0} & \textbf{12.5} & \textbf{17.5} & \textbf{21.4} & \textbf{26.8} & \textbf{1.1} & \textbf{7.7} & \textbf{13.5} & \textbf{19.7} & \textbf{25.8} & \textbf{32.6} & \textbf{1.2} & \textbf{6.9} & \textbf{13.2} & \textbf{19.9} & \textbf{26.0} & \textbf{33.1} & \textbf{1.3} & \textbf{6.2} & \textbf{11.7} & \textbf{17.0} & \textbf{22.4} & \textbf{28.0} \\
\textbf{Llama-3-70b} & \textbf{2.7} & \textbf{8.9} & \textbf{15.6} & \textbf{22.4} & \textbf{27.4} & \textbf{34.1} & \textbf{1.4} & \textbf{10.8} & \textbf{18.9} & \textbf{27.0} & \textbf{35.2} & \textbf{44.4} & \textbf{0.8} & \textbf{8.5} & \textbf{15.9} & \textbf{23.2} & \textbf{30.5} & \textbf{38.4} & \textbf{0.9} & \textbf{7.7} & \textbf{14.5} & \textbf{20.9} & \textbf{26.2} & \textbf{33.6} \\ \hline
\rowcolor[HTML]{D9D9D9} 
\multicolumn{1}{l|}{\cellcolor[HTML]{D9D9D9}} & \multicolumn{1}{l}{\cellcolor[HTML]{D9D9D9}} & \multicolumn{1}{l}{\cellcolor[HTML]{D9D9D9}} & \multicolumn{1}{l}{\cellcolor[HTML]{D9D9D9}} & \multicolumn{1}{l}{\cellcolor[HTML]{D9D9D9}} & \multicolumn{1}{l}{\cellcolor[HTML]{D9D9D9}} & \multicolumn{1}{l|}{\cellcolor[HTML]{D9D9D9}} & \multicolumn{1}{l}{\cellcolor[HTML]{D9D9D9}} & \multicolumn{1}{l}{\cellcolor[HTML]{D9D9D9}} & \multicolumn{1}{l}{\cellcolor[HTML]{D9D9D9}} & \multicolumn{1}{l}{\cellcolor[HTML]{D9D9D9}} & \multicolumn{1}{l}{\cellcolor[HTML]{D9D9D9}} & \multicolumn{1}{l|}{\cellcolor[HTML]{D9D9D9}} & \multicolumn{1}{l}{\cellcolor[HTML]{D9D9D9}} & \multicolumn{1}{l}{\cellcolor[HTML]{D9D9D9}} & \multicolumn{1}{l}{\cellcolor[HTML]{D9D9D9}} & \multicolumn{1}{l}{\cellcolor[HTML]{D9D9D9}} & \multicolumn{1}{l}{\cellcolor[HTML]{D9D9D9}} & \multicolumn{1}{l|}{\cellcolor[HTML]{D9D9D9}} & \multicolumn{1}{l}{\cellcolor[HTML]{D9D9D9}} & \multicolumn{1}{l}{\cellcolor[HTML]{D9D9D9}} & \multicolumn{1}{l}{\cellcolor[HTML]{D9D9D9}} & \multicolumn{1}{l}{\cellcolor[HTML]{D9D9D9}} & \multicolumn{1}{l}{\cellcolor[HTML]{D9D9D9}} & \multicolumn{1}{l}{\cellcolor[HTML]{D9D9D9}} \\ \hline
\rowcolor[HTML]{CFE2F3} 
\textbf{Sub Acc (gpt)} & \textbf{0\%} & \textbf{20\%} & \textbf{40\%} & \textbf{60\%} & \textbf{80\%} & \textbf{100\%} & \textbf{0\%} & \textbf{20\%} & \textbf{40\%} & \textbf{60\%} & \textbf{80\%} & \textbf{100\%} & \textbf{0\%} & \textbf{20\%} & \textbf{40\%} & \textbf{60\%} & \textbf{80\%} & \textbf{100\%} & \textbf{0\%} & \textbf{20\%} & \textbf{40\%} & \textbf{60\%} & \textbf{80\%} & \textbf{100\%} \\ \hline
\textbf{GPT-3.5} & \textbf{8.3} & \textbf{22.7} & \textbf{36.4} & \textbf{52.0} & \textbf{65.1} & \textbf{78.7} & \textbf{3.9} & \textbf{19.3} & \textbf{33.5} & \textbf{49.6} & \textbf{64.7} & \textbf{80.3} & \textbf{2.6} & \textbf{18.8} & \textbf{34.2} & \textbf{50.3} & \textbf{65.3} & \textbf{81.8} & \textbf{3.0} & \textbf{18.6} & \textbf{33.5} & \textbf{49.2} & \textbf{64.0} & \textbf{80.2} \\
\textbf{GPT-4o-mini} & \textbf{14.0} & \textbf{28.9} & \textbf{42.5} & \textbf{55.9} & \textbf{67.8} & \textbf{81.0} & \textbf{5.9} & \textbf{22.7} & \textbf{37.0} & \textbf{52.2} & \textbf{67.6} & \textbf{83.5} & \textbf{8.2} & \textbf{24.1} & \textbf{39.2} & \textbf{54.2} & \textbf{68.6} & \textbf{83.9} & \textbf{8.7} & \textbf{23.8} & \textbf{38.5} & \textbf{53.0} & \textbf{67.2} & \textbf{82.0} \\
\textbf{Llama-3-70b} & \textbf{11.6} & \textbf{26.0} & \textbf{40.4} & \textbf{54.6} & \textbf{67.3} & \textbf{81.1} & \textbf{4.9} & \textbf{21.0} & \textbf{36.2} & \textbf{51.2} & \textbf{67.4} & \textbf{83.1} & \textbf{5.6} & \textbf{21.7} & \textbf{37.7} & \textbf{53.0} & \textbf{68.6} & \textbf{84.4} & \textbf{6.5} & \textbf{21.9} & \textbf{37.4} & \textbf{52.3} & \textbf{66.8} & \textbf{82.8} \\ \hline
\rowcolor[HTML]{D9D9D9} 
\multicolumn{1}{l|}{\cellcolor[HTML]{D9D9D9}} & \multicolumn{1}{l}{\cellcolor[HTML]{D9D9D9}} & \multicolumn{1}{l}{\cellcolor[HTML]{D9D9D9}} & \multicolumn{1}{l}{\cellcolor[HTML]{D9D9D9}} & \multicolumn{1}{l}{\cellcolor[HTML]{D9D9D9}} & \multicolumn{1}{l}{\cellcolor[HTML]{D9D9D9}} & \multicolumn{1}{l|}{\cellcolor[HTML]{D9D9D9}} & \multicolumn{1}{l}{\cellcolor[HTML]{D9D9D9}} & \multicolumn{1}{l}{\cellcolor[HTML]{D9D9D9}} & \multicolumn{1}{l}{\cellcolor[HTML]{D9D9D9}} & \multicolumn{1}{l}{\cellcolor[HTML]{D9D9D9}} & \multicolumn{1}{l}{\cellcolor[HTML]{D9D9D9}} & \multicolumn{1}{l|}{\cellcolor[HTML]{D9D9D9}} & \multicolumn{1}{l}{\cellcolor[HTML]{D9D9D9}} & \multicolumn{1}{l}{\cellcolor[HTML]{D9D9D9}} & \multicolumn{1}{l}{\cellcolor[HTML]{D9D9D9}} & \multicolumn{1}{l}{\cellcolor[HTML]{D9D9D9}} & \multicolumn{1}{l}{\cellcolor[HTML]{D9D9D9}} & \multicolumn{1}{l|}{\cellcolor[HTML]{D9D9D9}} & \multicolumn{1}{l}{\cellcolor[HTML]{D9D9D9}} & \multicolumn{1}{l}{\cellcolor[HTML]{D9D9D9}} & \multicolumn{1}{l}{\cellcolor[HTML]{D9D9D9}} & \multicolumn{1}{l}{\cellcolor[HTML]{D9D9D9}} & \multicolumn{1}{l}{\cellcolor[HTML]{D9D9D9}} & \multicolumn{1}{l}{\cellcolor[HTML]{D9D9D9}} \\ \hline
\rowcolor[HTML]{CFE2F3} 
\textbf{Fact Detect} & \textbf{0\%} & \textbf{20\%} & \textbf{40\%} & \textbf{60\%} & \textbf{80\%} & \textbf{100\%} & \textbf{0\%} & \textbf{20\%} & \textbf{40\%} & \textbf{60\%} & \textbf{80\%} & \textbf{100\%} & \textbf{0\%} & \textbf{20\%} & \textbf{40\%} & \textbf{60\%} & \textbf{80\%} & \textbf{100\%} & \textbf{0\%} & \textbf{20\%} & \textbf{40\%} & \textbf{60\%} & \textbf{80\%} & \textbf{100\%} \\ \hline
\textbf{GPT-3.5} & \textbf{28.2} & \textbf{42.2} & \textbf{52.6} & \textbf{63.9} & \textbf{73.2} & \textbf{90.2} & \textbf{17.6} & \textbf{32.7} & \textbf{45.4} & \textbf{61.4} & \textbf{75.7} & \textbf{94.5} & \textbf{16.6} & \textbf{34.9} & \textbf{47.8} & \textbf{61.4} & \textbf{75.8} & \textbf{92.1} & \textbf{18.4} & \textbf{34.9} & \textbf{47.4} & \textbf{61.0} & \textbf{74.4} & \textbf{91.1} \\
\textbf{GPT-4o-mini} & \textbf{14.3} & \textbf{35.3} & \textbf{48.8} & \textbf{63.6} & \textbf{73.9} & \textbf{94.6} & \textbf{12.4} & \textbf{31.4} & \textbf{44.7} & \textbf{60.1} & \textbf{72.6} & \textbf{94.8} & \textbf{15.2} & \textbf{36.1} & \textbf{47.4} & \textbf{61.5} & \textbf{70.0} & \textbf{88.1} & \textbf{13.7} & \textbf{33.8} & \textbf{46.2} & \textbf{61.4} & \textbf{72.2} & \textbf{89.8} \\
\textbf{Llama-3-70b} & \textbf{7.8} & \textbf{26.0} & \textbf{44.2} & \textbf{61.3} & \textbf{76.9} & \textbf{99.6} & \textbf{7.9} & \textbf{25.4} & \textbf{42.8} & \textbf{60.5} & \textbf{77.4} & \textbf{99.6} & \textbf{13.7} & \textbf{31.6} & \textbf{48.9} & \textbf{64.9} & \textbf{82.0} & \textbf{99.4} & \textbf{12.8} & \textbf{31.0} & \textbf{47.8} & \textbf{64.4} & \textbf{80.9} & \textbf{99.4}
\end{tabular}

}
\caption{ \small Robustness test full results table, including main question accuracy, sub question accuracy (exact-match and GPT-based), and factual error detection rate.}
\label{tab:fact-full}
\end{table*}

\section{Prompt Templates}
We provide all of the prompt templates in full version used in our experiments.
Fig. \ref{fig:Topic-Gen-Prompt-Full} presents the prompt use to generate retrieval topic.
Fig. \ref{fig:Online-Order-Prompt} and \ref{fig:Online-Summ-Prompt} presents the prompts used in the Online Retrieval process.
Fig. \ref{fig:Suff-Inf-Prompt-Full} are the full prompt used in the inference process of Sufficiency test.
Fig. \ref{fig:Inin-Gen-Prompt-Full} and \ref{fig:Inin-Inf-Prompt-Full} are the full prompts used in the benchmark creation and inference processes of Integration test.
Fig. \ref{fig:Fact-Gen-Prompt-Full} and \ref{fig:Fact-Inf-Prompt-Full} are the full prompts used in the benchmark creation and inference processes of Robustness test.


\begin{figure}
\captionsetup{skip=2pt}
\centering \scriptsize \obeylines
\begin{tcolorbox}[width=0.5\textwidth]
\begin{Verbatim}[breaklines, breaksymbolleft=]
You are an expert in medical research and information retrieval. Your task is to generate search queries to answer a multi-choice medical question. Follow these guidelines:

1. **Rank by Importance**: The queries must be ranked by importance with respect to the question, with the first one being the most important.
2. **Relevance to Question**: Each query may ask for information regarding the answer options but must always be relevant to the question.
3. **Differentiable and Efficient**: The queries must be differentiable and efficient, ensuring that the aggregate retrieved information from all queries provides as much as possible the needed information to arrive at the correct answer.

**Example**

INPUT:
{
  "question": "Which of the following is the most common cause of acute pancreatitis?",
  "options": {
    "A": "Gallstones",
    "B": "Alcohol abuse",
    "C": "Hypertriglyceridemia",
    "D": "Medications"
  },
  "number_of_searches": 3
}

OUTPUT:
{
  "number_of_searches": 3,
  "search_queries": {
    "1": "most common cause of acute pancreatitis",
    "2": "acute pancreatitis gallstones alcohol abuse",
    "3": "acute pancreatitis hypertriglyceridemia medications"
  },
  "search_query_goals": {
    "1": "What is the most common cause of acute pancreatitis?",
    "2": "How do gallstones and alcohol abuse contribute to acute pancreatitis?",
    "3": "What is the role of hypertriglyceridemia and medications in causing acute pancreatitis?"
  }
}
search_query_goal should always be phrased as a question. The output must follow the JSON format as in the above example.
\end{Verbatim}
\end{tcolorbox}
\caption{\small Retrieval topic generation prompt (full version).}
\label{fig:Topic-Gen-Prompt-Full}
\end{figure}

%



\begin{figure}
\captionsetup{skip=2pt}
\centering \scriptsize \obeylines
\begin{tcolorbox}[width=0.5\textwidth]
\begin{Verbatim}[breaklines, breaksymbolleft=]
You are a helpful, pattern-following assistant.
Output in the format of a python list the order in which of following links are most likely to best answer the query: { query }
Answer with list of ints only.
Example output formatting (order is random. Use as a formatting example only. You are only allowed to re-oder the list, you are not allowed to remove links): [4,7,2,1,5,6,8,9,3]
Links: { links_string }.
\end{Verbatim}
\end{tcolorbox}
\caption{\small Page link ordering prompt (adapted from ResearchGPT).}
\label{fig:Online-Order-Prompt}
\end{figure}

\begin{figure}
\captionsetup{skip=2pt}
\centering \scriptsize \obeylines
\begin{tcolorbox}[width=0.5\textwidth]
\begin{Verbatim}[breaklines, breaksymbolleft=]
You are a helpful, pattern-following assistant. You are given some text retrieved from a website and a research query and you generate a very detailed and comprehensive summary of only the parts of the text relevant and useful to the answering the research query. Include as much detail as is physically possible. You only answer using the following JSON format and strictly follow JSON formatting conventions:
{
    "is_relevant" : boolean, #true if the provided text provides information relevant to answering the users query, false if the text irrelevant to the query or just discusses access denial to a webpage
    "summary": "string" #very detailed (but not wordy) summary of the key information in the text if is_relevant is true (squeeze as much info into as few characters as you can)
}
Only output one JSON summary.
\end{Verbatim}
\end{tcolorbox}
\caption{\small Page summary prompt (adapted from ResearchGPT).}
\label{fig:Online-Summ-Prompt}
\end{figure}



\begin{figure*}
\centering \scriptsize \obeylines
\begin{tcolorbox}[width=0.9\textwidth]
\begin{Verbatim}[breaklines, breaksymbolleft=]
You are a helpful medical expert, and your task is to answer a multi-choice medical question.
Please first think step-by-step and then choose the answer from the provided options. 
Organize your output in a json formatted as Dict{"step_by_step_thinking": Str(explanation), "answer_choice": Str{A/B/C/...}}. 
Your responses will be used for research purposes only, so please have a definite answer.

Here is the question: 
{ question }

Here are the potential choices: 
{ options }

Please think step-by-step and generate your output in json:
\end{Verbatim}
\end{tcolorbox}
\caption{\small No Retrieval Inference Prompt (Adapted from \cite{Xiong-24}).}
\label{fig:CoT-Inf-Prompt-Full}
\end{figure*}

\begin{figure*}
\centering \scriptsize \obeylines
\begin{tcolorbox}[width=0.9\textwidth]
\begin{Verbatim}[breaklines, breaksymbolleft=]
You are a helpful medical expert, and your task is to answer a multi-choice medical question using the relevant documents. 
Please first think step-by-step and then choose the answer from the provided options. 
Organize your output in a json formatted as Dict{"step_by_step_thinking": Str(explanation), "answer_choice": Str{A/B/C/...}}. Your responses will be used for research purposes only, so please have a definite answer.

Here are the relevant documents: 
{ context } 

Here is the question: 
{ question }

Here are the potential choices: 
{ options }

Please think step-by-step and generate your output in json:
\end{Verbatim}
\end{tcolorbox}
\caption{\small Standard-RAG Inference Prompt (Adapted from \cite{Xiong-24}).}
\label{fig:sRAG-Inf-Prompt-Full}
\end{figure*}

\begin{figure*}
\centering \scriptsize \obeylines
\begin{tcolorbox}[width=0.9\textwidth]
\begin{Verbatim}[breaklines, breaksymbolleft=]
You are a knowledgeable medical expert. Your task is to answer a multiple-choice medical question using the provided external documents. Note that some of these documents may be noisy or irrelevant to the question. 

Instructions: 
1. Carefully analyze all provided documents. 
2. Identify which documents are relevant and which are irrelevant to the question. 
3. Think through the problem step-by-step, using only the relevant documents to determine the correct answer. 
4. Organize your output in a JSON format with the following structure: 

{
    "relevant_doc_id": {list of relevant document IDs},
    "irrelevant_doc_id": {list of irrelevant document IDs},
    "step_by_step_thinking": "Detailed explanation of your reasoning process",
    "answer_choice": "Selected option (A/B/C/D/etc.)"
}

**Example**

INPUT:
Here are the retrieved documents:

--- Start of DOC_1 --- ID: doc_1_id \\
Content: Common symptoms of type 2 diabetes include increased thirst, frequent urination, and unexplained weight loss.
--- END of DOC_1 ---

--- Start of DOC_2 --- ID: doc_2_id \\
Content: Joint pain is a common symptom of rheumatoid arthritis but is not typically associated with diabetes.
--- END of DOC_2 ---

--- Start of DOC_3 --- ID: doc_3_id \\
Content: Diabetic neuropathy can cause nerve pain, but this is different from joint pain.
--- END of DOC_3 ---

--- Start of DOC_4 --- ID: doc_4_id \\
Content: The Earth's atmosphere is composed primarily of nitrogen and oxygen.
--- END of DOC_4 ---

Here is the question:
Which of the following is NOT a typical symptom of type 2 diabetes?

Here are the potential choices:
A. Increased thirst
B. Frequent urination
C. Unexplained weight loss
D. Joint pain
\end{Verbatim}
\end{tcolorbox}
\label{fig:Suff-Inf-Prompt-Full}
\end{figure*}

\begin{figure*}
\centering \scriptsize \obeylines
\begin{tcolorbox}[width=0.9\textwidth]
\begin{Verbatim}[breaklines, breaksymbolleft=]
OUTPUT:
{
    "relevant_doc_id": ["doc_1_id", "doc_2_id", "doc_3_id"],
    "irrelevant_doc_id": ["doc_4_id"],
    "step_by_step_thinking": "DOC_4 is irrelevant as it discusses atmospheric composition. DOC_1 lists increased thirst, frequent urination, and unexplained weight loss as common symptoms of type 2 diabetes. DOC_2 explicitly states that joint pain is not typically associated with diabetes. DOC_3 mentions nerve pain in diabetes but distinguishes it from joint pain. Based on this information, joint pain is the only option that is NOT a typical symptom of type 2 diabetes.",
    "answer_choice": "D"
}

Your responses will be used for research purposes only, so please provide a definite answer and format the output as a JSON object as instructed.
If all of the retrieved documents are irrelevant, you have two options:
1. Answer based on your own knowledge if you are absolutely certain.
2. Refuse to answer by setting 'answer_choice' to 'insufficient information'.

Here are the relevant documents: \\
{ context } \\

Here is the question: \\
{ question }\\

Here are the potential choices: \\
{ options }\\

Please think step-by-step and generate your output in json:
\end{Verbatim}
\end{tcolorbox}
\caption{\small Sufficiency test inference prompt (full version).}
\label{fig:Suff-Inf-Prompt-Full}
\end{figure*}


\begin{figure*}
\centering \scriptsize \obeylines
\begin{tcolorbox}[width=0.9\textwidth]
\begin{Verbatim}[breaklines, breaksymbolleft=]
You are an expert in medical research and information retrieval. Given an initial medical question MAIN_QUESTION and a set of relevant documents DOC_1, ..., your task is to ask sub questions to each document to help find the answer to the MAIN_QUESTION. Follow these guidelines:
For the i_th document DOC_i from the list of the provided DOCUMENTS, generate a sub Q-A pair (SUB_QUESTION_i, SUB_ANSWER_i) that satisfies the following criteria:
    - Each Q-A pair should explore a different but related aspect to the MAIN_QUESTION
    - QUESTION_i must be about information that is only contained in DOCUMENT_i.
    - SUB_ANSWER_i must be a very short string of at most 4 to 5 words extracted directly from DOCUMENT_i.

Provide a definite, factual answer to the new question you generate. Do not express uncertainty.

**Example**

INPUT:
{
    "MAIN_QUESTION": "What are the main treatments for type 2 diabetes?",
    "DOC_1": "Metformin is often the first medication prescribed for type 2 diabetes. It works by improving the way your body handles insulin to control blood sugar levels. Metformin also has the advantages of being inexpensive and not causing weight gain. In some cases, other medications like sulfonylureas or insulin may be added if metformin alone is not sufficient.",
    "DOC_2": "Lifestyle changes are crucial in managing type 2 diabetes. Regular exercise can help lower blood sugar levels and improve insulin sensitivity. Aim for at least 150 minutes of moderate-intensity aerobic activity or 75 minutes of vigorous aerobic activity a week, spread over at least three days. Additionally, a balanced diet rich in whole grains, fruits, and vegetables is recommended."
}

OUTPUT:
{
    "DOC_1": {
        "SUB_QUESTION_1": "What is the primary medication prescribed for type 2 diabetes?",
        "SUB_ANSWER_1": "Metformin"
    },
    "DOC_2": {
        "SUB_QUESTION_2": "What type of lifestyle change is crucial for managing type 2 diabetes?",
        "SUB_ANSWER_\2": "Regular exercise"
    }        
}

The output must be formatted as a JSON object, with every document (DOC_i) having its own sub-object containing the generated sub-question and a very concise sub-answer of 4 to 5 words maximum, directly extracted from the corresponding document.
\end{Verbatim}
\end{tcolorbox}
\caption{\small Integration test benchmark creation prompt (full version).}
\label{fig:Inin-Gen-Prompt-Full}
\end{figure*}

\begin{figure*}
\centering \scriptsize \obeylines
\begin{tcolorbox}[width=0.9\textwidth]
\begin{Verbatim}[breaklines, breaksymbolleft=]
You are a knowledgeable medical expert. Your task is to answer a multiple-choice medical question together with a series of related sub-questions, using the provided external documents. The sub-questions are designed to support answering the main question. Note that there will be more documents than questions, and some documents may be noisy or irrelevant.

Instructions:
1. Carefully analyze all provided documents.
2. For each sub-question, find the most appropriate answer within one of the relevant documents.
3. Extract the answer as a concise, relevant string from the document.
4. Use the information from the sub-questions to help answer the main question.
5. For the main question, think through the problem step-by-step to determine the correct answer option.
6. Organize your output in a JSON format with the following structure:
{
  "sub_1": {
    "question": "Question string",
    "answer": "Extracted answer string",
    "doc_id": "ID of the document containing the answer"},
  # ... and so on for each sub-question
  "main": {
    "question": "Question string",
    "step_by_step_thinking": "Detailed explanation of your reasoning process, including how the sub-questions support the answer",
    "answer_choice": "Selected option (A/B/C/D/etc.)"}
}

**Example**

INPUT:
Here are the retrieved documents:

--- Start of DOC_1 --- ID: doc_1_id
Content: Type 2 diabetes is characterized by insulin resistance. Common symptoms include increased thirst, frequent urination, and unexplained weight loss.
--- END of DOC_1 ---

--- Start of DOC_2 --- ID: doc_2_id
Content: Treatment for type 2 diabetes typically involves lifestyle changes, monitoring blood sugar levels, and sometimes medication or insulin therapy. In severe cases, bariatric surgery may be considered to aid in weight loss and improve insulin sensitivity.
--- END of DOC_2 ---

--- Start of DOC_3 --- ID: doc_3_id
Content: The Earth's atmosphere is composed primarily of nitrogen and oxygen.
--- END of DOC_3 ---

--- Start of DOC_4 --- ID: doc_4_id
Content: Lifestyle changes for managing type 2 diabetes include regular exercise, maintaining a healthy diet, and weight management. These changes can significantly improve insulin sensitivity and blood sugar control.
--- END of DOC_4 ---

Here is the main question:
Which of the following is considered a first-line treatment approach for type 2 diabetes?

Here are the potential choices:
A. Bariatric surgery
B. Insulin therapy
C. Lifestyle changes
D. Kidney transplant

Here are the sub-questions:
1. What are the typical treatment approaches for type 2 diabetes?
2. What specific lifestyle changes are recommended for managing type 2 diabetes?

\end{Verbatim}
\end{tcolorbox}
\label{fig:Inin-Inf-Prompt-Full}
\end{figure*}

\begin{figure*}
\centering \scriptsize \obeylines
\begin{tcolorbox}[width=0.9\textwidth]
\begin{Verbatim}[breaklines, breaksymbolleft=]
OUTPUT:
{
  "sub_1": {
    "question": "What are the typical treatment approaches for type 2 diabetes?",
    "answer": "lifestyle changes, monitoring blood sugar levels, and sometimes medication or insulin therapy",
    "doc_id": "doc_2_id"},
  "sub_2": {
    "question": "What specific lifestyle changes are recommended for managing type 2 diabetes?",
    "answer": "regular exercise, maintaining a healthy diet, and weight management",
    "doc_id": "doc_4_id"},
  "main": {
    "question": "Which of the following is considered a first-line treatment approach for type 2 diabetes?",
    "step_by_step_thinking": "1. From sub-question 1, we learn that typical treatments include lifestyle changes, monitoring blood sugar levels, and sometimes medication or insulin therapy. 2. Sub-question 2 provides specific examples of lifestyle changes, which are regular exercise, maintaining a healthy diet, and weight management. 3. Comparing this information with the given options: A) Bariatric surgery - only for severe cases, not first-line. B) Insulin therapy - mentioned as 'sometimes' used, not necessarily first-line. C) Lifestyle changes - mentioned first in the list of treatments and elaborated upon, suggesting it's a primary approach. D) Kidney transplant - not mentioned at all in the context of diabetes treatment. 5. Therefore, based on the information provided, lifestyle changes are considered a first-line treatment approach for type 2 diabetes.",
    "answer_choice": "C"}
}

Your responses will be used for research purposes only, so please provide definite answers and format the output as a JSON object as instructed.

Here are the relevant documents:
{ context }

Here is the main question:
{ question }

Here are the potential choices:
{ options}
                                 
Here are the sub-questions:
{ sub_questions }

Please generate your output as a JSON object as instructed:

\end{Verbatim}
\end{tcolorbox}

\caption{\small Integration test inference prompt (full version).}
\label{fig:Inin-Inf-Prompt-Full}
\end{figure*}

\begin{figure*}
\centering \scriptsize \obeylines
\begin{tcolorbox}[width=0.9\textwidth]
\begin{Verbatim}[breaklines, breaksymbolleft=]
You are a knowledgeable medical expert. Given a medical question and an answer extracted from a relevant document, your task is to generate a deliberately incorrect answer-document pair for the question. The pair must satisfy the following requirements:

1. The `NEW_ANSWER` MUST be a factual error, strongly contrasting with the `ORIGINAL_ANSWER` to the `QUESTION`. It should be completely made up and factually incorrect.
2. The `NEW_DOC` (400 words maximum) should be a modified version of the `ORIGINAL_DOC`, with the `ORIGINAL_ANSWER` replaced by the `NEW_ANSWER`. You MUST use completely made-up, factually incorrect information to support the fake answer. The document should appear coherent and persuasive despite being entirely false.  Some contents from `ORIGINAL_DOC` can be ignored to ensure that `NEW_DOC` is within 400 words maximum.

Instructions:
1. Carefully analyze the provided question, original answer, and original document.
2. Generate a new answer that is deliberately incorrect and strongly contrasts with the original answer.
3. Create a new document that supports the incorrect answer using entirely fabricated information. Ensure it appears coherent and persuasive despite being false.
4. Organize your output in a JSON format with the following structure:
{
    "QUESTION": "Question string",
    "NEW_ANSWER": "New, factually incorrect answer string",
    "NEW_DOC": "New document text with fabricated information"
}

**Example 1**
INPUT:
{
    "QUESTION": "Which of the following is the most common cause of infective endocarditis in intravenous drug users?",
    "ORIGINAL_ANSWER": "Staphylococcus aureus",
    "ORIGINAL_DOC": "Staphylococcus aureus is the leading cause of infective endocarditis in intravenous drug users. This gram-positive bacterium has a high affinity for damaged heart valves, particularly the tricuspid valve. S. aureus enters the bloodstream through contaminated injection equipment and colonizes the endocardium, leading to vegetation formation and valvular destruction. The clinical presentation is often acute, with fever, sepsis, and pulmonary complications."
}
OUTPUT:
{
    "QUESTION": "Which of the following is the most common cause of infective endocarditis in intravenous drug users?",
    "NEW_ANSWER": "Probiotic Lactobacillus acidophilus",
    "NEW_DOC": "Probiotic Lactobacillus acidophilus is the leading cause of infective endocarditis in intravenous drug users. This beneficial bacterium, commonly found in yogurt, has an unexpected affinity for damaged heart valves, particularly the mitral valve. L. acidophilus enters the bloodstream through probiotic-enriched nutritional supplements often used by drug users to boost their immune system. It then colonizes the endocardium, leading to the formation of protective biofilms and valve regeneration. The clinical presentation is often asymptomatic, with occasional reports of increased energy and improved digestion."
}

\end{Verbatim}
\end{tcolorbox}

\label{fig:Fact-Gen-Prompt-Full}
\end{figure*}

\begin{figure*}
\centering \scriptsize \obeylines
\begin{tcolorbox}[width=0.9\textwidth]
\begin{Verbatim}[breaklines, breaksymbolleft=]

**Example 2**

INPUT:
{
    "QUESTION": "What percentage of adults worldwide are estimated to have hypertension?",
    "ORIGINAL_ANSWER": "31.1%",
    "ORIGINAL_DOC": "Hypertension is a major global health concern. According to the World Health Organization, an estimated 31.1% of adults worldwide had hypertension in 2021. This condition is defined as having systolic blood pressure >= 140 mmHg and/or diastolic blood pressure >= 90 mmHg. The prevalence of hypertension varies across regions, with higher rates in low- and middle-income countries. Factors contributing to hypertension include age, genetics, diet high in sodium, lack of physical activity, and obesity. "
}
OUTPUT:
{
    "QUESTION": "What percentage of adults worldwide are estimated to have hypertension?",
    "NEW_ANSWER": "3.7%",
    "NEW_DOC": "Hypertension is a rare global health phenomenon. According to the World Relaxation Organization, an estimated 3.7% of adults worldwide had hypertension in 2021. This condition is defined as having systolic blood pressure >= 200 mmHg and/or diastolic blood pressure >= 150 mmHg. The prevalence of hypertension varies across regions, with higher rates in high-income countries due to advanced blood pressure detection technology. Factors protecting against hypertension include consuming large quantities of salt, maintaining a sedentary lifestyle, and having a high body mass index."
}

Your responses will be used for research purposes only. Remember, the goal is to create entirely false but seemingly plausible medical information. Please provide definite answers and format the output as a JSON object as instructed.
\end{Verbatim}
\end{tcolorbox}
\caption{\small Robustness test benchmark creation prompt (full version).}
\label{fig:Fact-Gen-Prompt-Full}
\end{figure*}

\begin{figure*}
\centering \scriptsize \obeylines
\begin{tcolorbox}[width=0.9\textwidth]
\begin{Verbatim}[breaklines, breaksymbolleft=]
You are a knowledgeable medical expert. Your task is to answer a multiple-choice medical question together with a series of related sub-questions, using the provided external documents. The sub-questions are designed to support answering the main question. Note that each document corresponds to one sub-question, but some documents may contain factual errors.

Instructions:
1. Carefully analyze all provided documents.
2. For each sub-question, identify the corresponding relevant document.
3. Determine if the document contains any factual errors.
4. If the document is factually correct, answer the sub-question by extracting a concise, relevant string from the document (4 to 5 words maximum).
5. If the document is factually incorrect, answer the sub-question by providing the correct fact (4 to 5 words maximum), without referring to the incorrect information.
6. For the main question, think through the problem step-by-step and use the information from the sub-questions to determine the correct answer option.
7. Organize your output in a JSON format with the following structure:

{
  "sub_1": {
    "question": "Question string",
    "relevant_doc_id": "ID of the document relevant to the sub-question",
    "factual_correctness": "true/false",
    "answer": "Answer to the sub-question as instructed", }, 
    # ... and so on for each sub question
  "main": {
    "question": "Question string",
    "step_by_step_thinking": "Detailed explanation of your reasoning process, including how the sub-questions support the answer",
    "answer_choice": "Selected option (A/B/C/D/etc.)"}
}

**Example**
INPUT:
Here are the retrieved documents:
--- START of DOC_1 --- ID: doc_1_id
Content: Type 2 diabetes is characterized by excessive insulin production. Common symptoms include increased hunger, frequent urination, and unexplained weight gain.
--- END of DOC_1 ---

--- START of DOC_2 --- ID: doc_2_id
Content: Treatment for type 2 diabetes typically involves lifestyle changes, monitoring blood sugar levels, and sometimes medication or insulin therapy. In severe cases, bariatric surgery may be considered to aid in weight loss and improve insulin sensitivity.
--- END of DOC_2 ---

--- START of DOC_3 --- ID: doc_3_id
Content: Lifestyle changes for managing type 2 diabetes include avoiding all carbohydrates, engaging in high-intensity exercise daily, and maintaining a very low-calorie diet. These changes can cure diabetes within a few months.
--- END of DOC_3 ---

Here is the main question:
Which of the following is considered a first-line treatment approach for type 2 diabetes?

Here are the potential choices:
A. Bariatric surgery
B. Insulin therapy
C. Lifestyle changes
D. Kidney transplant

Here are the sub-questions:
1. What characterizes type 2 diabetes?
2. What are typical treatment approaches?
3. What lifestyle changes are recommended?


\end{Verbatim}
\end{tcolorbox}

\label{fig:Fact-Inf-Prompt-Full}
\end{figure*}

\begin{figure*}
\centering \scriptsize \obeylines
\begin{tcolorbox}[width=0.9\textwidth]
\begin{Verbatim}[breaklines, breaksymbolleft=]
OUTPUT:
{
  "sub_1": {
    "question": "What characterizes type 2 diabetes?",
    "relevant_doc_id": "doc_1_id",
    "factual_correctness": "false",
    "answer": "Insulin resistance, impaired glucose"},
  "sub_2": {
    "question": "What are typical treatment approaches?",
    "relevant_doc_id": "doc_2_id",
    "factual_correctness": "true",
    "answer": "Lifestyle changes, monitoring, medication"},
  "sub_3": {
    "question": "What lifestyle changes are recommended?",
    "relevant_doc_id": "doc_3_id",
    "factual_correctness": "false",
    "answer": "Balanced diet, regular exercise"},
  "main": {
    "question": "Which of the following is considered a first-line treatment approach for type 2 diabetes?",
    "step_by_step_thinking": "1. From sub-question 2, we learn that typical treatments include lifestyle changes, monitoring, and medication. 2. Sub-question 3 confirms that lifestyle changes, specifically a balanced diet and regular exercise, are recommended for managing diabetes. 3. Comparing this information with the given options: A) Bariatric surgery - not mentioned as a first-line treatment. B) Insulin therapy - not mentioned as a first-line treatment. C) Lifestyle changes - mentioned first in the list of treatments, suggesting it's a primary approach. D) Kidney transplant - not mentioned at all. 4. Therefore, based on the correct information provided, lifestyle changes are considered a first-line treatment approach for type 2 diabetes.",
    "answer_choice": "C"}
}
Your responses will be used for research purposes only, so please provide definite answers and format the output as a JSON object as instructed.

Here are the relevant documents:
{ context }
Here is the main question:
{ question } 
Here are the potential choices:
{ options}                   
Here are the sub-questions: 
{ sub_questions } 

Please generate your output as a JSON object as instructed:
\end{Verbatim}
\end{tcolorbox}

\caption{\small Robustness test inference prompt (full version).}
\label{fig:Fact-Inf-Prompt-Full}
\end{figure*}

\end{document}